%% file: 00_main.tex
\documentclass[10pt,twocolumn,letterpaper]{article}

\usepackage[usenames,dvipsnames]{xcolor}
\usepackage{iccv}
\usepackage{times}
\usepackage{epsfig}
\usepackage{graphicx}
\usepackage{amsmath}
\usepackage{multicol}
\usepackage{amssymb}
\usepackage{flushend}
\usepackage{capt-of}
\usepackage{multirow}
\usepackage[accsupp]{axessibility} 
\usepackage[numbers,sort]{natbib}
\usepackage[normalem]{ulem}

\input{macros}

\usepackage[breaklinks=true,bookmarks=false]{hyperref}

\iccvfinalcopy

\def\iccvPaperID{10231}

\ificcvfinal\fi

\begin{document}
\title{Preface: A Data-driven Volumetric Prior \\for Few-shot Ultra High-resolution Face Synthesis}

\author{Marcel C. B\"uhler$^{1,2}$ \quad Kripasindhu Sarkar$^{2}$ \quad Tanmay Shah$^{2}$ \quad Gengyan Li$^{1,2}$ \quad Daoye Wang$^{2}$ \\Leonhard Helminger$^{2}$ \quad Sergio Orts-Escolano$^{2}$ \quad Dmitry Lagun$^{2}$ \\ Otmar Hilliges$^{1}$ \quad Thabo Beeler$^{2}$ 
\quad Abhimitra Meka$^{2}$ \vspace{0.1cm} \\
$^1$ETH Zurich \quad $^2$Google\\
{\small \href{https://syntec-research.github.io/Preface}{https://syntec-research.github.io/Preface}}\vspace{-0.7cm}
}

\input{figures/00_teaser}
\begin{abstract}
NeRFs have enabled highly realistic synthesis of human faces including complex appearance and reflectance effects of hair and skin. 
These methods typically require a large number of multi-view input images, making the process hardware intensive and cumbersome, limiting applicability to unconstrained settings.
We propose a novel volumetric human face prior that enables the synthesis of ultra high-resolution novel views of subjects that are not part of the prior's training distribution. 
This prior model consists of an identity-conditioned NeRF, trained on a dataset of low-resolution multi-view images of diverse humans with known camera calibration.
A simple sparse landmark-based 3D alignment of the training dataset allows our model to learn a smooth latent space of geometry and appearance despite a limited number of training identities. 
A high-quality volumetric representation of a novel subject can be obtained by model fitting to 2 or 3 camera views of arbitrary resolution. Importantly, our method requires as few as two views of casually captured images as input at inference time.
\end{abstract}

%%%%%%%%% BODY TEXT
\input{01_intro}

\input{02_related}
\input{04_method}
\input{03_dataset}
\input{05_experiments}
\input{06_results}
\input{07_conclusion}

\small
\noindent\textbf{Acknowledgments.}
We thank Emre Aksan for insightful discussions and Malte Prinzler for sharing DINER results.

{\small
\bibliographystyle{ieee_fullname}
\bibliography{egbib}
}
\clearpage
\appendix
\input{99_supp}

\end{document}

%% file: macros.tex
\makeatletter
\newcommand*{\addFileDependency}[1]{
  \typeout{(#1)}
  \@addtofilelist{#1}
  \IfFileExists{#1}{}{\typeout{No file #1.}}
}
\makeatother

\DeclareMathOperator*{\argmin}{arg\,min} % Jan Hlavacek

%% file: figures/00_teaser.tex
\twocolumn[{
\renewcommand\twocolumn[1][]{#1}
\maketitle
\begin{center}
  \centerline{\includegraphics[width=\textwidth]{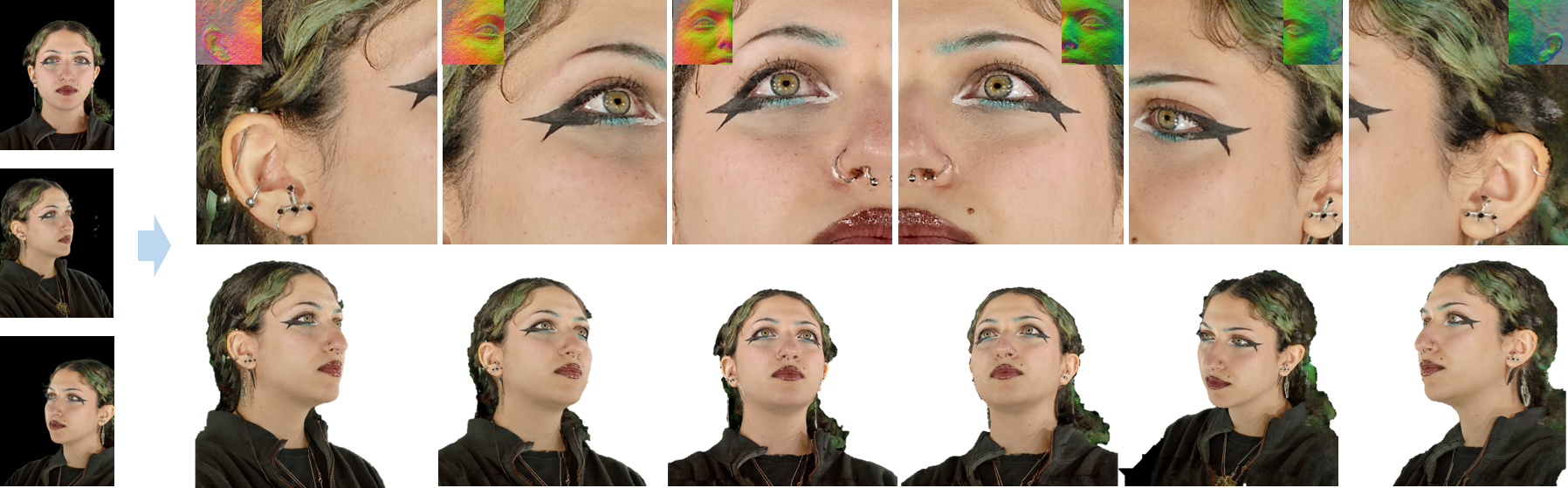}}
    \captionof{figure}{
    We propose a method for synthesising novel views of faces at ultra high-resolution from very sparse inputs. This figure shows novel view renderings at \textbf{4K resolution} reconstructed from \textbf{only three views} of the target identity.}
    \label{fig:teaser}
\end{center}
}]

%% file: 01_intro.tex
\begin{figure*}[t]
    \centering
  \includegraphics[width=\textwidth]
              {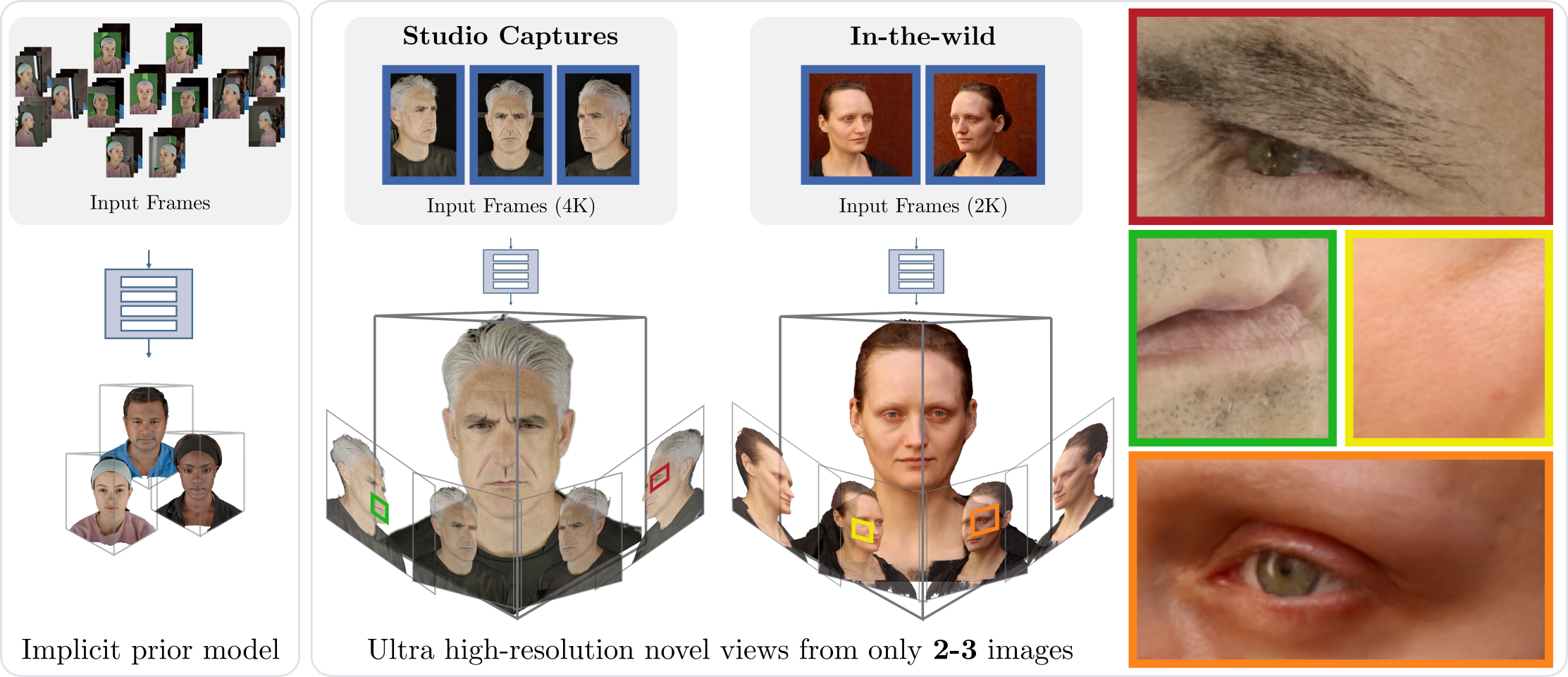}
    \caption{Our key contribution is a prior face model (left), learned from a multiview dataset of faces captured in a controlled setting. The prior model is resolution independent and can be fine-tuned to synthesise novel views at high resolution given as few as two images from a target identity captured in the studio (middle left) or in-the-wild (middle right).
    \label{fig:method_teaser}}
\end{figure*}

\section{Introduction}
Reconstruction and novel view synthesis of faces are challenging problems in 3D computer vision. Achieving high-quality photorealistic synthesis is difficult due to the underlying complex geometry and light transport effects exhibited by organic surfaces. Traditional techniques use explicit geometry and appearance representations for modeling individual face parts such as hair \cite{hair}, skin \cite{gotardo2018}, eyes \cite{eyes}, teeth \cite{wu2016teeth} and lips \cite{garrido2016lips}. Such methods often require specialised expertise and hardware and limit the applications to professional use cases. 

Recent advances in volumetric modelling \cite{nv,nerf,plenoxels,mipnerf360} have enabled learned, photorealistic view synthesis of both general scenes and specific object categories such as faces from 2D images alone. Such approaches are particularly well-suited to model challenging effects such as hair strands and skin reflectance. 
The higher dimensionality of the volumetric reconstruction problem is inherently more ambiguous than surface-based methods. Thus, initial developments in neural volumetric rendering methods \cite{nerf, mipnerf360} relied on an order-of-magnitude higher number of input images ($>100$) to make the solution tractable. Such a large image acquisition cost limits application to wider casual consumer use cases. Hence, few-shot volumetric reconstruction, of both general scenes and specific object categories such as human faces, remains a prized open problem.

This problem of the inherent ambiguity of volumetric neural reconstruction from few images has generally been approached in 3 ways:
i) Regularisation: using natural statistics to constrain the density field better such as low entropy ~\cite{mipnerf360,lolnerf} along camera rays, 3D spatial smoothness ~\cite{regnerf} and deep surfaces \cite{nerfpp} to avoid degenerate solutions such as floating artifacts; 
ii) initialisation: meta-learnt initialisation ~\cite{tancik2021learned} of the underlying representation (network weights) to aid faster and more accurate convergence during optimisation;
iii) data-driven subspace priors: using large and diverse datasets to learn generative ~\cite{pigan,eg3d,gram,gu2021stylenerf,deng2022gram,chen2021sofgan,zhou2021CIPS3D} or reconstructive ~\cite{lolnerf,h3dnet,cao2022authentic,morf} priors of the scene volume.

For human faces, large in-the-wild datasets ~\cite{ffhq,celeba,karras2017progressive} have proved to be particularly attractive in learning a smooth, diverse, and differentiable subspace that allow for few-shot reconstruction of novel subjects by performing inversion and finetuning of the model on a small set of images of the target identity ~\cite{roich2021pivotal}. But such general datasets and generative models also suffer from disadvantages:
i) The sharp distribution of frontal head poses in these datasets prevents generalisation to more extreme camera views, and ii) the computational challenge of training a 3D volume on such large datasets results in very limited output resolutions.

In this paper, we propose a novel volumetric prior for faces that is learned from a multi-view dataset of diverse human faces. 
Our model consists of a neural radiance field (NeRF) conditioned on learnt per-identity embeddings trained to generate 3D consistent views from the dataset. We perform a pre-processing step that aligns the geometry of the captured subjects~\cite{lolnerf}. 
This geometric alignment of the training identities allows our prior model to learn a continuous latent space using only image reconstruction losses. At test time, we perform model inversion to compute the embedding for a novel target identity from the given small set of views of arbitrary high resolution. 
In an out-of-model finetuning step, the resulting embedding and model are further trained with the given images. This results in NeRF model of the target subject that can synthesise high-quality images. Without our prior, the model cannot estimate a 3D consistent volume and overfits to the sparse training views (Fig.~\ref{fig:ablation_prioreffect}).
\input{figures/ablation/ablation_prioreffect}

While we present a novel data-driven subspace prior, we also extensively evaluate the role of regularisation and initialisation in achieving plausible 3D face volumes from few images by comparing with relevant state-of-the-art techniques and performing design ablations of our method.

In summary, we contribute:
\vspace{-0.1in}
\begin{itemize}
    \item A prior model for faces that can be finetuned to generate a high-quality volumetric 3D representation of a target identity from two or more views.
    \item Ultra high-resolution 3D consistent view-synthesis (demonstrated up to 4k resolution).
    \item Generalisation to in-the-wild indoor and outdoor captures, including challenging lighting conditions.
\end{itemize}

%% file: figures/ablation/ablation_prioreffect.tex
\begin{figure}

\begin{center}
\small
\setlength{\tabcolsep}{2pt}

\newcommand{\height}{1.7cm}
\newcommand{\inputheight}{1.2cm}
\begin{tabular}{cccccc}
\multirow{2}{*}{
\includegraphics[height=\inputheight]{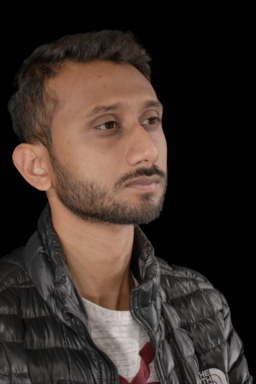} \includegraphics[height=\inputheight]{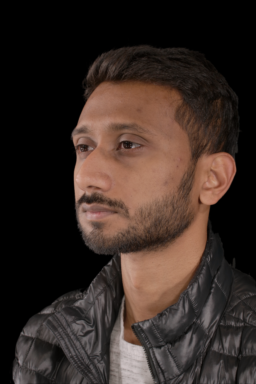}
} &
\rotatebox{90}{W/o Prior} & 
\includegraphics[height=\height]{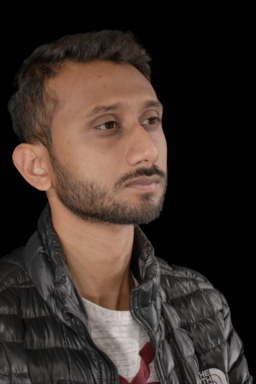} 
\includegraphics[height=\height]{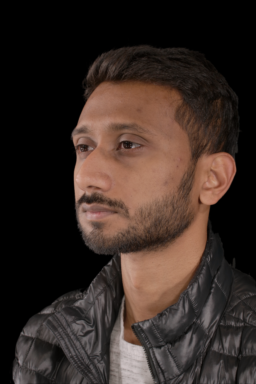} && \includegraphics[height=\height]{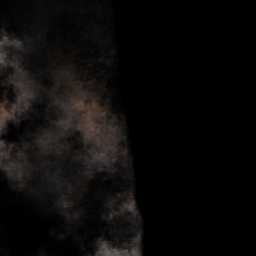}  \includegraphics[height=\height]{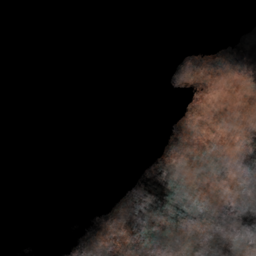} 
\\
&
 \rotatebox{90}{With Prior} & \includegraphics[height=\height]{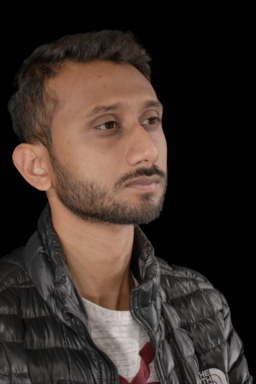} 
\includegraphics[height=\height]{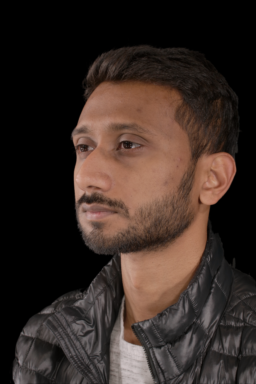} & &
 \includegraphics[height=\height]{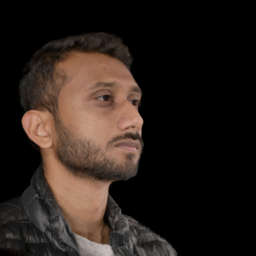}  \includegraphics[height=\height]{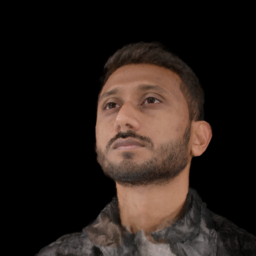} 
 \\
 Input &  & Training Result & & Novel Views \\
 \end{tabular}
\end{center}
\caption{\label{fig:ablation_prioreffect} Naively training on two images leads to overfitting and the model fails to synthesise novel views. With the proposed prior, the model can render view-consistent novel views.}

\end{figure}

%% file: 02_related.tex
\begin{figure*}[ht]
    \centering
  \includegraphics[width=\textwidth]
                  {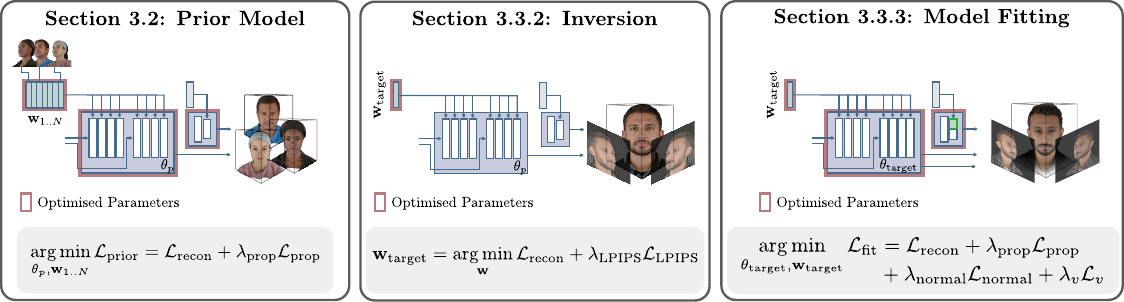}
    \caption{Overview. We train an implicit prior model on low-resolution multi-view images (left). At test time, we fit the prior model to as few as two images of a target identity. 
    A na\"ive optimisation without inversion or regularisation leads to strong view-dependent colour distortions and fuzzy surface structures, see Sec. \ref{sec:ablations} and Fig. \ref{fig:ablation_regularization_short}. To solve this, we first find a good initialisation through  inversion (middle) and then finetune all model parameters under additional constraints for geometry $\mathcal{L}_\text{normal}$ and appearance $\mathcal{L}_v$ (right). 
    \label{fig:overview}}
\end{figure*}

\section{Related works}

Volumetric reconstruction techniques~\cite{mildenhall2020nerf,park2021nerfies,park2021hypernerf,li2022eyenerf,mipnerf360} achieve a high-level of photorealism. However, they provide a wider space of solutions than surface based representations~\cite{park2019deepsdf,mescheder2019occupancy}, and hence often perform very poorly in the absence of sufficient constraints~\cite{regnerf,sparf2023,pixelnerf,keypointnerf,diner,Yang2023FreeNeRF}. To mitigate this, related works employ additional regularisation~\cite{regnerf,Yang2023FreeNeRF,lolnerf,dietnerf,keypointnerf,sparf2023}, perform sophisticated initialisation~\cite{tancik2021learned,h3dnet,vora2021nesf,kundu2022panoptic}, and leverage data-driven priors~\cite{lolnerf,eg3d,gu2021stylenerf,or2021stylesdf,h3dnet,voluxgan,pixelnerf,ibrnet,keypointnerf,chen2021mvsnerf,diner,sparf2023,dietnerf,zhang2022fdnerf}.

\paragraph{Regularisation}
A common solution to novel view synthesis from sparse views is employing regularisation and consistency losses for novel views. 

RegNeRF~\cite{regnerf} proposes a smoothness regulariser on the expected depth and a patch-based appearance regularisation from a pretrained normalising flow. 
A concurrent work, FreeNeRF~\cite{Yang2023FreeNeRF}, observes that NeRFs tend to overfit early in training because of the high frequencies in the positional encoding. They propose a training schedule where the training starts with the positional encodings masked to the low frequencies only and continuously fade in higher frequencies during the course of training. 
These methods have shown promising results for in-the-wild scenes but struggle to output high-quality results for human faces~\ref{fig:comparison_1k_light}.

It is also possible to leverage priors from large pretrained models. DietNeRF\cite{dietnerf} follows a strategy of constraining high-level semantic features of novel view images to map to the same scene object in the ``CLIP" \cite{clip} space. These methods require generating image patches per mini-batch rather than individual pixels. This is compute and memory intensive and reduces the effective batch size and resolution at which the models can be trained, limiting the overall quality.

\paragraph{Initialisation}
Recent papers explore the effect of initialisation~\cite{tancik2021learned,h3dnet,kundu2022panoptic,vora2021nesf}. Metalearning~\cite{finn2017modelmaml,nichol2018first,sitzmann2020metasdf,Zakharov_2019_ICCV} initial model parameters from a large collection of images~\cite{tancik2021learned} has shown promising results for faster convergence. However, the inner update loop in metalearning becomes very expensive for large neural networks. This limits its applicability in high-resolution settings.

\paragraph{Data-driven Priors}

Recent works propose generative neural fields models in 3D~\cite{eg3d,gu2021stylenerf,zhou2021CIPS3D,gram,lolnerf,pigan,niemeyer2021giraffe,schwarz2020graf,xiang2022gram,voluxgan,prao2022vorf}. These models typically map a random latent vector to a radiance field. At inference time, the model can generate novel views by inverting a target image to the latent space~\cite{abdal2019image2stylegan}.

GRAF and PiGAN \cite{schwarz2020graf,pigan} are the first technique to learn a 3D volumetric generative model trained with an adversarial loss on in-the-wild datasets. 
Since neural radiance fields are computationally expensive, training them in an adversarial setting requires an efficient representation.
EG3D \cite{eg3d} proposes a tri-plane representation, which enables training lightweight neural radiance field as a 3D GANs, resulting in state-of-the-art synthesis results. 

Due to memory limitations, such generative models can be trained only at limited resolutions. They commonly rely on an additional 2D super-resolution module to generate more details~\cite{eg3d,gu2021stylenerf,voluxgan,pigan}, which results in the loss of 3D consistency.

Recent works render 3D consistent views by avoiding a 2D super-resolution module~\cite{morf,cao2022authentic}.
MoRF \cite{morf} learns a conditional NeRF~\cite{mildenhall2020nerf} for human heads from multiview images captured using a polarisation based studio setup that helps to learn separate diffuse and specular image components. Their dataset consists of 15 real identities and is supplemented with synthetic renderings to generate more views. Their method is limited to generating results in the studio setting and does not generalise to in-the-wild scenes. 
Cao et al. 2022~\cite{cao2022authentic} train a universal avatar prior that can be finetuned to a target subject with a short mobile phone capture of RGB and depth. Their underlying representation follows Lombardi et al.~\cite{lombardi2021mvp}.

A popular option for novel view synthesis from sparse inputs is formulating the task as an auto-encoder and perform image-based rendering. This family of methods \cite{pixelnerf,ibrnet,keypointnerf,chen2021mvsnerf} follow a feedforward approach of generalisation to novel scenes by training a convolutional encoder that maps input images to pixel aligned features that condition a volumetric representation of the scene. 

Multiple works extend this approach with additional priors including keypoints~\cite{keypointnerf}, depth maps~\cite{diner,xu2022sinnerf,wang2023sparsenerf}, or correspondences~\cite{sparf2023}.
KeypointNeRF~\cite{keypointnerf} employs an adapted positional encoding strategy based on 3D keypoints. DINER~\cite{diner} includes depth maps estimated from pretrained models to bootstrap the learning of density field and sample the volume more efficiently around the expected depth value. Employing our face prior outperforms these methods (see Tbl.~\ref{tab:comp_holobooth_short}, Fig. \ref{fig:comparison_1k_light} and \ref{fig:comparison_facescape_light}).

%% file: 04_method.tex
\section{Method}
We propose a prior model for faces that can be finetuned to very sparse views. The finetuned model can generate ultra-high resolution novel view synthesis with intricate details like individual hair strands, eyelashes, and skin pores (Fig.~\ref{fig:teaser}). In this section, we first introduce neural radiance fields~\cite{mildenhall2020nerf} in Sec.~\ref{sec:background} and our prior model in Sec.~\ref{sec:prior}. We then outline our reconstruction pipeline in Sec.~\ref{sec:reconstruction}.

\subsection{Background}
\label{sec:background}
A NeRF~\cite{nerf} represents a scene as a volumetric function $f : (\textbf{x}, \textbf{d}) \rightarrow (\textbf{c}, \sigma) $ which maps 3D locations $\textbf{x}$ to a radiance $\textbf{c}$ and a density $\sigma$, which is modelled using a multi-layer perceptron (MLP). The radiance is additionally conditioned on the view direction $\textbf{d}$ to support view dependent effects such as specularity. In order to more effectively represent and learn high frequency effects, each location is positionally encoded before being passed to the MLP.

Given a NeRF, a pixel can be rendered by integrating along its corresponding camera ray in order to obtain the radiance or colour value $\bf \hat c = \textbf{F}(\textbf{r})$. Assuming a predetermined near and far camera plane ${t_n}$ and ${t_f}$, the integrated radiance of the camera ray can be computed using the following equation: \vspace{-2em}

\begin{align}
    \textbf{F}(\textbf{r}) = \int_{t_n}^{t_f}T(t)\sigma(\textbf{r}(t))\textbf{c}(\textbf{r}(t),\textbf{d}) dt \text{,} \\
    \text{ where } T(t) = \text{exp} \left( - \int_{t_n}^{t} \sigma(\textbf{r}(s)) ds \right).
\end{align}

In practice, this is estimated using raymarching. The original NeRF implementation approximated the ray into a discrete number of sample points, and estimated the alpha value of each sample by multiplying its density with the distance to the next sample. They further improve quality using a coarse-to-fine rendering method, by first distributing samples uniformly between the near and far planes, and then importance sampling the quadrature weights.

Mip-NeRF~\cite{barron2021mip} solves the classic anti-aliasing problem resulting from discrete sampling in a continuous space. This is achieved by sampling conical volumes along the ray. MipNeRF360~\cite{mipnerf360} also introduced an efficient pre-rendering step; a uniformly sampled coarse rendering pass by a proposal network, which predicts the sampling weights instead of the density and colour values using a lightweight MLP. This is followed by an importance-sampled NeRF rendering step. We incorporate both of these ideas in our model. 

\subsection{Face Prior Model}
\label{sec:prior}
Our prior model is a conditional neural radiance field $F_\theta$ that is trained as an auto-decoder ~\cite{bojanowski2018optimizing,lolnerf}.
Given a ray $\bf r$ and a latent code $\bf w$, $F_\theta$ predicts a colour $\bf{\hat c} = F_\theta (\bf r, \bf w)$ with volumetric rendering~\cite{mildenhall2020nerf}.

The architecture of the prior model is based on Mip-NeRF360~\cite{mipnerf360} and consists of two MLPs. Unlike Mip-NeRF360, the MLPs are conditioned on a latent code $\textbf{w}_\text{identity}$, representing the identity.

The first MLP---the \emph{proposal} network---predicts density only. The second MLP---the \emph{NeRF} MLP---predicts both density and colour. Both MLPs take an encoded point $\tilde \gamma_{\bf x} (\bf x)$ and a latent code $\bf w$ as input, where $\tilde \gamma_{\bf x}(\cdot)$ denotes a function for integrated positional encodings~\cite{barron2021mip}. The NeRF MLP further takes the positionally encoded view direction $\gamma_{\bf v}(\bf d)$ as input (without integration for the positional encoding).

\begin{figure}[t]
    \centering
  \includegraphics[width=0.35\textwidth]
                  {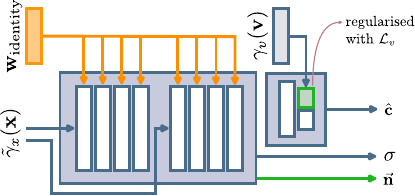}
    \caption{Prior Model Architecture. Our prior model extends the Mip-NeRF360~\cite{mipnerf360} architecture with a conditioning input at each layer of the trunk MLP. Unlike SOTA generative NeRF models \cite{eg3d,gu2021stylenerf,lolnerf}, our model conditions both on a latent code \emph{and} a view direction, which enables view-dependent effects. During model fitting to very few images, we prevent overfitting by regularising the view direction weights. See Fig.~\ref{fig:ablation_regularization_short} for an example.
    \label{fig:prior}}
\end{figure}

Fig.~\ref{fig:prior} gives an overview of the backbone NeRF MLP of our prior model. The latent code is concatenated at each layer. Unlike state-of-the-art generative models~\cite{lolnerf,chan2022efficient,gu2021stylenerf}, our model also conditions on the view direction $\bf d$.

For training, we sample random rays $\bf r$ and render the output colour $\bf \hat c$ as described in Sec.~\ref{sec:background}. Given $N$ training subjects, we optimise over both the network parameters $\bf \theta$ and the latent codes $\mathbf{w}_{1..N}$.
Our objective function is
\begin{align}
\label{eq:priorloss}
    \argmin_{\bf \theta, \textbf{w}_\text{1..N}} \mathcal{L}_\text{prior} = \mathcal{L}_\text{recon} + \lambda_\text{prop} \mathcal{L}_\text{prop},
\end{align}

with $\lambda_\text{prop}=1$.
We describe the loss terms $\mathcal{L}_\text{recon}$ and $\mathcal{L}_\text{prop}$ for a single ray. The final loss is computed as the expectation over all rays in the training batch.

The objective function has a data term comparing the predicted colour with the ground truth $\mathcal{L}_\text{recon} = \lVert{\bf F_\theta(\bf r, \bf w) - \bf c}\rVert_1$, as well as a weight distribution matching loss term between the NeRF MLP and the proposal MLP $\mathcal{L}_\text{prop}$. The latter is the same as in Mip-NeRF360~\cite{mipnerf360}. We refrain from regularising the latent space, and we disable the distortion loss. As our scene is not unbounded, we also disable the 360-parameterisation or space-warping of MipNeRF360. 

We train the prior model for 1 Mio. steps on multi-view images of resolution $512\times768$. Please refer to Sec.~\ref{sec:dataset} for details about the training set.

\subsection{Volumetric Reconstruction Pipeline}
\label{sec:reconstruction}
Figure~\ref{fig:overview} illustrates the reconstruction pipeline, which comprises three steps: 1) Preprocessing and head alignment, 2) inversion, and 3) model fitting. This section describes each step in detail.

\subsubsection{Preprocessing}
\label{sec:preprocessing}
We estimate camera parameters and align the heads to a predefined canonical pose during the data preprocessing stage.
For the studio setting, we calibrate the cameras and estimate 3D keypoints by triangulating detected 2D keypoints; for in-the-wild captures, we use Mediapipe~\cite{lugaresi2019mediapipe} to estimate the camera positions and 3D keypoints.
We align and compute a similarity transform to a predefined set of five 3D keypoints (outer eye corners, nose, mouth centre, and the chin) in a canonical pose. Please see the supp. mat. for details.

\subsubsection{Inversion}
\label{sec:inversion}
The reconstruction results depend on a good initialisation of the face geometry (see Tbl.~\ref{tab:ablation_initialization_summary}). We solve an optimisation problem to find a latent code that produces a good starting point~\cite{abdal2019image2stylegan}.

Given $K$ views of a target identity, we optimise with respect to a new latent code while keeping the network weights frozen.
Let $P$ be a random patch sampled from one of the $K$ images of the target identity and $\hat P_{\bf w}$ be a patch rendered by our prior model when conditioning on the latent code $\bf w$.
The latent code of the target identity $\mathbf{w}_\text{target}$ is recovered by minimising the following objective function:

\begin{equation}
    \mathbf{w}_\text{target} = \argmin_{\mathbf{w}} \mathcal{L}_\text{recon} + \lambda_\text{LPIPS} \mathcal{L}_\text{LPIPS},
\end{equation}
where $\mathcal{L}_\text{recon} =\frac{1}{\lvert P \rvert}{\lVert \hat P_{\bf w} - P \rVert}$ is the same loss as in Eq.~\ref{eq:priorloss}, but computed over an image patch, and $\mathcal{L}_\text{LPIPS}(\hat P_{\bf w}, P)$ is a perceptual loss\cite{zhang2018perceptual} with $\lambda_\text{LPIPS}=0.2$.
We optimise at the same resolution as the prior model after removing the background~\cite{pandey2021total}.

\subsubsection{Model Fitting}
The goal of model fitting is to adapt the weights of the prior model for generating novel views of a target identity at high resolutions. We do this by finetuning the weights of the prior model to a target identity from sparse views.

Please note that the prior model is trained on \emph{low resolution} and is optimised to reconstruct a \emph{large set of identities} from \emph{many views} for each identity, see Sec.~\ref{sec:priormodeltraining}. After model fitting, the model should generate \emph{high-resolution novel views} with intricate details like individual hair strands for a \emph{single} target identity given as few as \emph{two} views. 

Training a NeRF model on sparse views leads to major artifacts because of a distorted geometry~\cite{niemeyer2022regnerf} and overfitting to high frequencies~\cite{Yang2023FreeNeRF}. 
We find that correctly initialising the weights of the model avoids floater artifacts and leads to high-quality novel view synthesis.
We initialise the model weights with the pretrained prior model and use the latent code $\mathbf{w}_{\text{target}}$ obtained through inversion (Sec.~\ref{sec:inversion}).
Fig.~\ref{fig:ablation_regularization_short} shows that na\"ively optimising without any further constraints leads to overfitting to the view direction (first column). Regularising the weights of the view branch causes fuzzy surface structures (second column), which can be mitigated using a normal consistency loss~\cite{verbin2022ref} (third column).
We initialise the model with the weights of the prior and optimise it given the objective function

\begin{align}
\label{eq:modelfittingloss}
    \argmin_{\mathbf{\theta}_{\text{target}}, \mathbf{w}_{\text{target}}} \mathcal{L}_\text{fit} &= \mathcal{L}_\text{recon} + \lambda_\text{prop} \mathcal{L}_\text{prop} \notag\\
    &+ \lambda_\text{normal} \mathcal{L}_\text{normal} + \lambda_{v} \mathcal{L}_{v},
\end{align}

where the loss terms $\mathcal{L}_\text{recon}$, $\mathcal{L}_\text{prop}$ and the hyperparameter $\lambda_\text{prop}$ are the same as in Eq.~\ref{eq:priorloss}.
The regulariser for the normals $\mathcal{L}_\text{normal}$ is the same as in RefNeRF~\cite{verbin2022ref}. We regularise the weights of the view branch with $\mathcal{L}_{v} = \lVert \theta_v \rVert^2$, where the parameters $\theta_v$ correspond to weights of the connections between the encoded view direction and the output, see the highlighted box in Fig.~\ref{fig:prior}.
We set $\lambda_\text{normal} = 0.001$ and $\lambda_{v} = 0.0001$ and optimise until convergence.

Since our model generates faces that are aligned to a canonical pose and location (Sec.~\ref{sec:exp_preprocessing}), the rendering volume can be bounded by a rectangular box. We set the density outside this box to zero for the final rendering.

%% file: 03_dataset.tex
\section{Dataset}
\begin{figure}
    \centering
  \includegraphics[width=0.9\linewidth]
                  {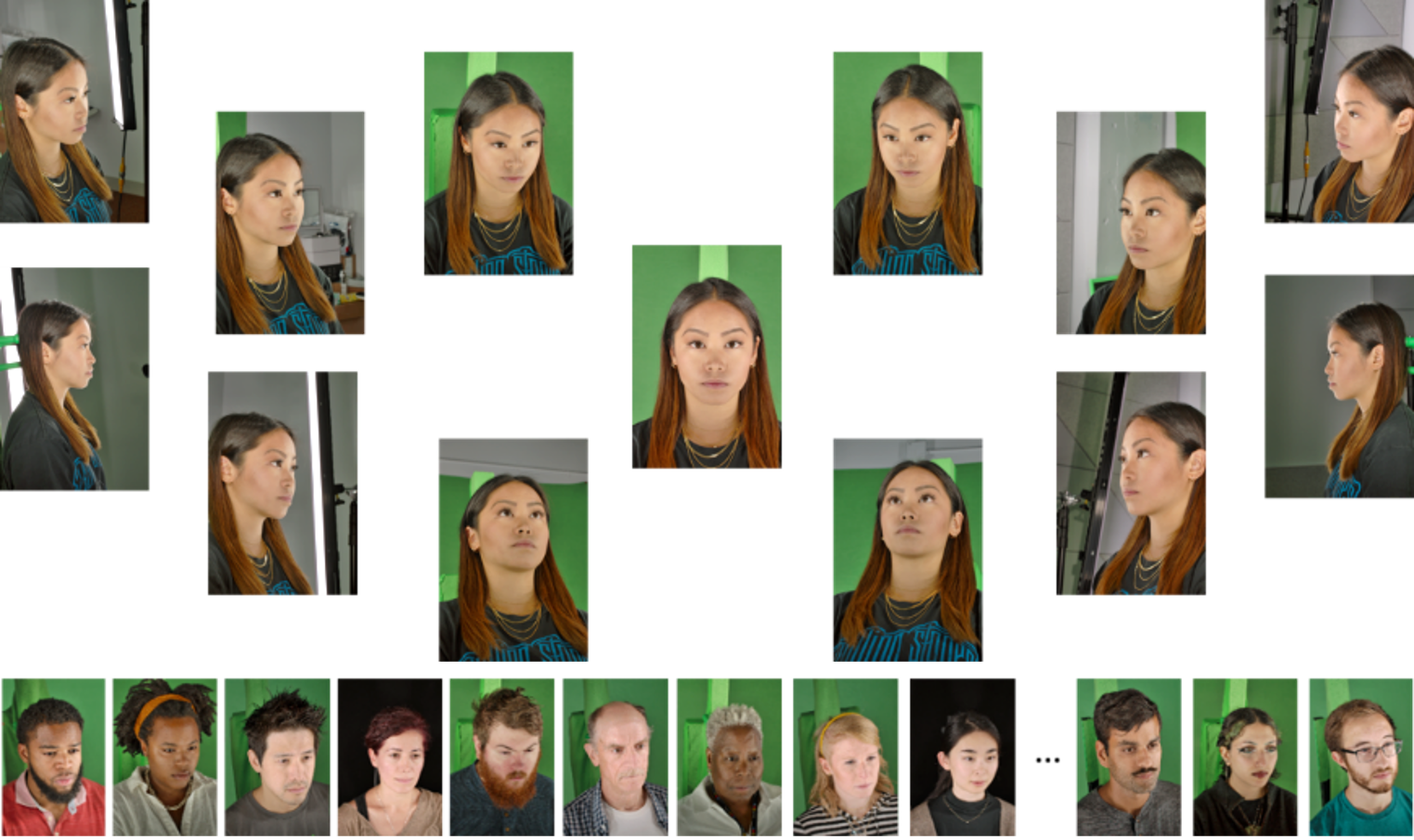}
    \caption{
    Exemplar images of our captured dataset. Our dataset contains 1450 different subjects (\textit{bottom row}) captured under 13 different cameras on the frontal hemisphere (\textit{top rows}).
    \label{fig:dataset}\vspace{-1.5em}}
\end{figure}

We capture a novel high-quality multi-view dataset of diverse human faces from 1450 identities with a neutral facial expression under uniform illumination, see Fig. \ref{fig:dataset}. 13 camera views are distributed uniformly across the frontal hemisphere. Camera calibration is performed prior to every take to obtain accurate camera poses. We hold out 15 identities for evaluation and train on the rest. The camera images are of $4096\times 6144$ resolution.
We made a concerted effort for a diverse representation of different demographic categories in our dataset, but acknowledge the logistical challenges in achieving an entirely equitable distribution. We provide more details of the demographic breakdown of the dataset in the supplementary document. 

To assess the out-of-distribution performance of our method we show results on the publicly available Facescape multi-view dataset~\cite{facescape}. We also acquire a handful of in-the-wild captures of subjects using a mobile camera to qualitatively demonstrate the generalisation capability of our method further. 
\label{sec:dataset}

%% file: 05_experiments.tex
\section{Experiments}

\paragraph{Preprocessing}
\label{sec:exp_preprocessing}
We perform an offline head alignment to a canonical location and pose. This step is crucial to learn a meaningful prior over human faces. For each subject, we estimate five 3D keypoints for the eyes, nose, mouth, and chin and align the head to a canonical location and orientation. The canonical location is defined as the median location of the five keypoints across the first 260 identities of our training set. For an illustration and more details, please see the supplementary document. 

\paragraph{Prior Model Training}
\label{sec:priormodeltraining}
We train the prior model with our pre-processed dataset containing 1450 identities and 13 camera views. 
To make our training computationally tractable, we train versions of our prior model at a lower resolution. We train two versions of our model, at 256$\times$384 and 512$\times$768 image resolution. The lower resolution model is trained only for the purpose of quantitative evaluation against other SOTA methods, to ensure fair comparison against other methods that cannot be trained at a higher resolution due to compute and memory limitations. We provide details about our training hardware and hyperparameters in the supplementary document. 

\paragraph{Comparisons}
We perform evaluations on three different datasets: Our high-quality studio dataset, a publicly available studio dataset (Facescape~\cite{facescape}), and in-the-wild captures from a mobile and a cellphone camera. For the studio datasets, we assume calibrated cameras. For the in-the-wild captures, we estimate camera parameters with Mediapipe~\cite{lugaresi2019mediapipe}.
The metrics for the quantitative comparisons are computed after cropping the images to squares and setting the background to black with foreground masks from a state-of-the-art network~\cite{pandey2021total}. For more details, please refer to the supplementary material.

%% file: 06_results.tex
\section{Results}
\input{tables/comparison_holobooth_views_light}

\input{figures/inthewild/inthewild_light}
\input{figures/comparison/comparison_holobooth_light}
\input{figures/comparison/comparison_facescape_light}

\input{figures/single_image/single_image}

\input{figures/ablation/ablation_regularization_short}
\input{tables/ablation_initialization_summary}
We perform extensive evaluation and experiments to demonstrate i) our core claims - high resolution, few shot, in-the-wild synthesis, ii) improved performance over the state-of-the-art methods, iii) ablation of various design choices. We also encourage the reader to see the video results and more insightful evaluations in the supp. mat.

\subsection{Ultra-high Resolution Synthesis}
We demonstrate ultra high resolution synthesis after finetuning our 512$\times$768 prior model to sparse high-resolution images in the studio setting (Fig.~\ref{fig:teaser}) and in-the-wild (Fig.~\ref{fig:inthewild_light}). 

\paragraph{4K Novel Views from Three Views}
Figure~\ref{fig:teaser} shows $4096 \times 4096$ (4K) renderings after finetuning to three views of a held-out subject from our studio dataset. Note the range of the rendered novel views and the quality of synthesis results for such an out-of-distribution test subject at 4K resolution. From just three images, our method learns a highly detailed and photorealistic volumetric model of a face. We synthesise smooth and 3D consistent camera trajectories while preserving challenging details such as individual hair strands, skin pores and eyelashes.
Our model learns both consistent geometry and fine details of individual hair strands and microgeometry of the skin, making the synthesised images barely distinguishable from captured views. Please see the supplementary material for video results and results on other subjects.

\paragraph{2K Novel Views from Two in-the-wild Views}
Our method also affords reconstruction from in-the-wild captures from a single camera. We use a digital camera to capture two images. Results are shown in Fig.~\ref{fig:inthewild_light}. The upper row was captured outdoors in front of a wall; the bottom was row was captured in a room. Please see the supplementary material for more examples and videos.

\subsection{Comparison with Related Work}
Our goal is high-resolution novel view synthesis from sparse inputs. 
We perform comparisons by training related works~\cite{tancik2021learned,keypointnerf,regnerf,Yang2023FreeNeRF,eg3d} on our studio dataset and rendering results for unseen views at resolution $1024\times 1024$ (1K). Since the task of novel view synthesis becomes substantially easier when given more views of the target subject, we perform comparisons for different number of views ranging from two to seven. Fig.~\ref{fig:comparison_1k_light} and Tbl.~\ref{tab:comp_holobooth_short} show that our method can handle difficult cases at high resolution and clearly outperforms all related works when reconstructing from two views. Please see the supp. mat. for results on more views.

We observe that some of the related methods perform significantly better at lower resolutions and when given more than just two views of the target subject. Hence, we complement our comparisons with a comparison on the FaceScape dataset~\cite{facescape}. We follow the setting of the best performing related work, DINER~\cite{diner}, and use four reference images at resolution $256\times 256$. 
 Fig.~\ref{fig:comparison_facescape_light} displays visuals and the supplementary document provides metrics.
Note that KeypointNerf~\cite{keypointnerf} and DINER~\cite{diner} were trained on Facescape while ours is not. This means that our scores represent results in the "out-of-distribution" setting.

\subsection{Single Image Fitting}
Our method is also capable of fitting to a single image and still produces detailed results. We show such result on held-out test subjects from our dataset in Fig.~\ref{fig:single_image}. Note the consistent depth and normal maps and photorealistic renderings. This indicates that our model learns a strong prior over head geometry which helps it resolve depth ambiguity to reconstruct a cohesive density field for the head, including challenging regions like hair.

\subsection{Ablations}
\label{sec:ablations}
\paragraph{Initialisation}
The initialisation of the latent code plays a key role in achieving good results. We ablate various initialisation choices such as: i) a zero vector, ii) Gaussian noise, iii) the mean over the training latent codes, iv) the nearest and furthest neighbour in the training set defined by a precomputed embedding~\cite{schroff2015facenet}, and v)
 inversion (Ours).
We finetune the prior model to two views of three holdout identities and report the results in Tbl.~\ref{tab:ablation_initialization_summary}. Inversion performs best in all metrcis.

\paragraph{Regularisation}
We also ablate the choice of regularisation for the model finetuning. 
Fig. \ref{fig:ablation_regularization_short} shows that without any regularisation, the view branch of the model overfits to the view direction from the sparse input signal. We observe that the parameter weights of the view branch become very large and dominate the colour observed from a particular view. To mitigate this, we regularise the L2 norm of the weights using $L_{v}$ (green highlight in Fig.~\ref{fig:prior}). However, the model still overfits by generating a  fuzzy surface that produces highly specular effects from the optimised views but has incorrect geometry. To regularise the geometry, we extend the trunk of our model with a branch predicting normal and supervise it with the analytical normals~\cite{verbin2022ref}. With both regularisation terms, the model can be robustly fit to a target identity from very sparse views. 

\paragraph{Challenging Lighting Conditions}
Our method can generate high-quality novel views even under challenging lighting conditions with shadows and specular reflections, see Fig.~\ref{fig:ablation_lighting}.
\input{figures/ablation/ablation_lighting}

\paragraph{Further Ablations}
We perform further ablations for fitting to a higher number of target views, for different configurations of our prior models, and for frozen latent codes during model finetuning. Please see the supplementary material for results.

%% file: tables/comparison_holobooth_views_light.tex
\begin{table}[t]
\small 
\setlength{\tabcolsep}{3pt}
\begin{center}

\begin{tabular}{r ccc}
\hline
\textbf{Method} & \textbf{PSNR} $\uparrow$ & \textbf{SSIM} $\uparrow$ & \textbf{LPIPS} $\downarrow$ \\
\hline
FreeNeRF \cite{Yang2023FreeNeRF} & 15.02 & 0.6795 & 0.3093 \\
EG3D-based prior \cite{eg3d} & 19.70 & 0.7588 & 0.2897 \\
Learnit \cite{tancik2021learned} & 20.04 & 0.7716 & 0.3299 \\
RegNeRF \cite{niemeyer2022regnerf} & 20.40 & 0.7432 & 0.2858 \\
KeypointNeRF \cite{keypointnerf} & 22.79 & 0.7878 &	0.2713 \\
\hline
\textbf{Ours} & \textbf{25.69} & \textbf{0.8039} & \textbf{0.1905} \\
\end{tabular}				
\end{center}
\caption{Comparison with related works at 1K resolution on \emph{two} views of our studio dataset. The metrics are computed as the average over six views of three holdout subjects. Our method outperforms the related works by a clear margin. For a visual comparison, please refer to Fig. \ref{fig:comparison_1k_light}. The supp. mat. contains metrics and visuals for more input views.
\label{tab:comp_holobooth_short}\vspace{-1.5em}}
\end{table}

%% file: figures/inthewild/inthewild_light.tex
\begin{figure*}
\begin{center}
\setlength\tabcolsep{1pt}
\newcommand{\crop}{0.8cm}
\newcommand{\cropsmall}{0.4cm}
\newcommand{\height}{1.9cm}
\begin{tabular}{cccccc}
\includegraphics[height=\height]{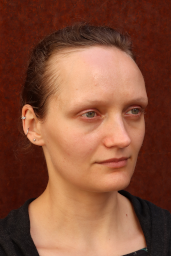}\includegraphics[height=\height]{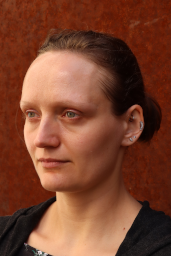}
&\includegraphics[height=\height]{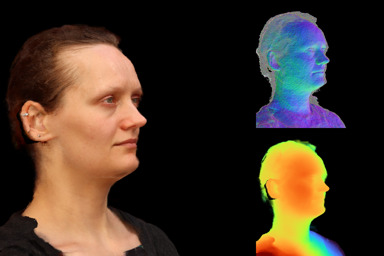}
&\includegraphics[height=\height]{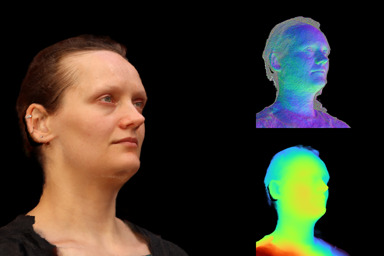}
&\includegraphics[height=\height]{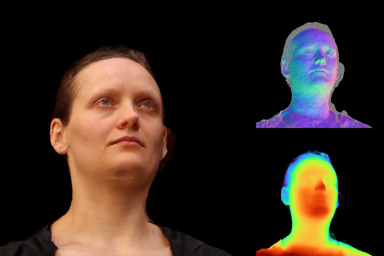}
&\includegraphics[height=\height]{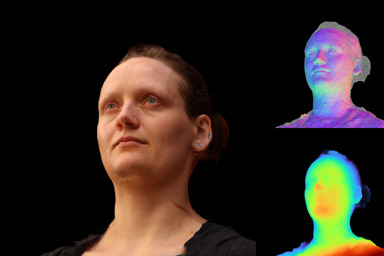}
&\includegraphics[height=\height]{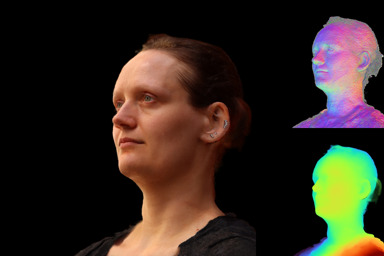}
\\
\includegraphics[height=\height]{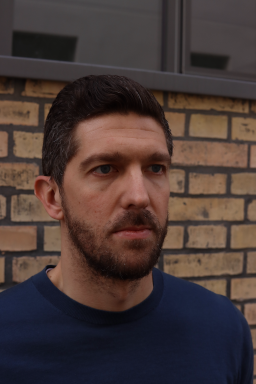}\includegraphics[height=\height]{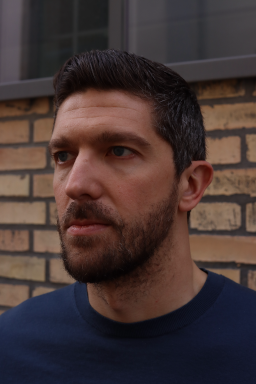}
&\includegraphics[height=\height]{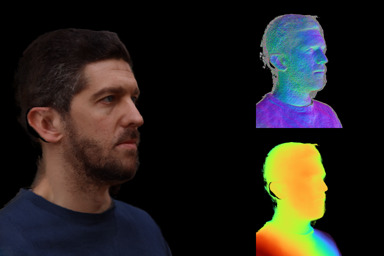}
&\includegraphics[height=\height]{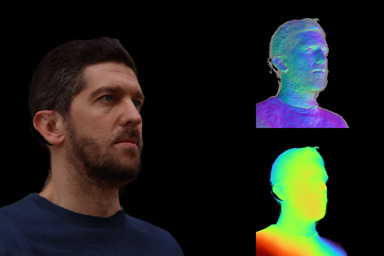}
&\includegraphics[height=\height]{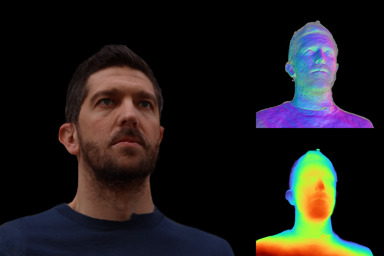}
&\includegraphics[height=\height]{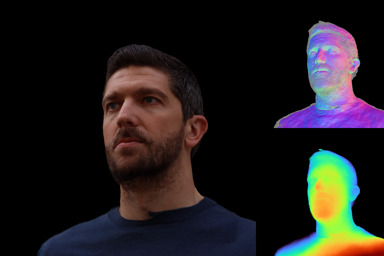}
&\includegraphics[height=\height]{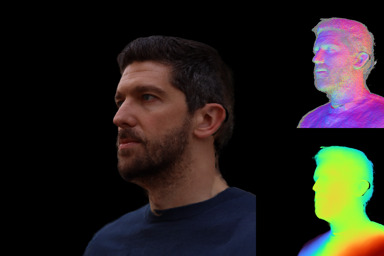}
\\
\end{tabular}
    \caption{In-the-wild Results. We reconstruct a target identity from two images acquired with a consumer camera (left). Note how the novel views can extrapolate from the input camera angles.
    The inlays show the normals (top) and depth (bottom). The hair density is low, thus the grey normal colour in that region. We encourage the reader to see the supp. mat. for the high-resolution results and videos.
    \label{fig:inthewild_light}\vspace{-1.5em}}
\end{center}
\end{figure*}

%% file: figures/comparison/comparison_holobooth_light.tex
\begin{figure*}[ht]

\begin{center}
\small
\setlength{\tabcolsep}{2pt}

\newcommand{\height}{2.3cm}
\begin{tabular}{ccccccc}
  \includegraphics[height=\height]{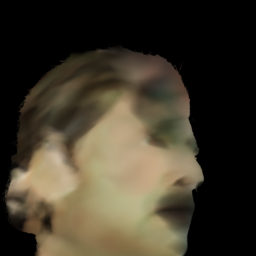}  
  & \includegraphics[height=\height] {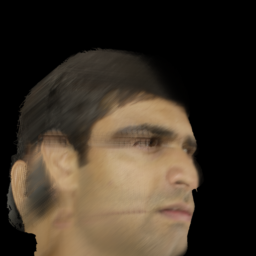}  
  & \includegraphics[height=\height]{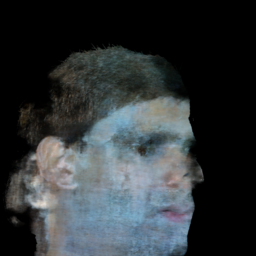} 
  & \includegraphics[height=\height]{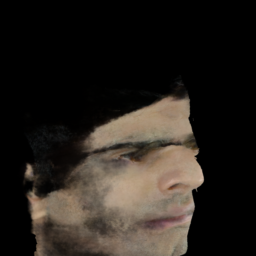}
 & \includegraphics[height=\height]{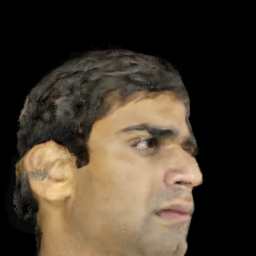}   
 & \includegraphics[height=\height]{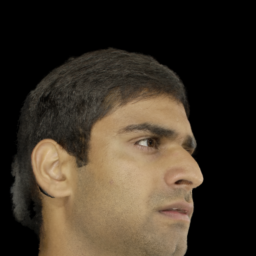}    
 & \includegraphics[height=\height]{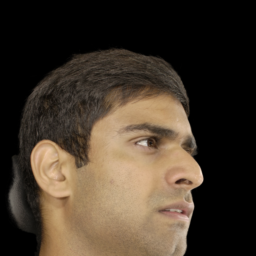} 
    \\
  \includegraphics[height=\height]{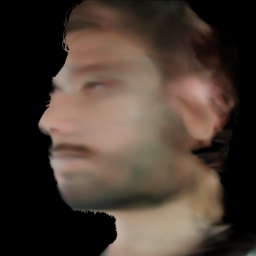}  
  & \includegraphics[height=\height] {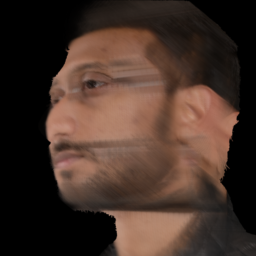}  
  & \includegraphics[height=\height]{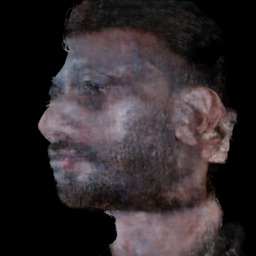} 
  & \includegraphics[height=\height]{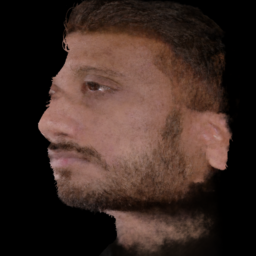}
 & \includegraphics[height=\height]{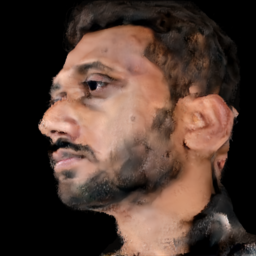}   
 & \includegraphics[height=\height]{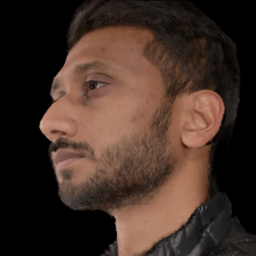}    
 & \includegraphics[height=\height]{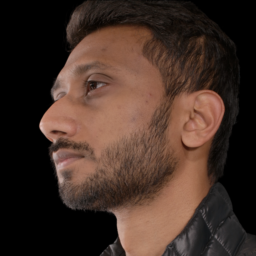} 
    \\
     \includegraphics[height=\height]{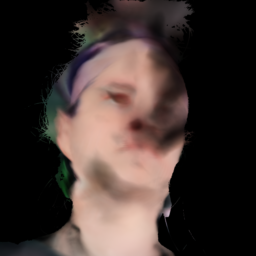}
  & \includegraphics[height=\height] {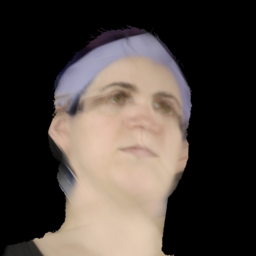}  
  & \includegraphics[height=\height]{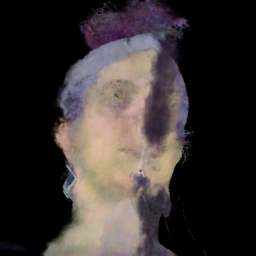} 
  & \includegraphics[height=\height]{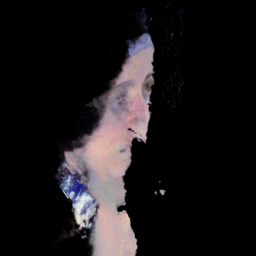}
 & \includegraphics[height=\height]{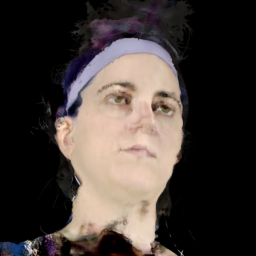}   
 & \includegraphics[height=\height]{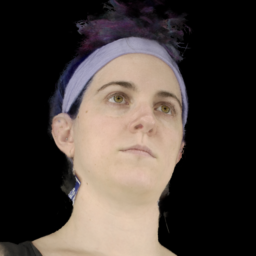}    
 & \includegraphics[height=\height]{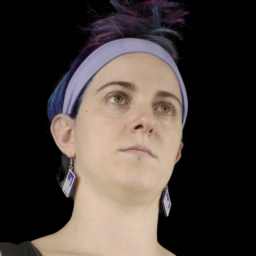} 
    \\ 
Learnit \cite{tancik2021learned} & KeypointNeRF \cite{keypointnerf} & RegNeRF \cite{regnerf} & FreeNeRF \cite{Yang2023FreeNeRF} & EG3D-based prior \cite{eg3d} & \textbf{Ours} & GT
\end{tabular}
\end{center}
\caption{\label{fig:comparison_1k_light} Visual comparison when given two target views. Our method consistently produces more pleasing results. Please see Tbl.~\ref{tab:comp_holobooth_short} for metrics and the supplementary material for implementation details and results on more than two target views.}
\end{figure*}

%% file: figures/comparison/comparison_facescape_light.tex
\begin{figure*}
\begin{center}
\small
\setlength{\tabcolsep}{2pt}
\newcommand{\height}{3.4cm}
\newcommand{\wordsubject}{facescape/sub212_01_target28_ref7-55-40-23}
\newcommand{\eyebrowsubject}{facescape/sub340_01_target52_ref3-45-17-54}
\newcommand{\eyesubject}{facescape/sub122_01_target13_ref52-37-21-44}
\newcommand{\mouthsubject}{facescape/sub344_01_target19_ref34-35-23-24}
\newcommand{\earsubject}{facescape/sub340_01_target29_ref7-50-37-48}

\begin{tabular}{cc|ccccc}
\includegraphics[height=\height]{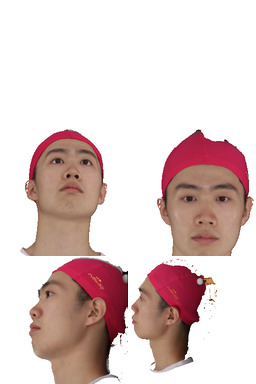} &
\includegraphics[height=\height]{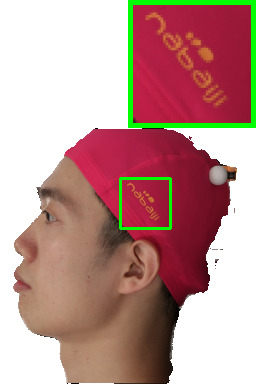} &
\includegraphics[height=\height]{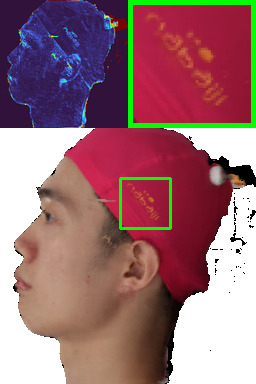} &
\includegraphics[height=\height]{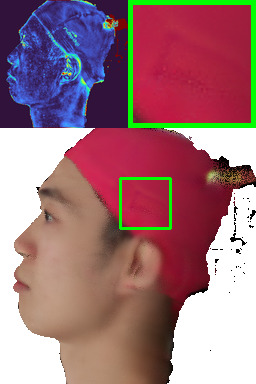} &
\includegraphics[height=\height]{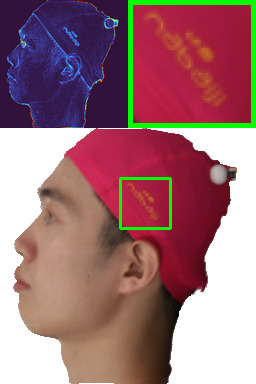} &
\includegraphics[height=\height]{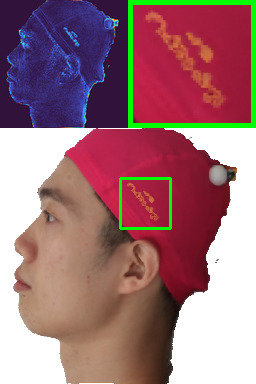} &
\includegraphics[height=\height]{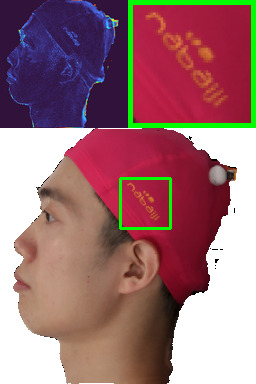} \\
\includegraphics[height=\height]{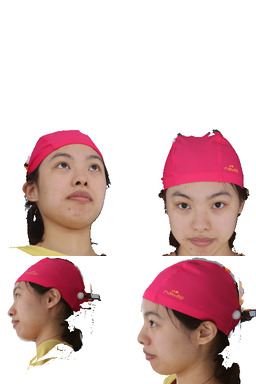} &
\includegraphics[height=\height]{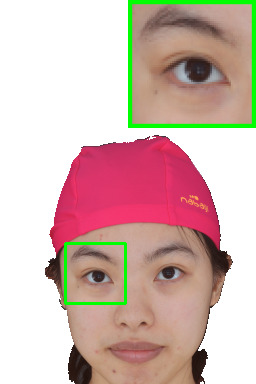} &
\includegraphics[height=\height]{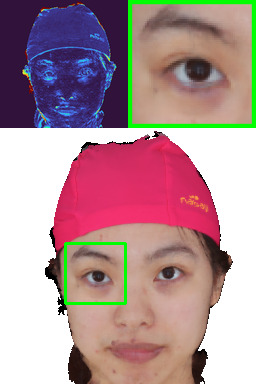} &
\includegraphics[height=\height]{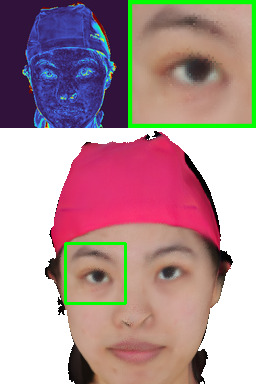} &
\includegraphics[height=\height]{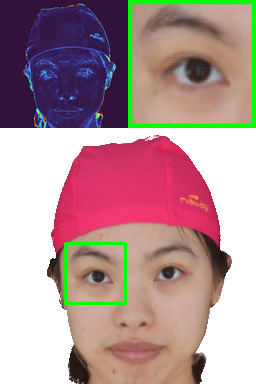} &
\includegraphics[height=\height]{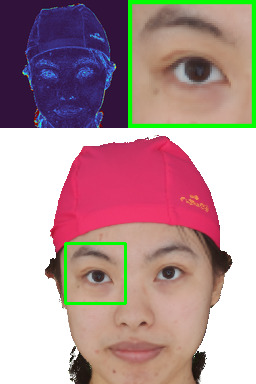} &
\includegraphics[height=\height]{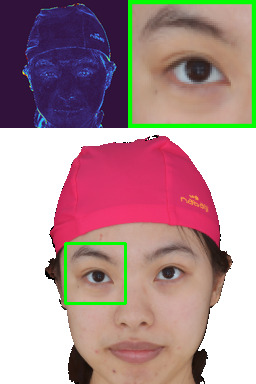} \\
Input & Ground Truth & RegNeRF \cite{regnerf} & EG3D-based prior \cite{eg3d} & KeypointNeRF \cite{keypointnerf} & DINER \cite{diner} & Ours
\end{tabular}
\end{center}
    \caption{Comparison with the state-of-the-art on holdout identities from FaceScape~\cite{facescape}.     
    Each method is given four input views and we show novel views and the L1 residue. Please see the supp. mat. for implementation details, more examples, and detailed metrics.
    \label{fig:comparison_facescape_light}\vspace{-1.5em}}
\end{figure*}

%% file: figures/single_image/single_image.tex
\begin{figure*}
\hspace{30pt}
\begin{center}
\setlength\tabcolsep{1pt}
\newcommand{\crop}{1.0cm}
\newcommand{\cropsmall}{0.4cm}
\newcommand{\height}{2.05cm}
\begin{tabular}{cccccc}
\\
\includegraphics[height=\height]{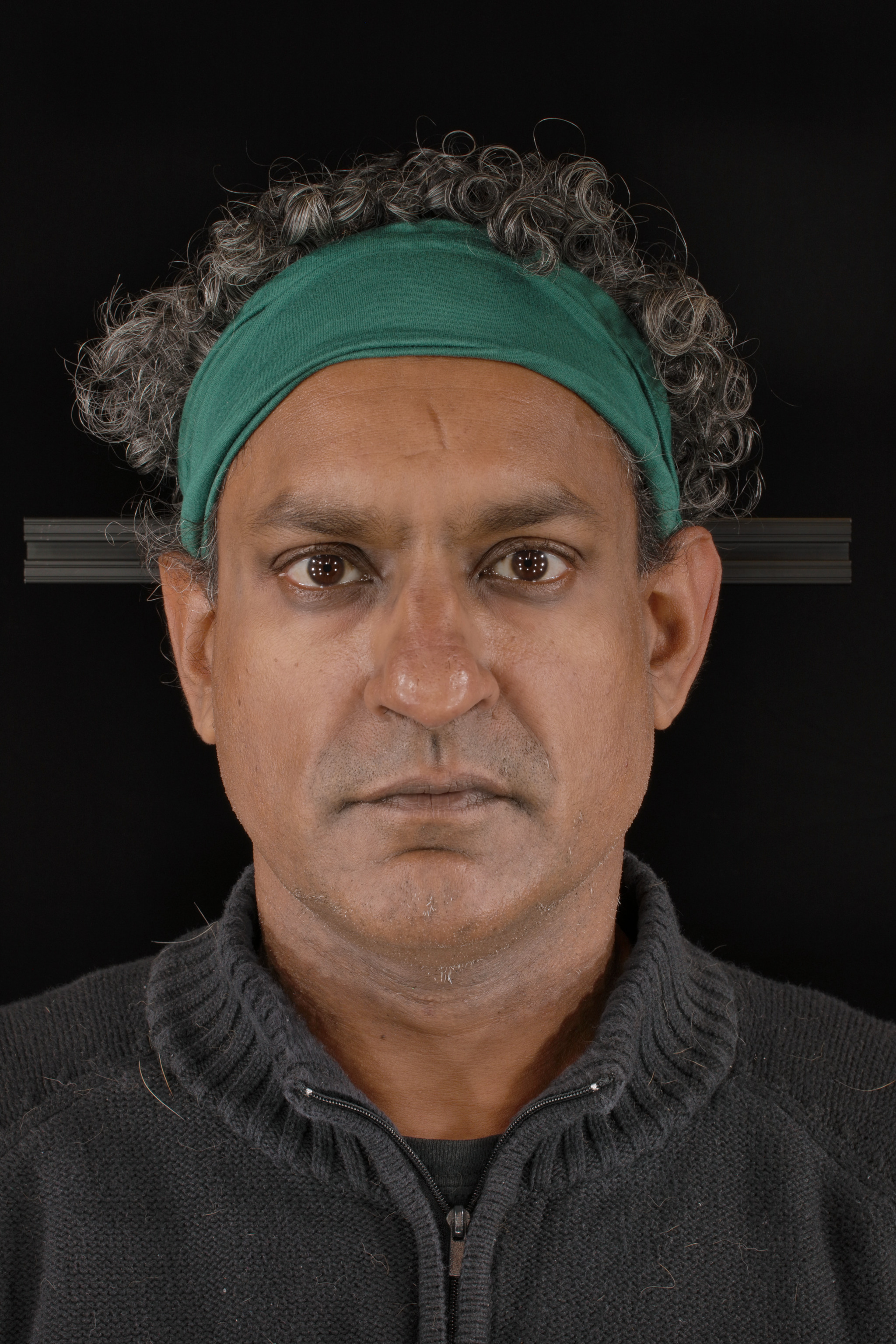}
&\includegraphics[height=\height]{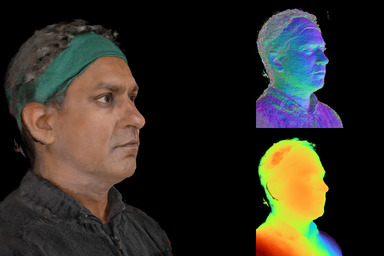}
&\includegraphics[height=\height]{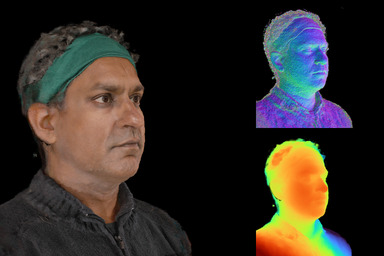}
&\includegraphics[height=\height]{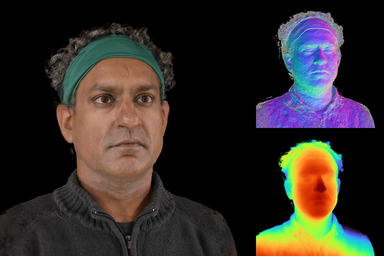}
&\includegraphics[height=\height]{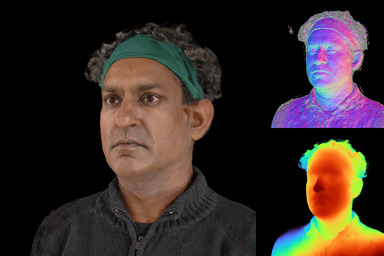}
&\includegraphics[height=\height]{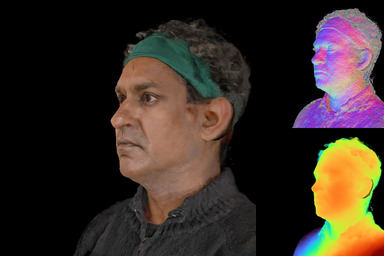}
\\
\includegraphics[height=\height]{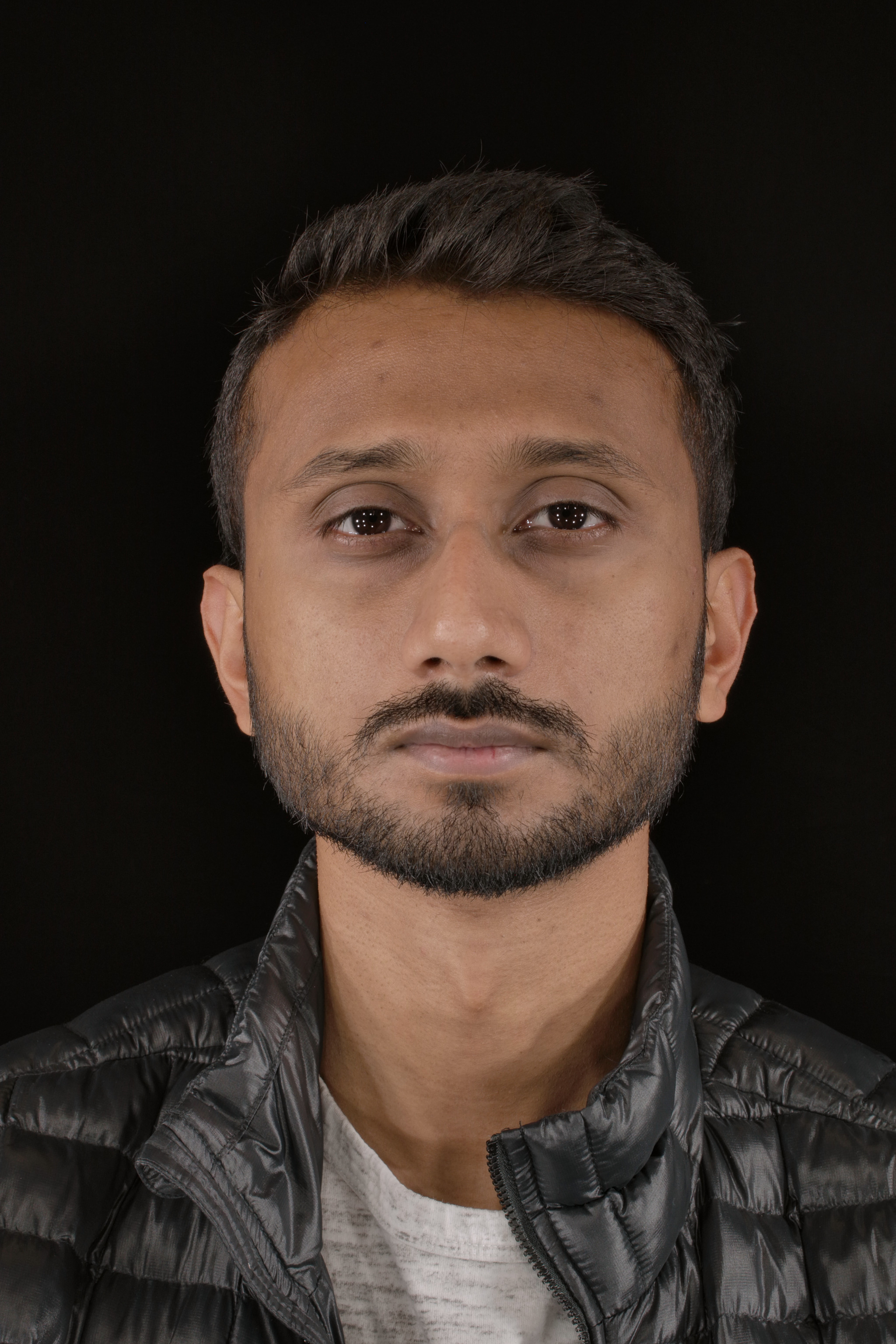}
&\includegraphics[height=\height]{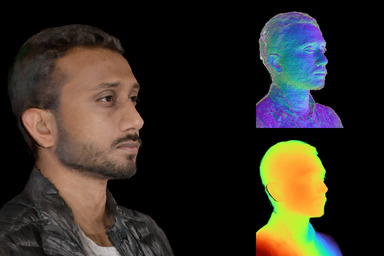}
&\includegraphics[height=\height]{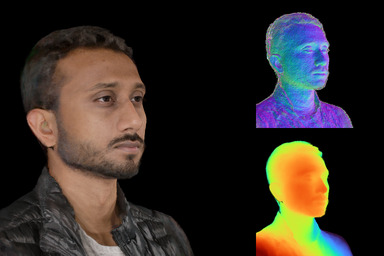}
&\includegraphics[height=\height]{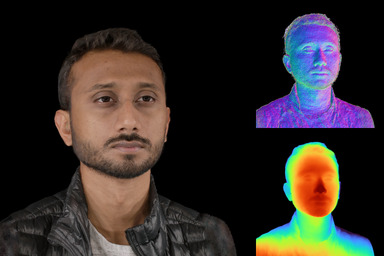}
&\includegraphics[height=\height]{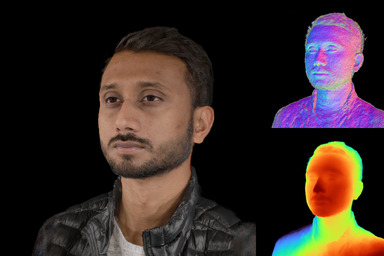}
&\includegraphics[height=\height]{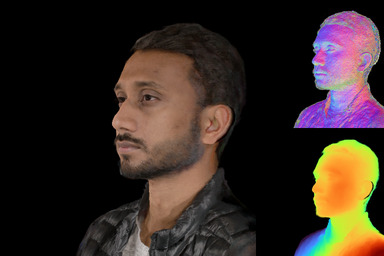}
\\
\includegraphics[height=\height]{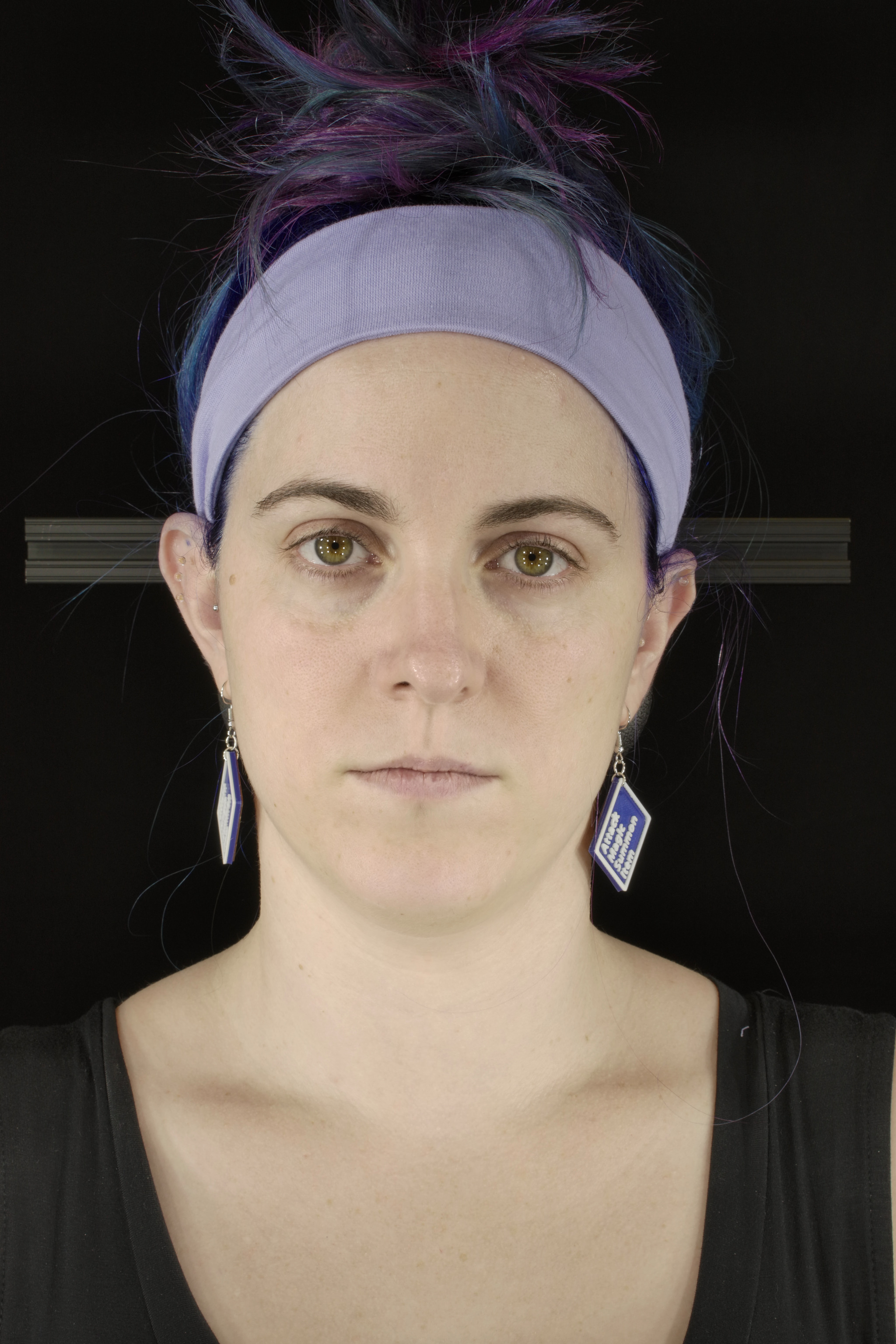}
&\includegraphics[height=\height]{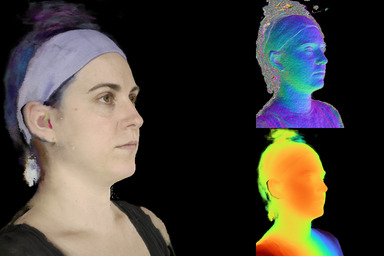}
&\includegraphics[height=\height]{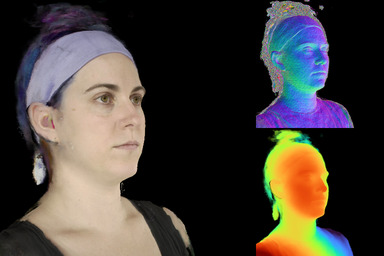}
&\includegraphics[height=\height]{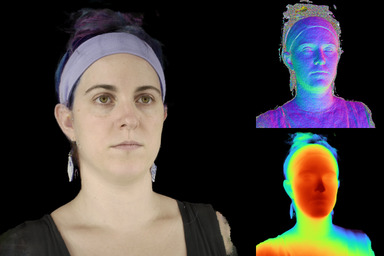}
&\includegraphics[height=\height]{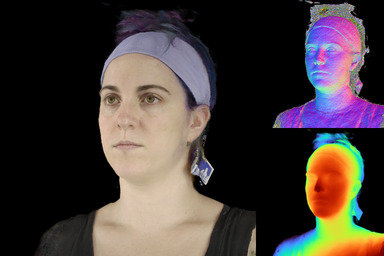}
&\includegraphics[height=\height]{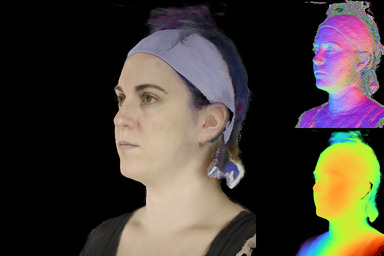}
\\
\end{tabular}
    \caption{Single Image Reconstruction Results. From left to right: input image captured using a studio setup, synthesised views around the subject face using a single frontal view for model fitting.
    \label{fig:single_image}\vspace{-1.5em}}
\end{center}
\end{figure*}

%% file: figures/ablation/ablation_regularization_short.tex
\begin{figure}
\centering
\newcommand{\height}{3.6cm}
\begin{tabular}{ccc}
\includegraphics[height=\height]{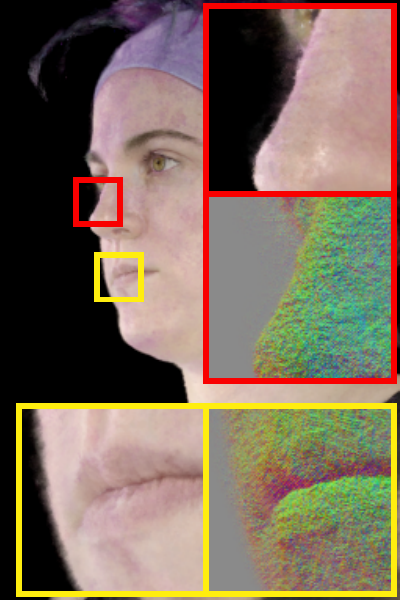} &
\includegraphics[height=\height]{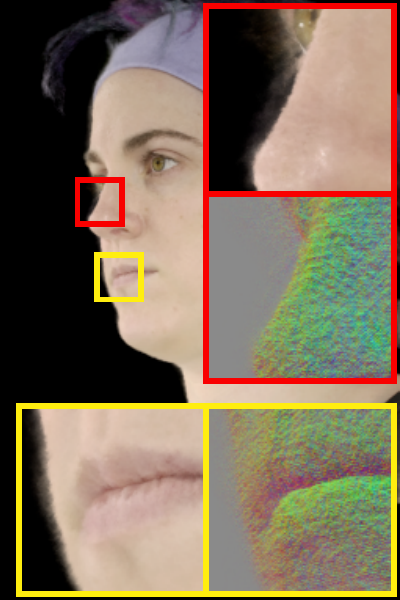} &
\includegraphics[height=\height]{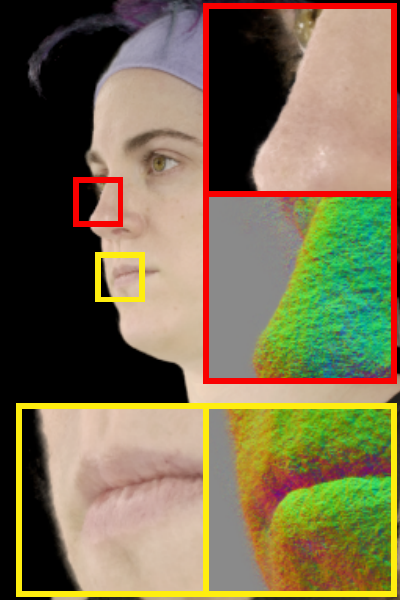} \\
$\mathcal{L}_\text{recon} + \mathcal{L}_\text{prop}$ & $+\;\mathcal{L}_v$ & $+ \;\mathcal{L}_\text{normal}$ \vspace{1em}
\end{tabular}
\caption{\label{fig:ablation_regularization_short} Ablation on the choice of regularisers. Without any regularisation, the view branch of
the model overfits to the view direction from the sparse in-
put signal. Additional regularisers allows the model to fit to a target identity from very sparse views.}
\end{figure}

%% file: tables/ablation_initialization_summary.tex
\begin{table}[t]
\small 
\begin{center}
\begin{tabular}{r ccc}
\hline
\textbf{Initialization} & 
    \multicolumn{1}{c}{\textbf{PSNR} $\uparrow$} &
    \multicolumn{1}{c}{\textbf{SSIM} $\uparrow$} & 
    \multicolumn{1}{c}{\textbf{LPIPS} $\downarrow$}\\
\hline
Furthest & 23.91 & 0.7876 & 0.2041  \\
Nearest & 24.41 & 0.7900 & 0.2002 \\
Mean & 24.61 & 0.7934 & 0.1959 \\
Noise & 24.66 & 0.7957 & 0.1998 \\ 
Zeros & 24.65 & 0.7941 & 0.1944 \\
\hline
Inversion (\textbf{Ours}) & \textbf{25.69} & \textbf{0.8040} & \textbf{0.1905} \\
\end{tabular}
\end{center}
\caption{Ablation on various types of initialising $\mathbf{w}_{\text{target}}$ when finetuning the model. We compare taking the mean across all latent codes during training; initialising it with zeros, Gaussian noise; and copying the latent code of the nearest or furthest neighbor in the training set. Inversion (\textbf{Ours}) performs best. Please refer to the supplementary material for visual examples.\vspace{-1.5em}}
\label{tab:ablation_initialization_summary}
\end{table}

%% file: figures/ablation/ablation_lighting.tex
\begin{figure}

\newcommand{\height}{1.5cm}
\newcommand{\psnr}[1]{\textcolor{cyan}{#1}}
\newcommand{\ssim}[1]{\textcolor{olive}{#1}}
\newcommand{\lpips}[1]{\textcolor{red}{#1}}

\begin{center}
\vspace{-1em}
\small
\setlength{\tabcolsep}{2pt}
\newcommand{\smallheight}{1cm}
\begin{tabular}{cccc ccc}
&& 
    \begin{tabular}{lll}
    \end{tabular}\\
\includegraphics[height=\smallheight]{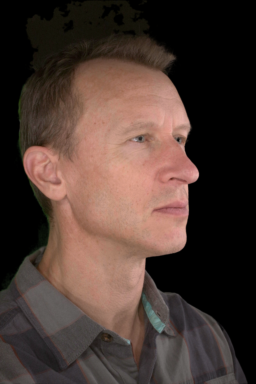}
    \includegraphics[height=\smallheight]{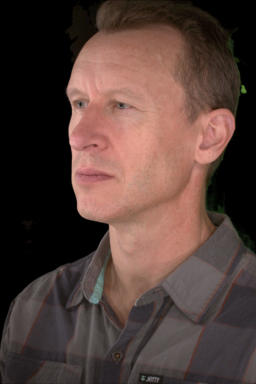} &
    \includegraphics[height=\height]{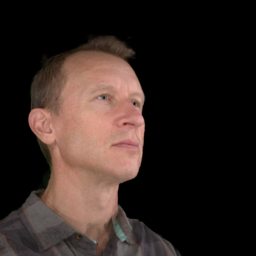} 
    \includegraphics[height=\height]{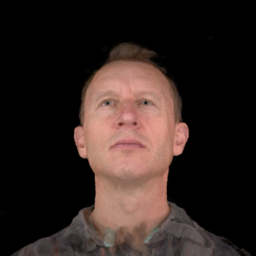}
    \includegraphics[height=\height]{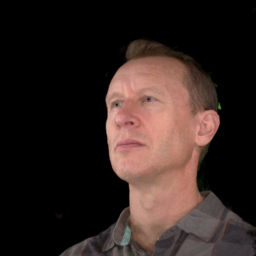} &
    \begin{tabular}[b]{l}
     \psnr{25.87} \\ \ssim{0.7688} \\ \lpips{0.1978}
    \end{tabular} \\
\includegraphics[height=\smallheight]{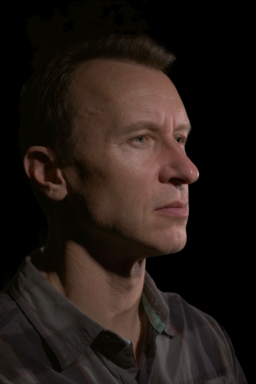}
    \includegraphics[height=\smallheight]{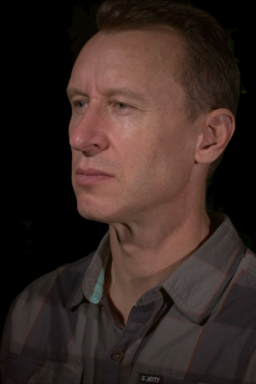} &
    \includegraphics[height=\height]{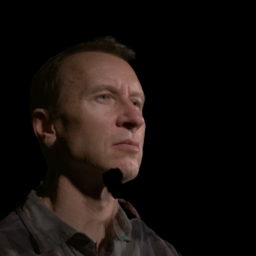} 
    \includegraphics[height=\height]{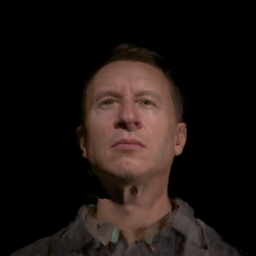}
    \includegraphics[height=\height]{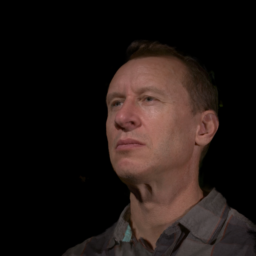} &
    \begin{tabular}[b]{l}
    \psnr{28.53} \\ \ssim{0.8500} \\ \lpips{0.2002}\\
    \end{tabular} \\
\includegraphics[height=\smallheight]{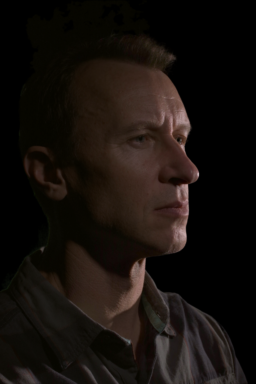}
    \includegraphics[height=\smallheight]{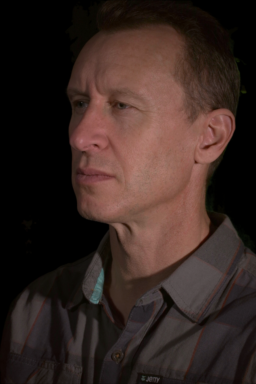} &
    \includegraphics[height=\height]{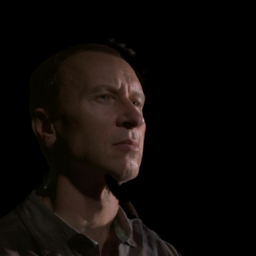} 
    \includegraphics[height=\height]{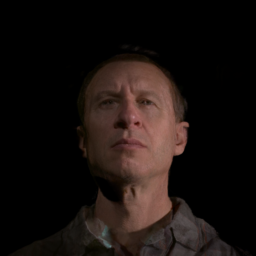}
    \includegraphics[height=\height]{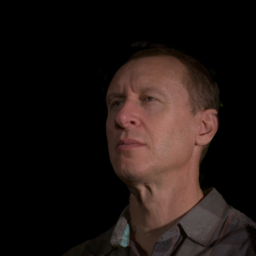} &
    \begin{tabular}[b]{l}
    \psnr{28.75} \\ \ssim{0.8480} \\ \lpips{0.1952}  \\
    \end{tabular} \\
 Input & Novel Views & Metrics\\
\end{tabular}
\end{center}

\caption{\label{fig:ablation_lighting}We show results for challenging lighting conditions with shadows and specular reflections, e.g., on the forehead. The right column lists \psnr{PSNR}, \ssim{SSIM}, and \lpips{LPIPS}.
}
\end{figure}

%% file: 07_conclusion.tex
\section{Conclusion}
We present a method that can create ultra high-resolution NeRFs of unseen subjects from as few as two images, yielding quality that surpasses other state-of-the-art methods.
While our method generalises well along several dimensions such as identity, resolution, viewpoint, and lighting, it is also impacted by the limitations of our dataset. While minor deviations from a neutral expression such as smiles can be synthesised, it struggles with extreme expressions. Clothing and accessories are also harder to synthesise. We show examples of such failure cases in the supplementary. Our model fitting process can take a considerable amount of time, particularly at higher resolutions. While some of these problems can be solved with more diverse data, others are excellent avenues for future work.

%% file: 99_supp.tex
\section{Supplementary Material}
This supplementary document provides more details about our experimental setting in Sec. \ref{sec:experimental_details} and supplementary results and ablations in Sec.~\ref{sec:supp_results}. For videos and ultra-high resolution results up to 4K, please see the project page.

\section{Detailed Experimental Setting}
\label{sec:experimental_details}
\subsection{Architecture and Hyperparameters}
\label{sec:supp_architecture}
In the following, we describe the architecture of the prior and the finetuned model in detail and list the hyperparameters we used for training and finetuning our models.

\subsubsection{Prior Model}
Following Mip-NeRF~\cite{mipnerf360}, the prior model consists of two MLPs. The first MLP is the \emph{proposal} network that only predicts density. The second MLP a neural radiance field (\emph{NeRF}) that predicts both density and colour. The proposal MLP has 4 linear layers with $(256 + 512)\times 256$ parameters: $256$ neurons for the features from the previous branch and 512 neurons for the concatenated latent code. The NeRF MLP has 8 linear layers with $(1024 + 512)\times 1024$ parameters: 1024 neurons for the features from the previous branch and 512 neurons for the concatenated latent code. The total parameter count of our prior model including all latent codes is 14.6 Mio.

During training and inference, we use three hierarchical sampling steps~\cite{mildenhall2020nerf}. The first step uses 256 proposal samples, the second step 256 refined proposal samples, and the third step 128 NeRF samples. 

We use the same number of positional encoding frequencies for both the proposal and the NeRF MLPs. The integrated positional encoding for the trunk networks $\hat \gamma_{\bf x}(\cdot)$ has 12 levels; the positional encoding $\gamma_{\bf v}(\cdot)$ for the view direction has 4 levels, and it appends the view direction without positional encoding. The view branch of the NeRF MLP has a bottleneck with width 256. The positionally-encoded view direction is concatenated to the bottleneck features and processed by a linear layer of width $(256 + 27) \times 128$ before being projected to RGB (256 bottleneck features and 27 features from the positional encoding of the view direction).

We optimise the prior model as an auto-decoder~\cite{bojanowski2018optimizing}, where each identity has a latent code with 512 dimensions. 
Each training step samples 128 random rays from 8 views of 64 identities, which yields a batch size of $65,536$. We train our prior model for 1 Mio. steps, which takes 144 hours (approximately 6 days) on 36 TPUs. We optimise our model using Adam~\cite{KingmB2015} with $\beta_1 = 0.9, \beta_2=0.999$. The learning rate starts at $0.002$ and exponentially decays to $0.00002$. We clip gradients with norms larger than 0.001.

\subsubsection{Inversion}
We perform inversion on the prior model to find a good initialisation for the finetuning. In each step, we sample 8 random patches of size $32 \times 32$ from all available views. 
We initialise the new latent code with zeros. The optimisation uses Adam with $\beta_1 = 0.9, \beta_2=0.999$ and a fixed learning rate of $0.001$. We optimise for $1,500$ steps on 4 TPUs, which takes 10 minutes.

\subsubsection{Finetuned Model}

The architecture of the finetuned model is the same as the prior model, except for an additional linear layer that maps the features from the trunk to 3-d normal vectors.

We create batches of $8,912$ rays by sampling random pixels from all available views. We start with a learning rate of 0.001 and exponentially decay to $0.00002$. The number of optimisation step depends on the resolution. For low-resolution ($256 \times 256$), we optimise for $25,000$ steps. We increase the number of optimisation steps for higher resolutions: $50,000$ steps for $512\times 512$; $100,000$ steps for $1024\times 1024$; $200,000$ steps for $2048\times 2048$; and $300,000$ steps for $4096\times 4096$. We always optimise on four TPUs. The model finetuning takes 4 hours for $25,000$ steps and linearly increases for more training steps.

\subsubsection{Camera Alignment}
A crucial preprocessing step is to align all cameras to a canonical pose. As described in the main paper, we estimate five 3D keypoints on the outer eye corners, nose, mouth, and chin and calculate a similarity transform the the same five keypoints in a canonical space using Procrustes analysis. The canonical keypoints are computed as the median keypoint location across the first 260 subjects in our training set. Fig.~\ref{fig:keypoints} shows an example.
\input{figures/keypoints}
\input{figures/dataset_dist}

\subsection{Studio Dataset}

\label{sec:studio_dataset}
Our studio dataset consists of 1450 volunteers who were prompted to optionally self-report various characteristics like age, gender, skin colour, and hair colour. We report the statistics here and in Fig.~
\ref{fig:dataset_dist}. 60\% of the participants were male, 38\% female, 0.2\% non-binary and the rest preferred not to state. The age of the participants was heavily centered in the 24-50 age group. We also note the bias in appearance characteristics.

The participants were also given the option to wear or remove their glasses, hence a very small percentage $\sim$1\% wore glasses during capture. The capture was performed over a period of many months. Initial captures contain a black background and were later changed to a green screen to allow for better foreground segmentation if required. We do not mask out the background during prior model training. During finetuning, we estimate a foreground mask with a robust pretrained estimator~\cite{pandey2021total}. Hence, our method works without any constraints on the background, as long as the camera poses are accurate.

\section{Supplementary Results and Analysis}
\label{sec:supp_results}

\input{figures/comparison/comparison_holobooth_256}
\input{figures/comparison/comparison_facescape_full}
\input{figures/single_image/single_image_full}

\input{figures/inthewild/inthewild}

\input{figures/ablation/ablation_regularization}
\input{tables/ablation_regularization}

\input{figures/ablation/ablation_frozen_latent}
\input{tables/ablation_frozenlatent}
\input{figures/ablation/ablation_initialization}
\input{tables/ablation_initialization}
\input{figures/ablation/ablation_expression}

\input{tables/ablation_views}

This section supplements the results in the main paper with more visuals and detailed metrics. We provide supplementary results for comparisons related works in Sec.~\ref{supp:sec:comparisons}, more visuals for one- and few-shot synthesis in the studio setting and in-the-wild in Sec.~\ref{supp:sec:fewshot}, and a detailed analysis of our ablations in Sec.~\ref{supp:sec:ablation}.

\subsection{Supplementary Comparisons}
\label{supp:sec:comparisons}

This section supplements the comparison from the main paper with detailed metrics and visuals for individual holdout subjects.

\subsubsection{Comparisons on Our Studio Dataset}
\label{sec:details_comp_studio}
This section provides supplementary results on our multiview studio dataset described in the main paper and in Sec.~\ref{sec:studio_dataset}. Note that our goal is novel view synthesis so we refrain from comparing with methods that explicitly target geometry reconstruction~\cite{h3dnet,niemeyer2020differentiable,oechsle2021unisurf,yariv2021volume,wang2021neus}. 

We train the competing methods~\cite{tancik2021learned,eg3d,regnerf,Yang2023FreeNeRF,keypointnerf} on our dataset and compare with our results in Tbl.~\ref{tab:comp_holobooth}. 

In the following, we describe the experimental details for each competing method.

\begin{figure}[t]
\centering
\newcommand{\height}{4cm}
\includegraphics[height=\height]{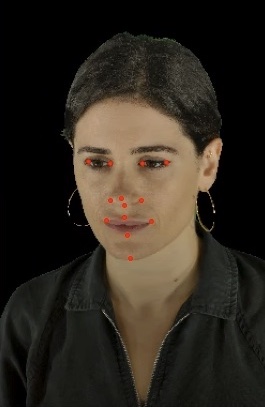}
\caption{Keypoints used for training KeypointNeRF~\cite{keypointnerf} with our data.}
\label{fig:keypoints_keypointnerf}
\end{figure}

For KeypointNeRF, we use their publicly available code and their default training and network settings. We manually chose 13 keypoints that closely resemble the ones shown in their paper (Fig. \ref{fig:keypoints_keypointnerf}) and compute the near and far planes from our own dataset. We made a considerable effort to train them at 1K resolution, but we found that their results at the resolution 256 is of much higher quality than their results at 1K. Therefore, we present their results at both 1K resolution (Tbl.~\ref{tab:comp_holobooth}) and at 256 resolution (Tbl.~\ref{tab:comparison256}). For the lower resolution comparison, we compare with our lower-resolution prior model trained at resolution $256\times 384$.

For the comparison with RegNeRF~\cite{niemeyer2022regnerf}, we train their model with the default settings provided by the authors for the DTU dataset~\cite{jensen2014large}, except for adjusting the near / far planes and scene scaling. We also disable the loss from the appearance regulariser because the model is not available. 

For FreeNeRF, we implement their frequency regularisation with a 90\% schedule into our pipeline. We do not employ their occlusion regularisation because it causes transparent surfaces and floaters on our dataset.

For learnit~\cite{tancik2021learned}, we adapt their publicly available notebook to work with our dataset. For training the meta model, we set the batch size to 4096, the number of inner steps to 64, the number of samples along the ray to 128, and train for 15,000 steps. We run the inference-time optimisation for the same number of steps as ours: 100,000 steps.

For the EG3D-based prior, we train a prior model with a triplane representation as proposed in Chan et al.~\cite{eg3d}. The model is trained as an auto-decoder model similar to ours. We simultaneously optimize a per-identity latent code and the network weights to obtain an EG3D prior model that is finetuned to sparse views of a target subject for the same number of steps as ours. We do not apply our additional regularisers when finetuning EG3D.

We train the EG3D prior on low-resolution images at resolution $256\times 256$ that are super-resolved to resolution $1024\times 1024$. The triplane resolution is $256\times 256$ and the per-identity latent codes have dimensionality $512$. Since the EG3D model requires rendering the full image, we reduce the number of initial samples per ray to 64 and the number of importance samples to 8.

For all methods, we perform the same inference-time bounding box based culling as we did for our method. Table~\ref{tab:comp_holobooth} lists metrics for experiments on 2, 3, 5, and 7 views and Fig.~\ref{fig:comparison_1k_suba}, \ref{fig:comparison_1k_subb}, and \ref{fig:comparison_1k_subc} show visual examples. Our method consistently outperforms related works.

We do not compare with DINER~\cite{diner}, Sparse NeRF~\cite{wang2023sparsenerf}, and SPARF~\cite{sparf2023} on our dataset because their training code is not publicly available at the time of submission.
\input{tables/comparison256}
\input{tables/comparison_holobooth_views}

\subsubsection{Comparison on FaceScape}
\label{sec:details_comp_facescape}
Figure~\ref{fig:comparison_facescape_full} adds more examples for the comparison with Facescape~\cite{facescape}, and Tbl.~\ref{tab:facescape} lists metrics.
 \input{tables/comparison_facescape}

For the comparison on FaceScape~\cite{facescape}, we obtain the outputs directly from the authors of DINER~\cite{diner}. For each target identity, we perform model finetuning on two different subset of four views and average the scores. Since we develop our method on neutral faces, we filter out faces with non-neutral expressions.

For the comparison with RegNeRF~\cite{niemeyer2022regnerf}, we follow the same protocol as described in Sec.~\ref{sec:details_comp_studio}. We follow the default settings provided by the authors for the DTU dataset~\cite{jensen2014large}, but adjust the near / far planes and scene scaling. Again, we  disable the loss from the appearance regulariser.

For the EG3D-based prior~\cite{chan2022efficient}, we train their model on Celeb-A~\cite{celeba} dataset at a 256 tri-plane and image resolution without the super-resolution module to ensure 3D consistent results. We note that their discretised volume representation leads to blurry results.  

\subsection{Few-shot Synthesis}
\label{supp:sec:fewshot}
\paragraph{Ultra High-res}
Our main setting is fitting to two or more views at a ultra-high resolution up to 4K. This goes far beyond the resolution of the prior model ($512\times 768$). Using at least two views provides the coverage from side angles such that the model can reconstruct intricate details like individual skin pores or a beard, which are not visible at lower resolutions. Please see the main paper and the project page for results.

\paragraph{Single Image} To showcase the robustness of our method, we show results for synthesising novel views from as little as a \emph{single image} at the resolution of our prior model ($512\times 768$), see the main paper and Fig.~\ref{fig:single_image_full}.

\paragraph{In-the-wild}
Fig. \ref{fig:inthewild} shows examples for in-the-wild captures with a mobile camera. The project page shows videos and adds high resolution results for in-the-wild captures with a smartphone camera.

\subsection{Ablation}
\label{supp:sec:ablation}
We perform extensive ablations on our prior model and on the finetuning algorithm. For the prior model, we ablate the impact of the number of training identities and the prior model resolution (Tbl.~\ref{tab:ablation_prior}). For the finetuning algorithm, we ablate regularisation terms (Tbl.~\ref{tab:ablation_regularization} and Fig.~\ref{fig:ablation_regularization}), number of views (Tbl.~\ref{tab:ablation_views} and Fig.~\ref{fig:comparison_1k_suba}, \ref{fig:comparison_1k_subb}, and \ref{fig:comparison_1k_subc}), and initialisation techniques (Tbl.~\ref{tab:ablation_initialization}). We also ablate the effect of finetuning the full model including the latent codes vs. only finetuning the model parameters (Tbl.~\ref{tab:frozenlatent} and Fig.~\ref{fig:ablation_frozen_latent}).

We provide all metrics cropped to the face region and evaluate on six holdout views to have comparable numbers across all ablations. All metrics are computed after finetuning for each of the three holdout subjects at resolution $1024 \times 1024$.

\subsubsection{Prior Model}
Table~\ref{tab:ablation_prior} ablates the effect of different variants of our prior model. We compare these variants of the prior model: lower resolution ($256 \times 384$ instead of $512 \times 768$) and fewer training identities. The results show that a more diverse prior model performs better while a lower resolution prior model might not necessarily be required.
\input{tables/ablation_prior}

\subsubsection{Model Finetuning}
\paragraph{Initialisation}
This supplementary document complements the ablations in the main paper with metrics showing the benefits of the chosen regularisation (Tbl.~\ref{tab:ablation_regularization} and Fig. \ref{fig:ablation_regularization}) and visual examples for different initialisation techniques (Tbl.~\ref{tab:ablation_regularization} and Fig.~\ref{fig:ablation_initialization}). For the \emph{Nearest} (\emph{Furthest}) Neighbour initialisation, we compute image embeddings using a pretrained face recognition network~\cite{schroff2015facenet}. We compute the similarity of the mean embedding of all target images with embeddings computed on a frontal rendering of all reconstructed training identities. 

\paragraph{Number of Views}
We also provide a supplementary ablation on the performance when a different number of views are available in Tbl.~\ref{tab:ablation_views}. 
Figures~\ref{fig:comparison_1k_suba}, \ref{fig:comparison_1k_subb}, and \ref{fig:comparison_1k_subc}), and the project page shows visual results.

\paragraph{Frozen Latent Code} 

Table~\ref{tab:frozenlatent} lists metrics and Fig.~\ref{fig:ablation_frozen_latent} shows the rendered images. We do not observe a strong difference in performance.

\subsection{Limitations}
\input{figures/comparison/comparison_holobooth_subA}
\input{figures/comparison/comparison_holobooth_subB}
\input{figures/comparison/comparison_holobooth_subC}

Our model is trained on neutral faces with a closed mouth. It can handle mild expressions (e.g., closed eyes and a slightly open mouth) but fails for strong expressions and teeth, see Fig.~\ref{fig:ablation_expression}.

While our results show robustness to in-the-wild settings, it is sensitive to correct camera calibration. In the reconstruction, this is particularly noticeable for thin structures like the eyes and eyelids. We also assume that the subject does not move during the capture.

Also, our prior model does not cover accessories like glasses or hats and reconstructions thereof are therefore not 3D consistent. Please see the project page for examples.

%% file: figures/keypoints.tex
\begin{figure}[t]
    \centering
  \includegraphics[width=0.3\textwidth,angle=90]
                  {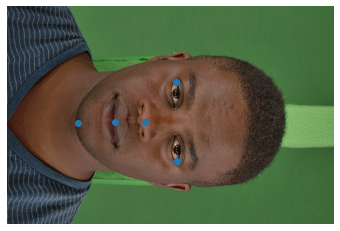}
    \caption{Visualisation of the five keypoints used for aligning captured subjects to a canonical pose.
    \label{fig:keypoints}}
\end{figure}

%% file: figures/dataset_dist.tex
\begin{figure*}[t]
    \centering
  \includegraphics[width=\textwidth]
                  {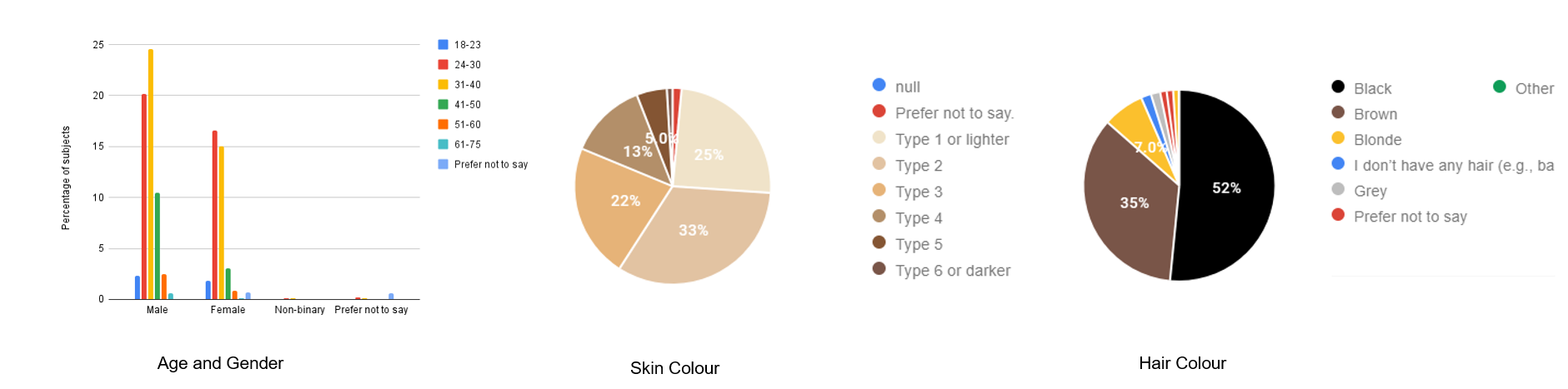}
    \caption{Distribution of characteristics in our dataset: we report the percentage distribution of our dataset by age, gender, skin colour and hair colour.
    \label{fig:dataset_dist}}
\end{figure*}

%% file: figures/comparison/comparison_holobooth_256.tex
\begin{figure*}[ht]

\begin{center}
\small
\setlength{\tabcolsep}{2pt}

\newcommand{\height}{4cm}
\begin{tabular}{cccccc}
\rotatebox{90}{KeypointNeRF}
  & \includegraphics[height=\height]
 {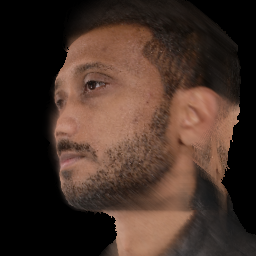} 
  & \includegraphics[height=\height]
 {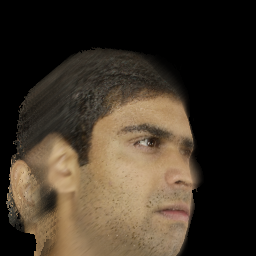} 
  & \includegraphics[height=\height]
 {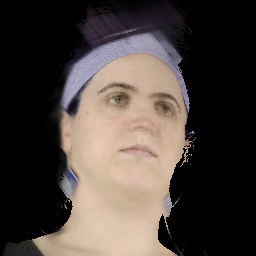}  \\
\rotatebox{90}{\textbf{Ours}}
  & \includegraphics[height=\height]
 {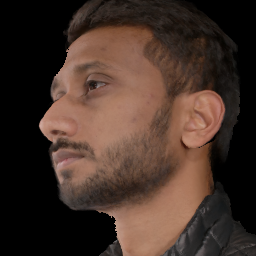} 
  & \includegraphics[height=\height]
 {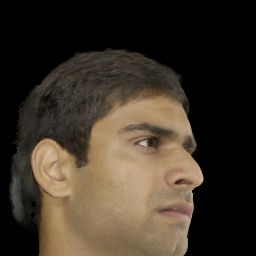} 
  & \includegraphics[height=\height]
 {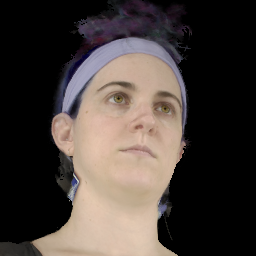} 
    \\  
\rotatebox{90}{GT}
  & \includegraphics[height=\height]
 {figures/comparison/holobooth/gt_0_gt} 
  & \includegraphics[height=\height]
 {figures/comparison/holobooth/gt_2_gt} 
  & \includegraphics[height=\height]
 {figures/comparison/holobooth/gt_1_gt} 
    \\    
& Subject A & Subject B & Subject C
\end{tabular}
\end{center}

\caption{\label{fig:comparison256} Visual comparison with KeypointNeRF \cite{keypointnerf} on low-resolution. Please see Tab.~\ref{tab:comparison256} for metrics.}

\end{figure*}

%% file: figures/comparison/comparison_facescape_full.tex
\begin{figure*}[ht]

\begin{center}
\small
\setlength{\tabcolsep}{2pt}

\newcommand{\height}{3cm}
\newcommand{\wordsubject}{facescape/sub212_01_target28_ref7-55-40-23}
\newcommand{\eyebrowsubject}{facescape/sub340_01_target52_ref3-45-17-54}
\newcommand{\eyesubject}{facescape/sub122_01_target13_ref52-37-21-44}
\newcommand{\mouthsubject}{facescape/sub344_01_target19_ref34-35-23-24}
\newcommand{\earsubject}{facescape/sub340_01_target29_ref7-50-37-48}

\begin{tabular}{cc|ccccc}
\includegraphics[height=\height]{figures/comparison/\wordsubject_input} &
\includegraphics[height=\height]{figures/comparison/\wordsubject_gt} &
\includegraphics[height=\height]{figures/comparison/\wordsubject_regnerf} &
\includegraphics[height=\height]{figures/comparison/\wordsubject_eg3d} &
\includegraphics[height=\height]{figures/comparison/\wordsubject_keypointnerf} &
\includegraphics[height=\height]{figures/comparison/\wordsubject_diner} &
\includegraphics[height=\height]{figures/comparison/\wordsubject_ours} \\
\includegraphics[height=\height]{figures/comparison/\eyebrowsubject_input} &
\includegraphics[height=\height]{figures/comparison/\eyebrowsubject_gt} &
\includegraphics[height=\height]{figures/comparison/\eyebrowsubject_regnerf} &
\includegraphics[height=\height]{figures/comparison/\eyebrowsubject_eg3d} &
\includegraphics[height=\height]{figures/comparison/\eyebrowsubject_keypointnerf} &
\includegraphics[height=\height]{figures/comparison/\eyebrowsubject_diner} &
\includegraphics[height=\height]{figures/comparison/\eyebrowsubject_ours} \\
\includegraphics[height=\height]{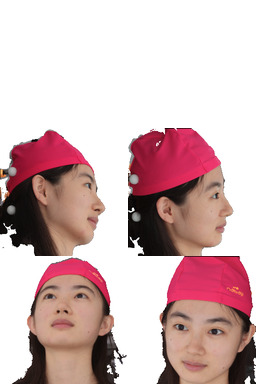} &
\includegraphics[height=\height]{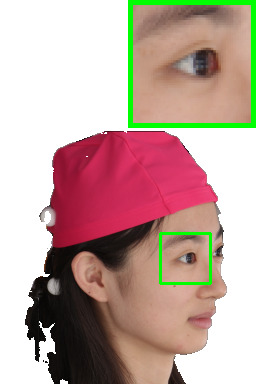} &
\includegraphics[height=\height]{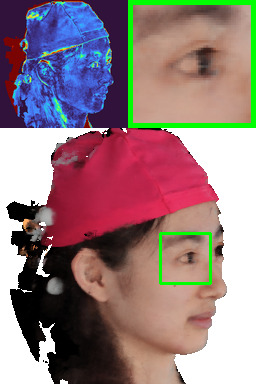} &
\includegraphics[height=\height]{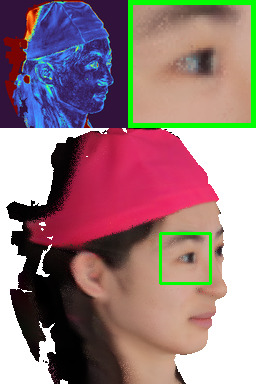} &
\includegraphics[height=\height]{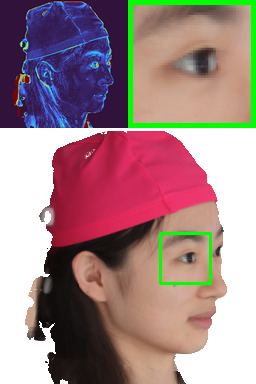} &
\includegraphics[height=\height]{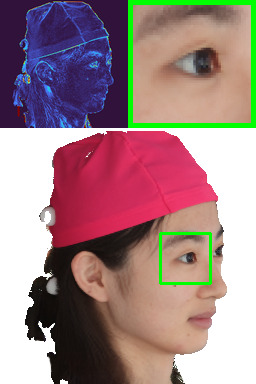} &
\includegraphics[height=\height]{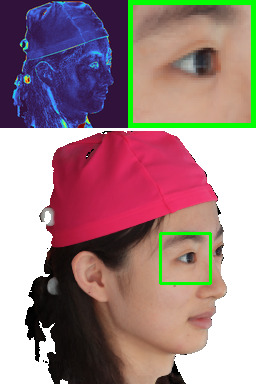} \\
\includegraphics[height=\height]{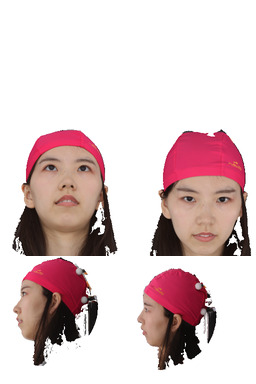} &
\includegraphics[height=\height]{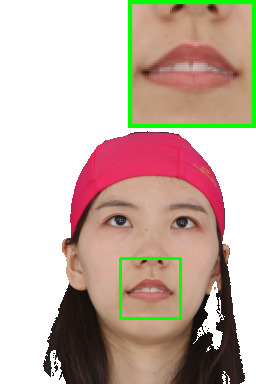} &
\includegraphics[height=\height]{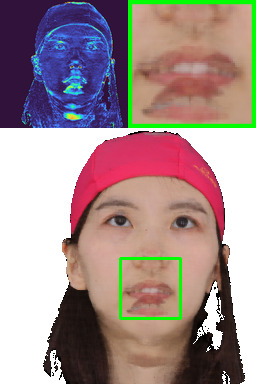} &
\includegraphics[height=\height]{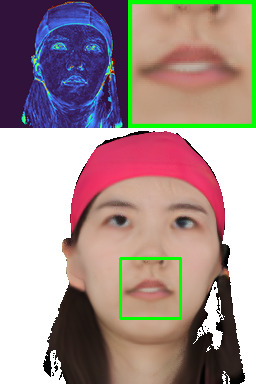} &
\includegraphics[height=\height]{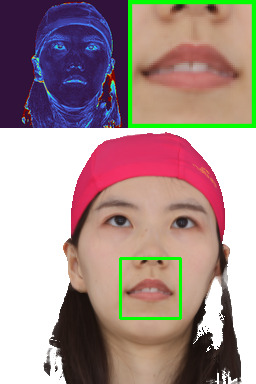} &
\includegraphics[height=\height]{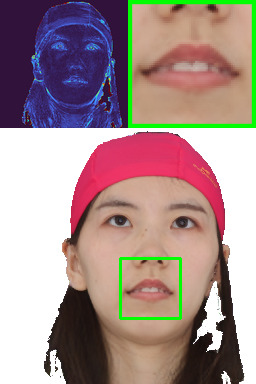} &
\includegraphics[height=\height]{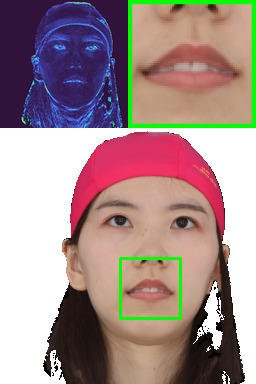} \\
\includegraphics[height=\height]{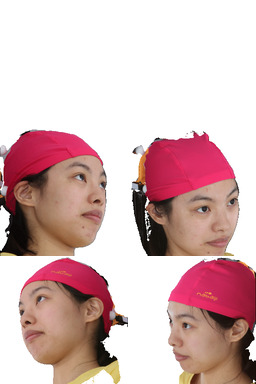} &
\includegraphics[height=\height]{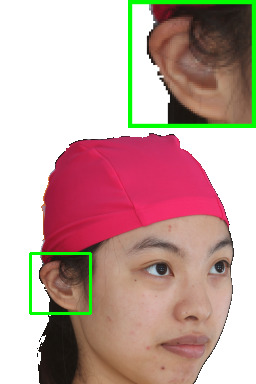} &
\includegraphics[height=\height]{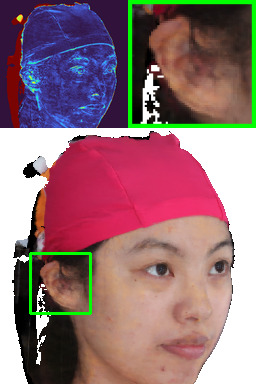} &
\includegraphics[height=\height]{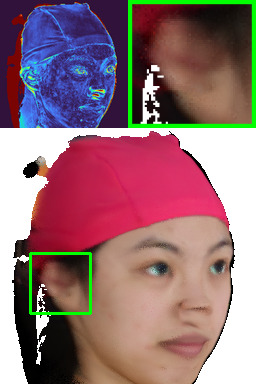} &
\includegraphics[height=\height]{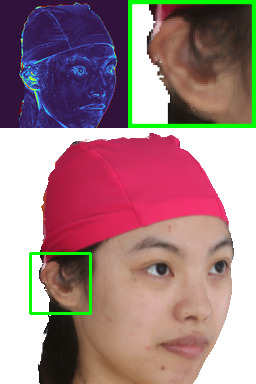} &
\includegraphics[height=\height]{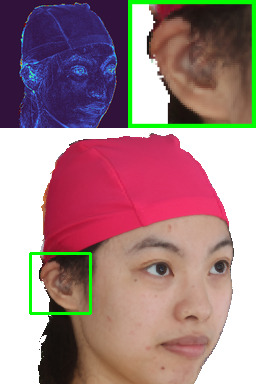} &
\includegraphics[height=\height]{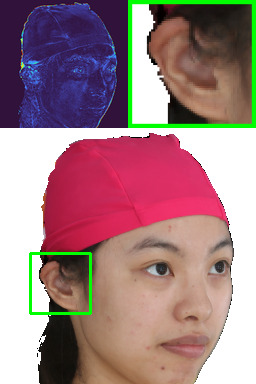} \\
Input & Ground Truth & RegNeRF~\cite{regnerf} & EG3D-based Prior~\cite{eg3d} & KeypointNeRF~\cite{keypointnerf} & DINER~\cite{diner} & Ours
\end{tabular}
\end{center}
    \caption{Comparison with the state-of-the-art for novel view synthesis from sparse views on holdout identities from FaceScape~\cite{facescape}. For each identities, given four views as input, we show novel view reconstruction results and the L1 residue. 
    \label{fig:comparison_facescape_full}}

\end{figure*}

%% file: figures/single_image/single_image_full.tex
\begin{figure*}
\hspace{30pt}
\begin{minipage}{443pt}
\begin{center}
\setlength\tabcolsep{1pt}
\newcommand{\crop}{1.0cm}
\newcommand{\cropsmall}{0.4cm}
\newcommand{\height}{3.5cm}
\begin{tabular}{ccc}
\includegraphics[height=\height]{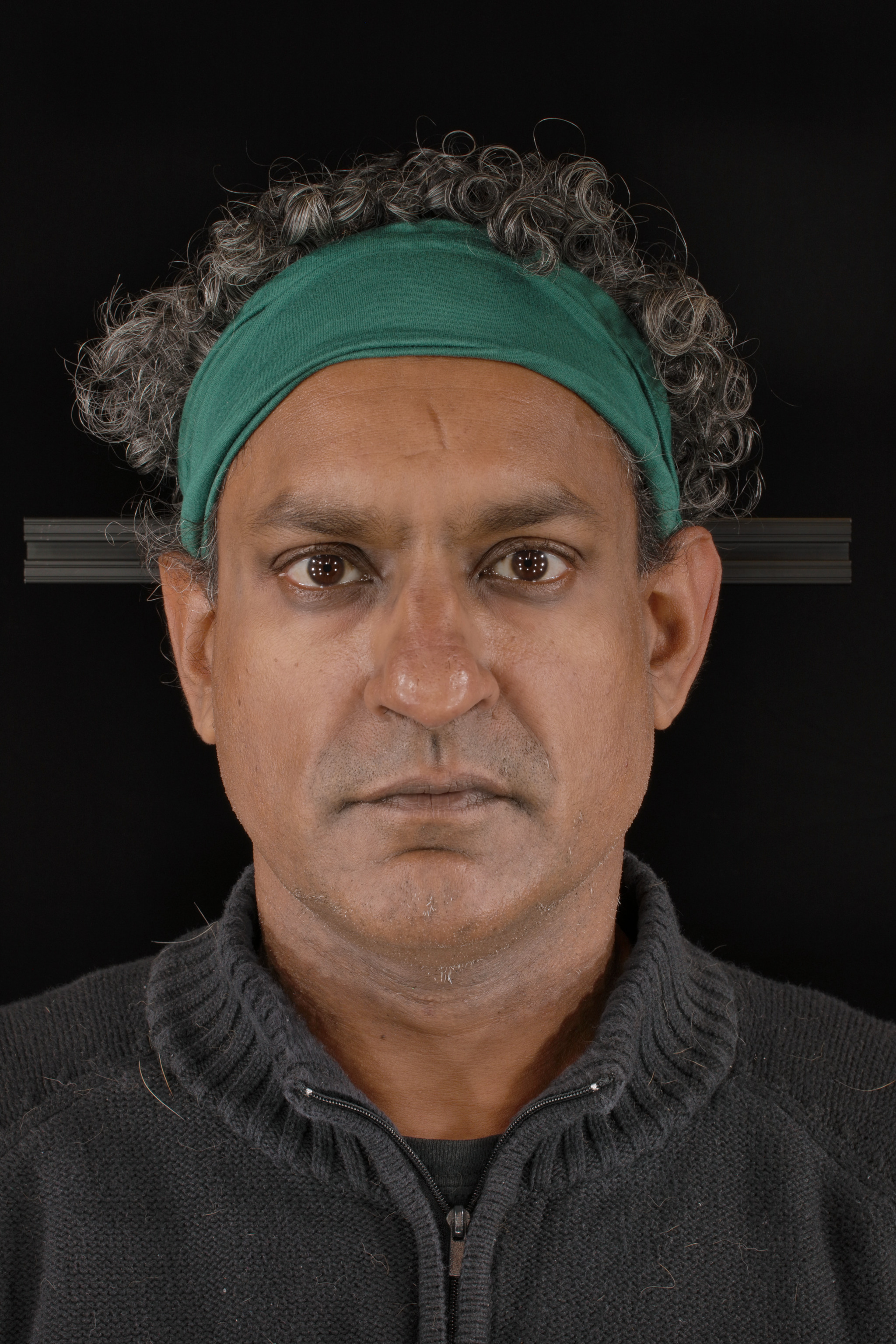}
&\includegraphics[height=\height]{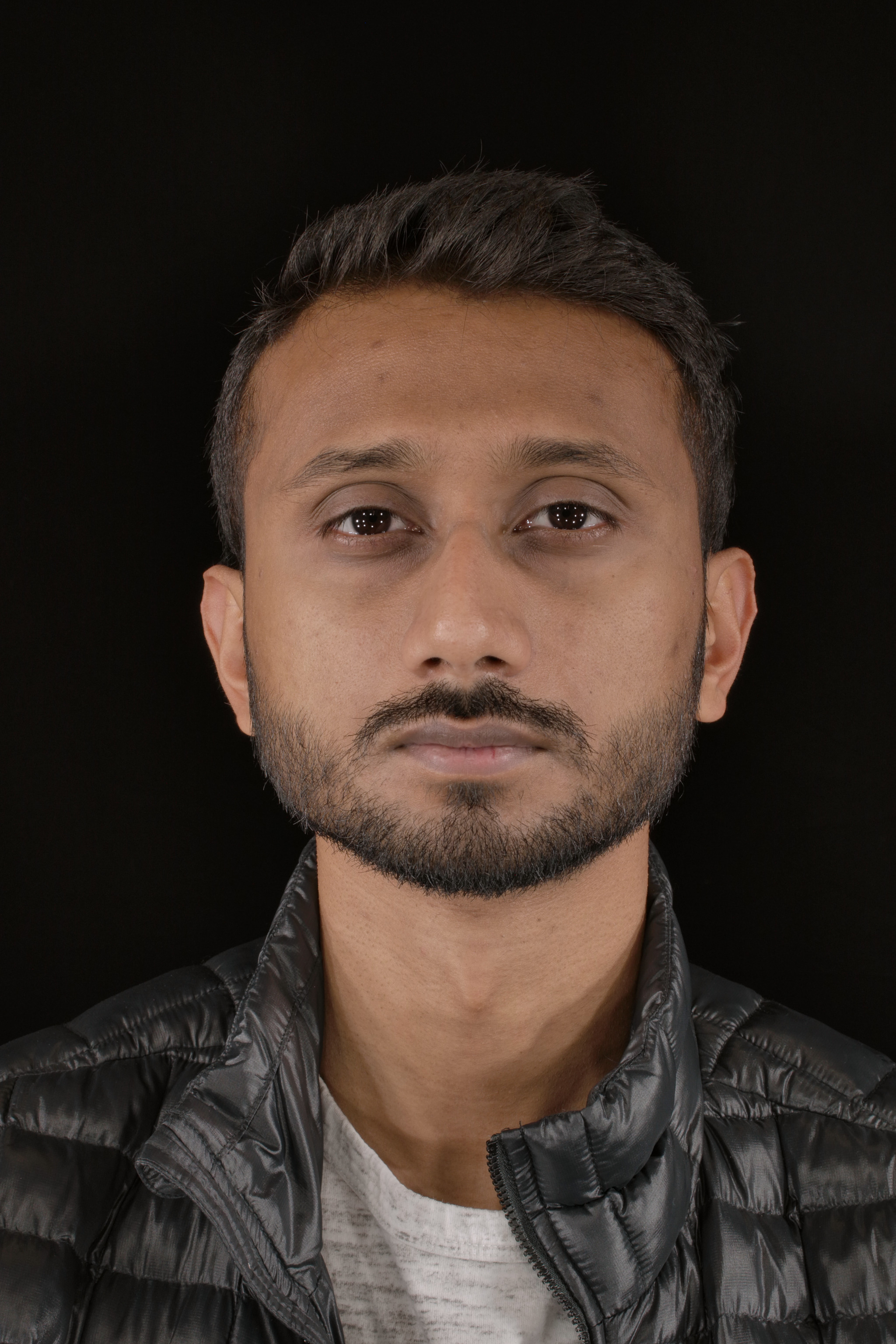}
&\includegraphics[height=\height]{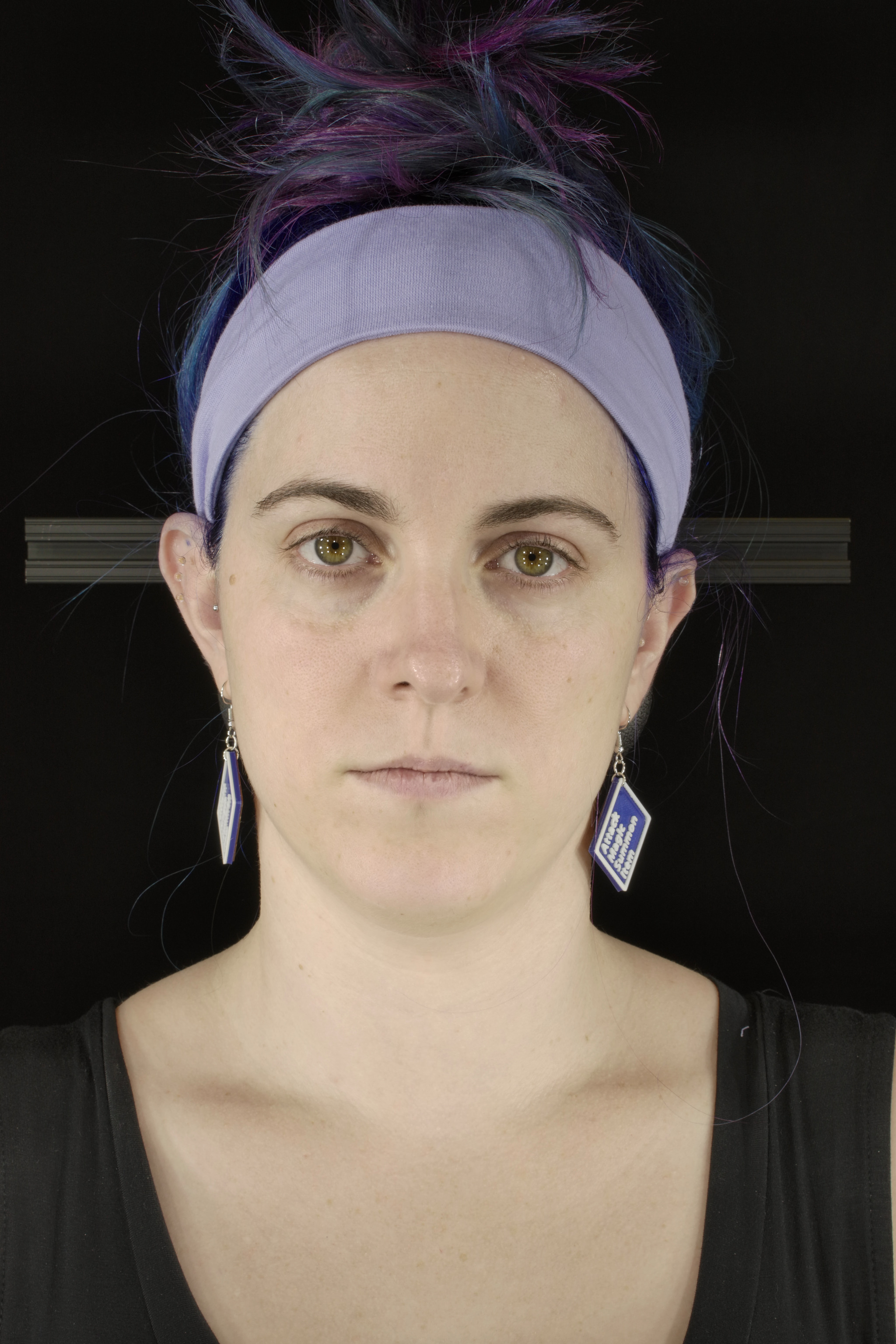}
\\
\includegraphics[height=\height]{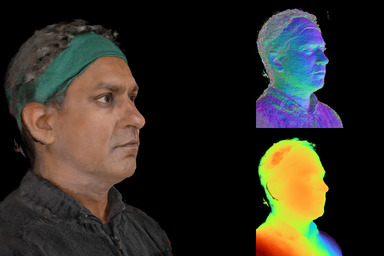}
&\includegraphics[height=\height]{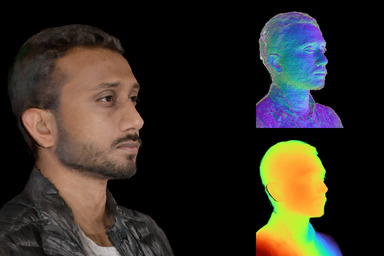}
&\includegraphics[height=\height]{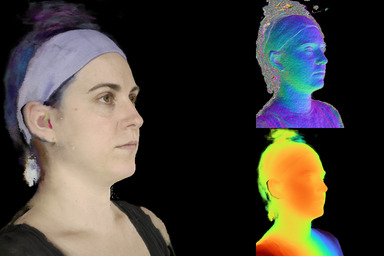}
\\
\includegraphics[height=\height]{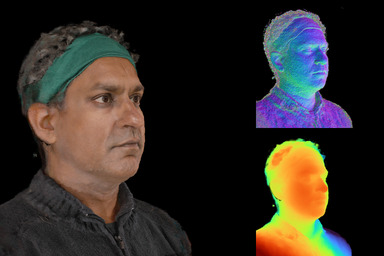}
&\includegraphics[height=\height]{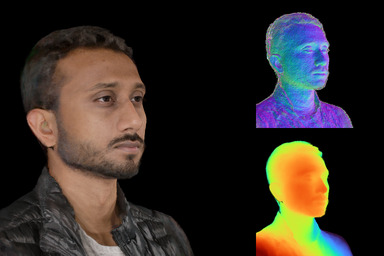}
&\includegraphics[height=\height]{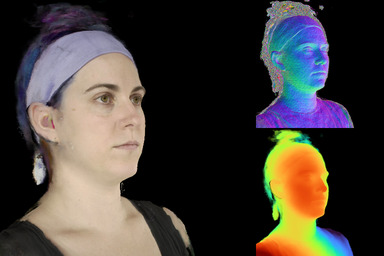}
\\
\includegraphics[height=\height]{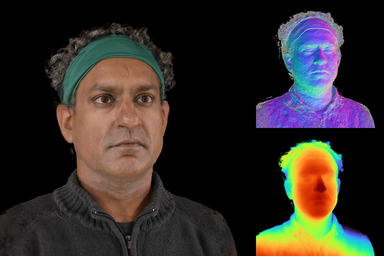}
&\includegraphics[height=\height]{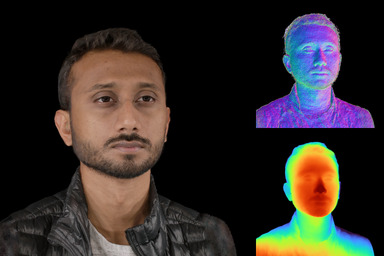}
&\includegraphics[height=\height]{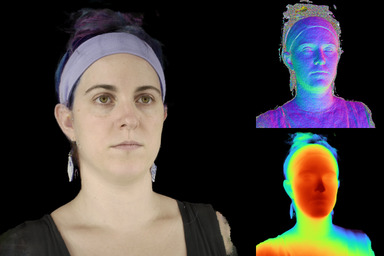}
\\
\includegraphics[height=\height]{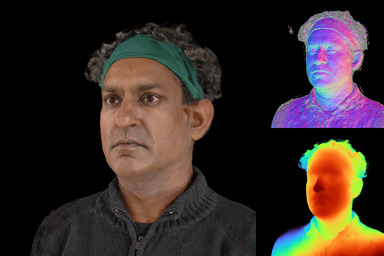}
&\includegraphics[height=\height]{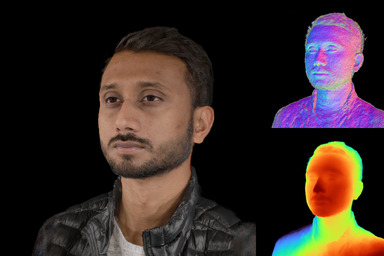}
&\includegraphics[height=\height]{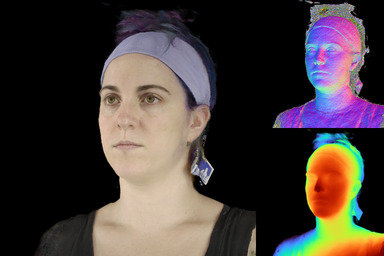}
\\
\includegraphics[height=\height]{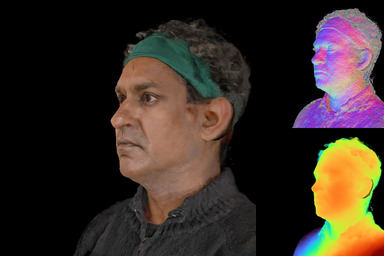}
&\includegraphics[height=\height]{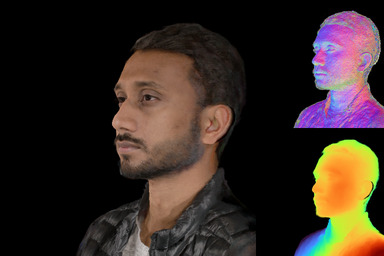}
&\includegraphics[height=\height]{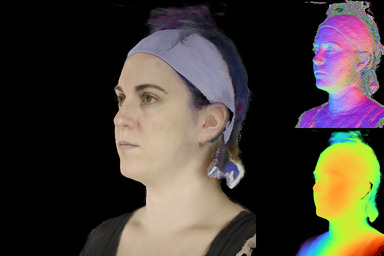}
\\
\end{tabular}
    \caption{Single image reconstruction results from the main paper at higher resolution. The top row shows the input image captured in a studio setup. The rows below show synthesised views around the subject face using the image in the top row for model fitting. The inlays show the normals (top) and depth (bottom).
    \label{fig:single_image_full}}
\end{center}
\end{minipage}
\end{figure*}

%% file: figures/inthewild/inthewild.tex
\begin{figure*}
\begin{center}
\setlength\tabcolsep{1pt}
\newcommand{\crop}{0.8cm}
\newcommand{\cropsmall}{0.4cm}
\newcommand{\height}{3.5cm}
\begin{tabular}{cccccc}
\includegraphics[height=\height]{figures/inthewild/gt/inthewild_00_cam05}\includegraphics[height=\height]{figures/inthewild/gt/inthewild_00_cam02} 
    & 
    \includegraphics[height=\height]{figures/inthewild/gt/inthewild_01_cam13}\includegraphics[height=\height]{figures/inthewild/gt/inthewild_01_cam15}
    & \includegraphics[height=\height]{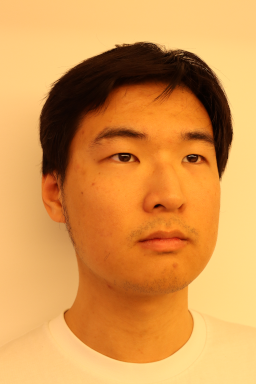}\includegraphics[height=\height]{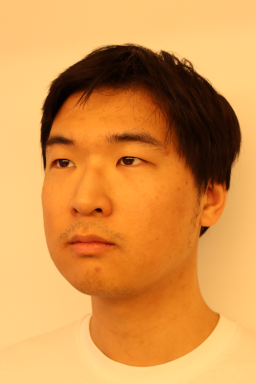} \\
\includegraphics[height=\height]{figures/inthewild/inthewild_0_0}
    & \includegraphics[height=\height]{figures/inthewild/inthewild_1_0}
    & \includegraphics[height=\height]{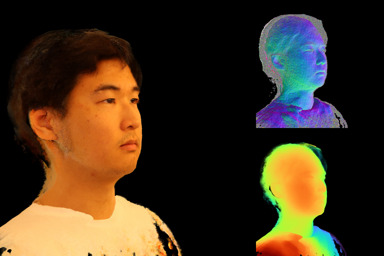} \\
\includegraphics[height=\height]{figures/inthewild/inthewild_0_1}
    & \includegraphics[height=\height]{figures/inthewild/inthewild_1_1}
    & \includegraphics[height=\height]{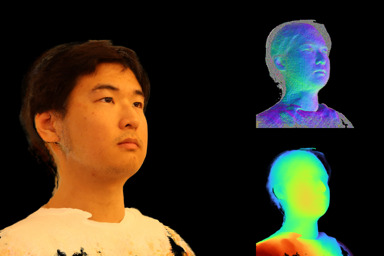} \\
\includegraphics[height=\height]{figures/inthewild/inthewild_0_2}
    & \includegraphics[height=\height]{figures/inthewild/inthewild_1_2}
    & \includegraphics[height=\height]{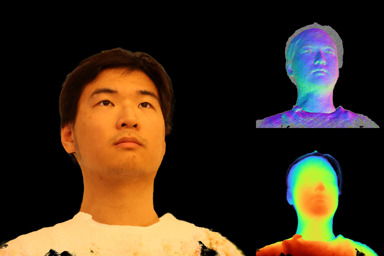} \\
\includegraphics[height=\height]{figures/inthewild/inthewild_0_3}
    & \includegraphics[height=\height]{figures/inthewild/inthewild_1_3}
    & \includegraphics[height=\height]{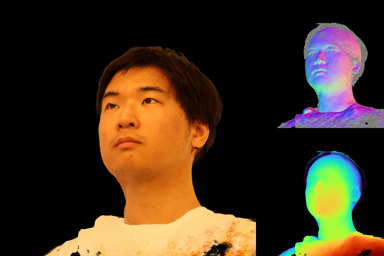} \\
\includegraphics[height=\height]{figures/inthewild/inthewild_0_4}
    & \includegraphics[height=\height]{figures/inthewild/inthewild_1_4}
    & \includegraphics[height=\height]{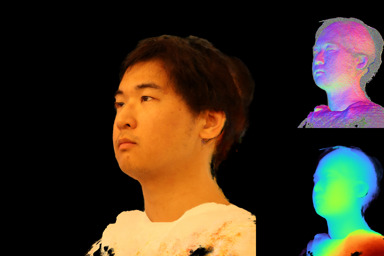}
\end{tabular}
    \caption{In-the-wild Results at Higher Resolution. We reconstruct a target identity from two images acquired with a consumer camera (left). Note how the novel views can extrapolate from the input camera angles.
    The inlays show the normals (top) and depth (bottom). The hair density is low, thus the grey normal colour in that region. We encourage the reader to see the supp. mat. for the high-resolution results and videos.
    \label{fig:inthewild}}
\end{center}
\end{figure*}

%% file: figures/ablation/ablation_regularization.tex
\begin{figure*}
\centering
\newcommand{\height}{5.5cm}
\begin{tabular}{ccccc}
\includegraphics[height=\height]{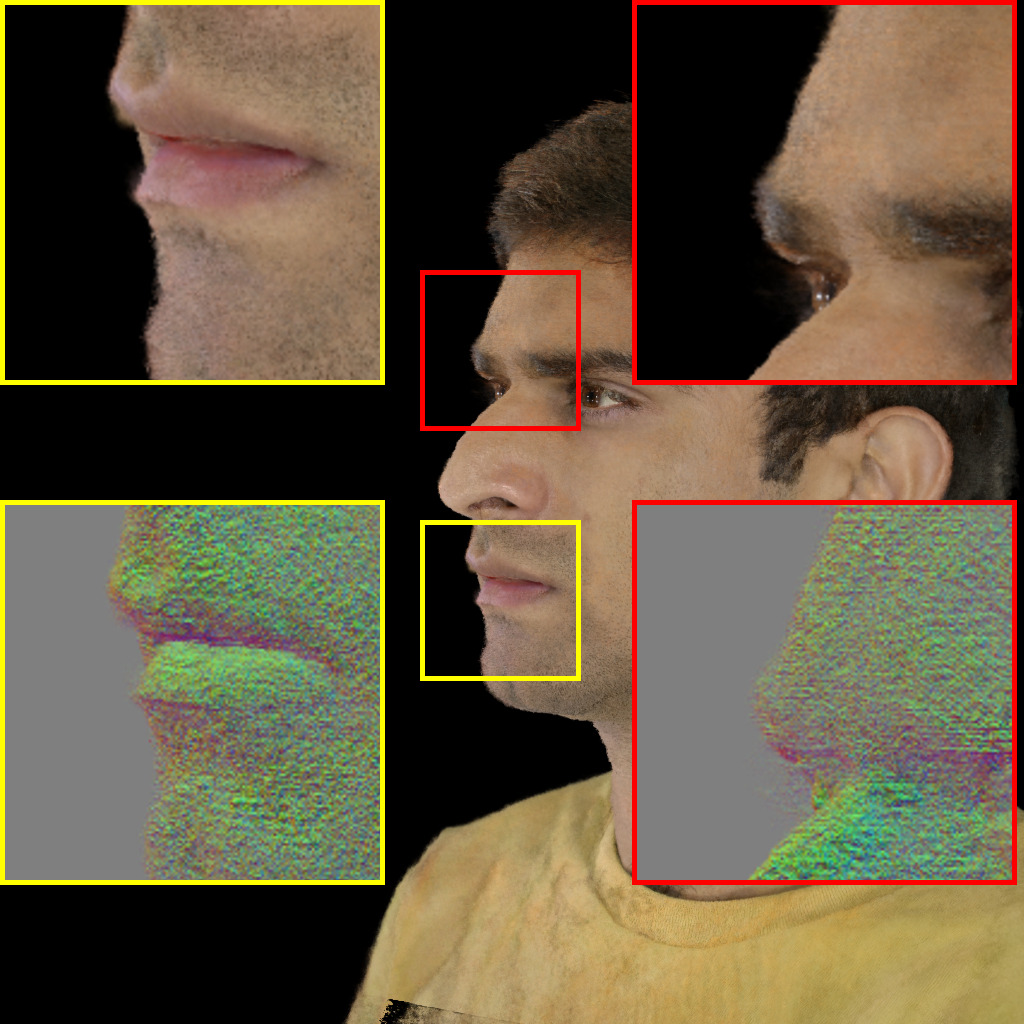} 
    & \includegraphics[height=\height]{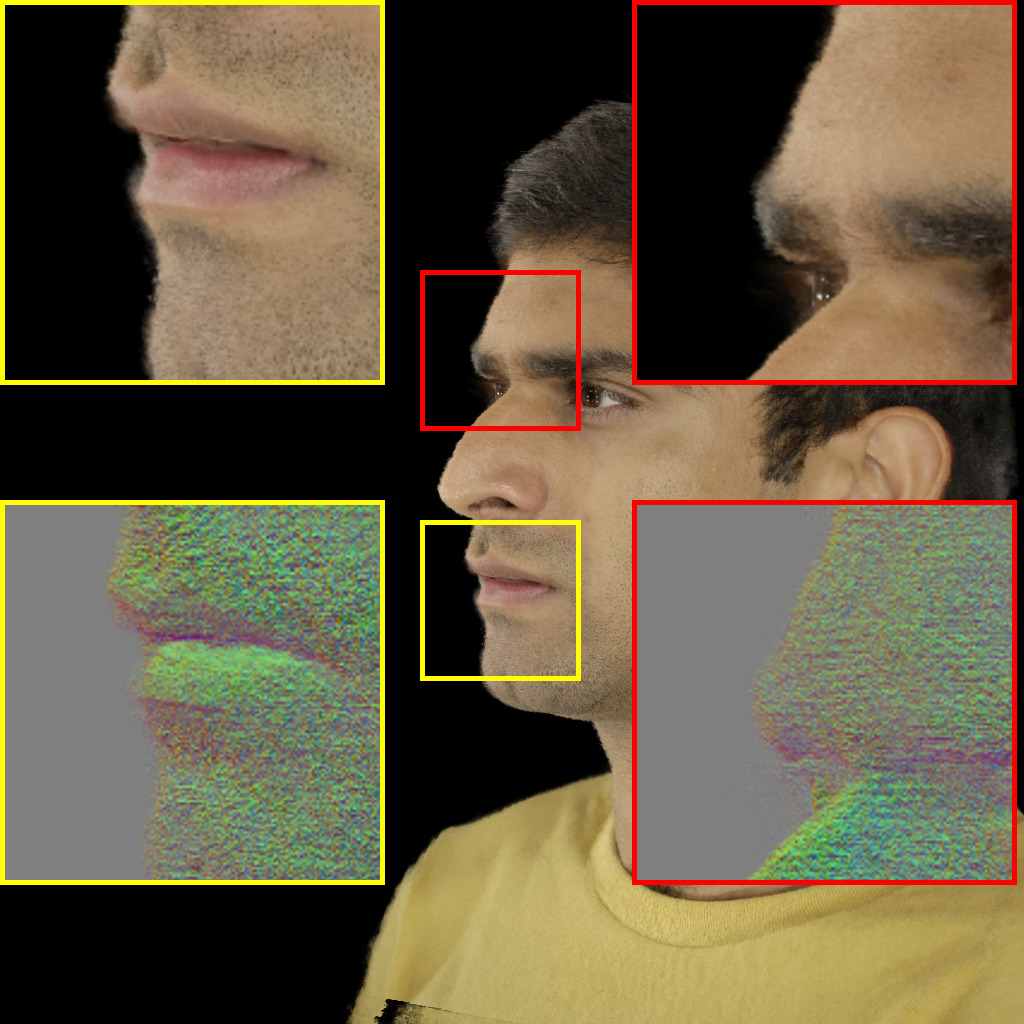}
    &  \includegraphics[height=\height]{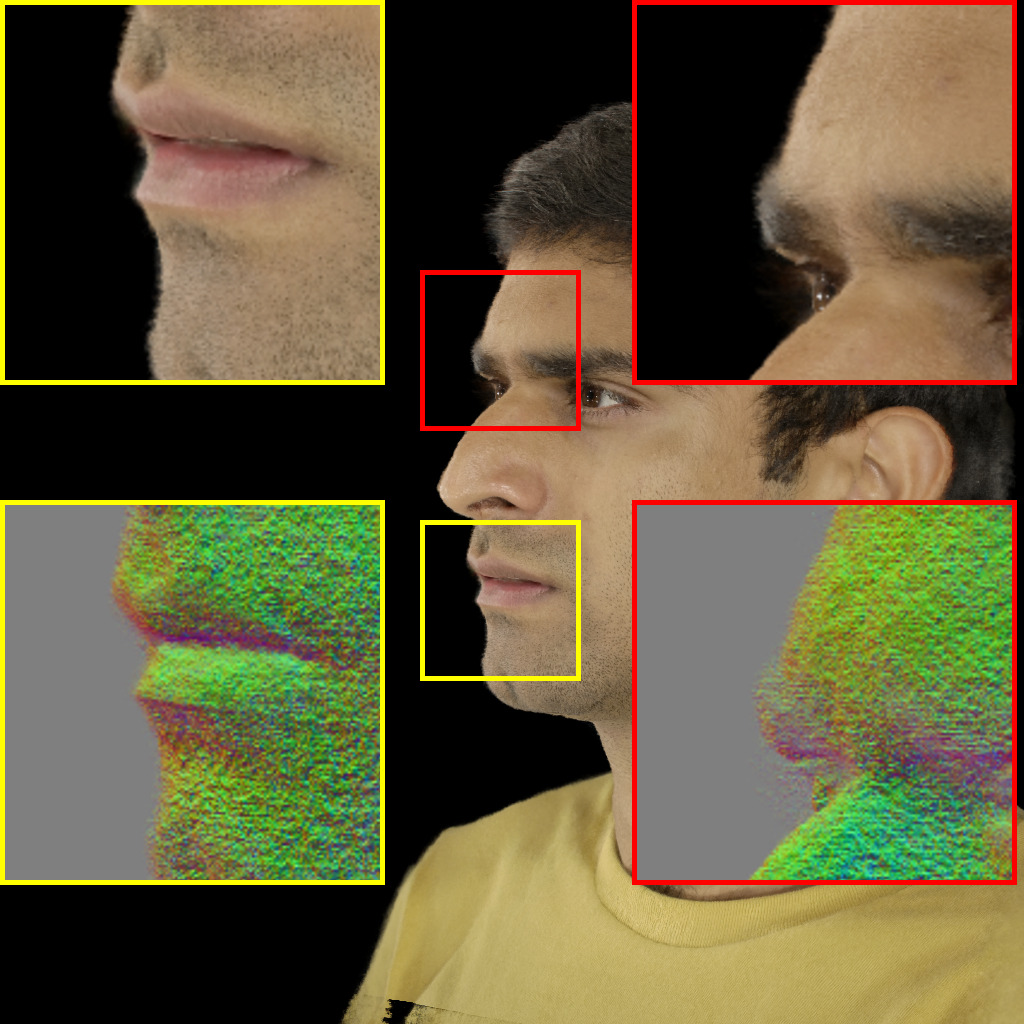}
    \\
\includegraphics[height=\height]{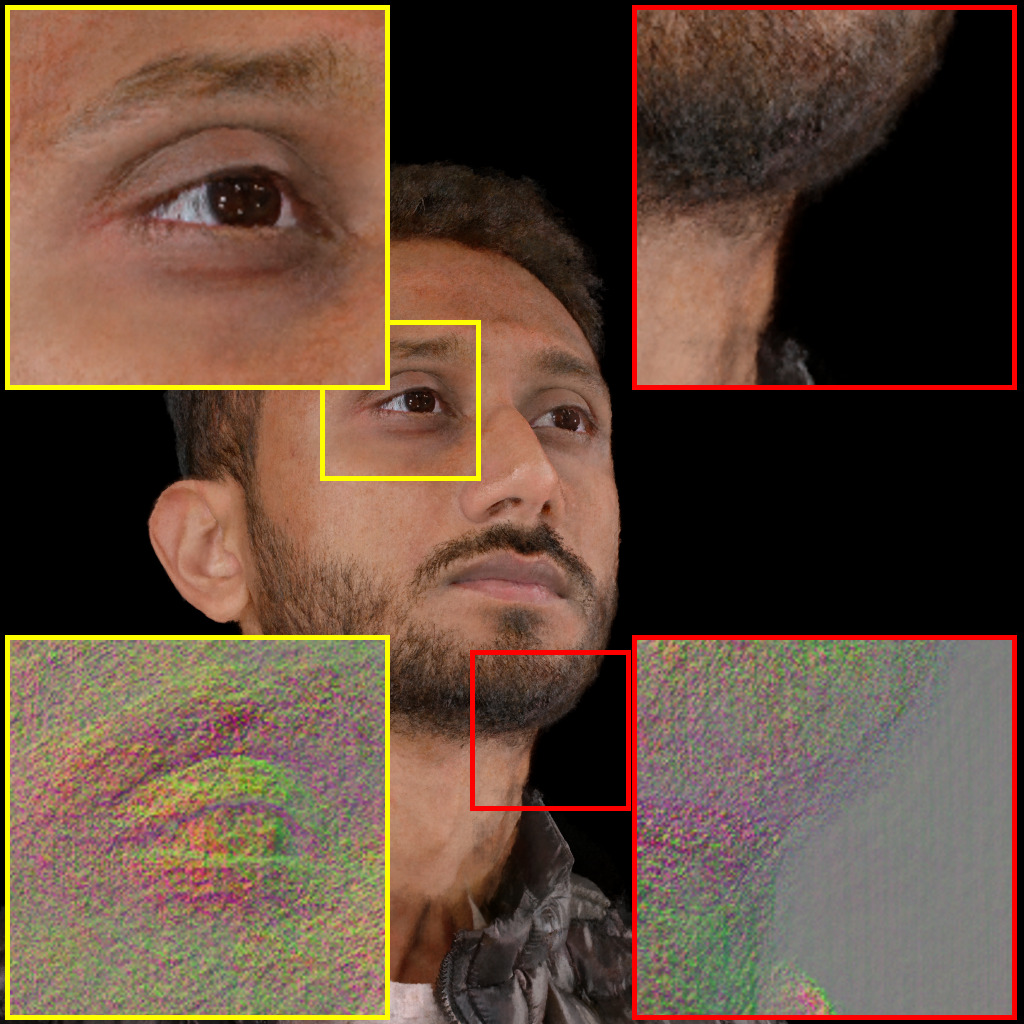} 
    & \includegraphics[height=\height]{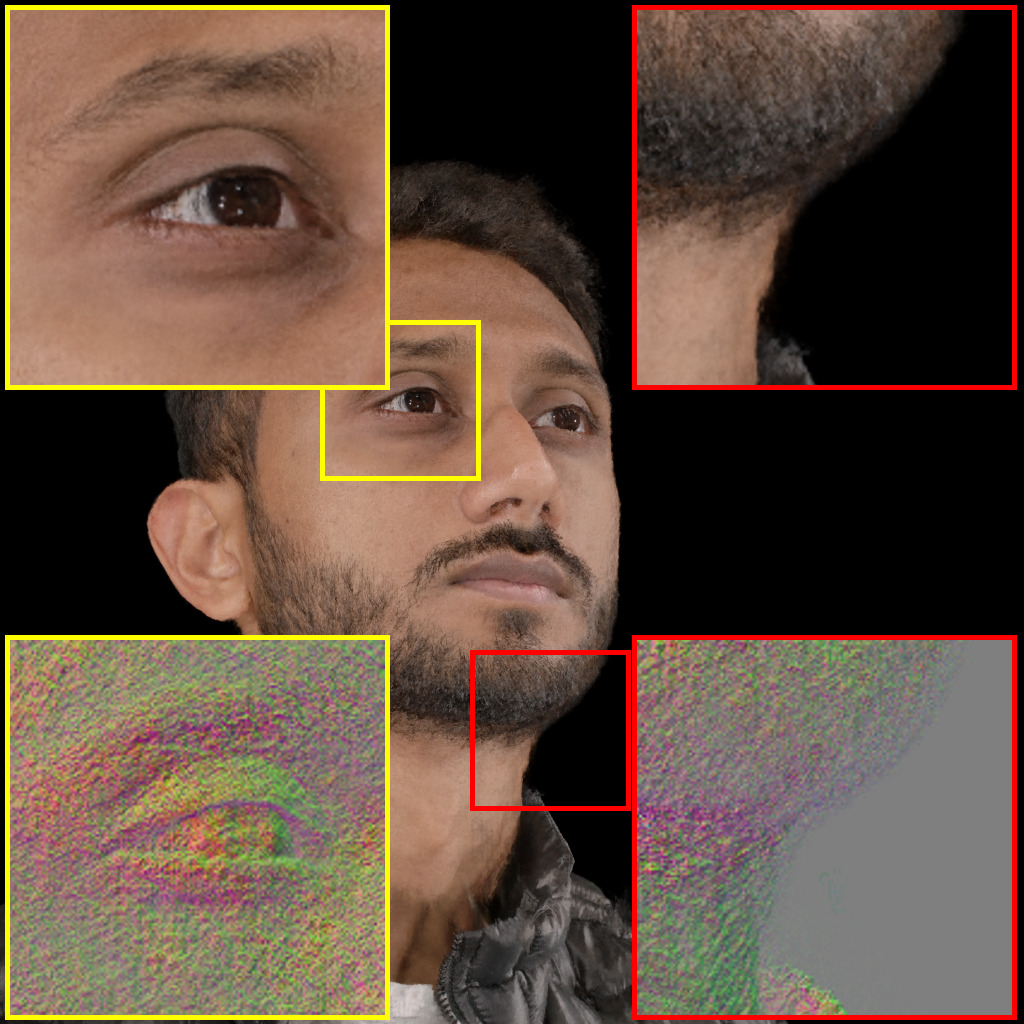}
    &  \includegraphics[height=\height]{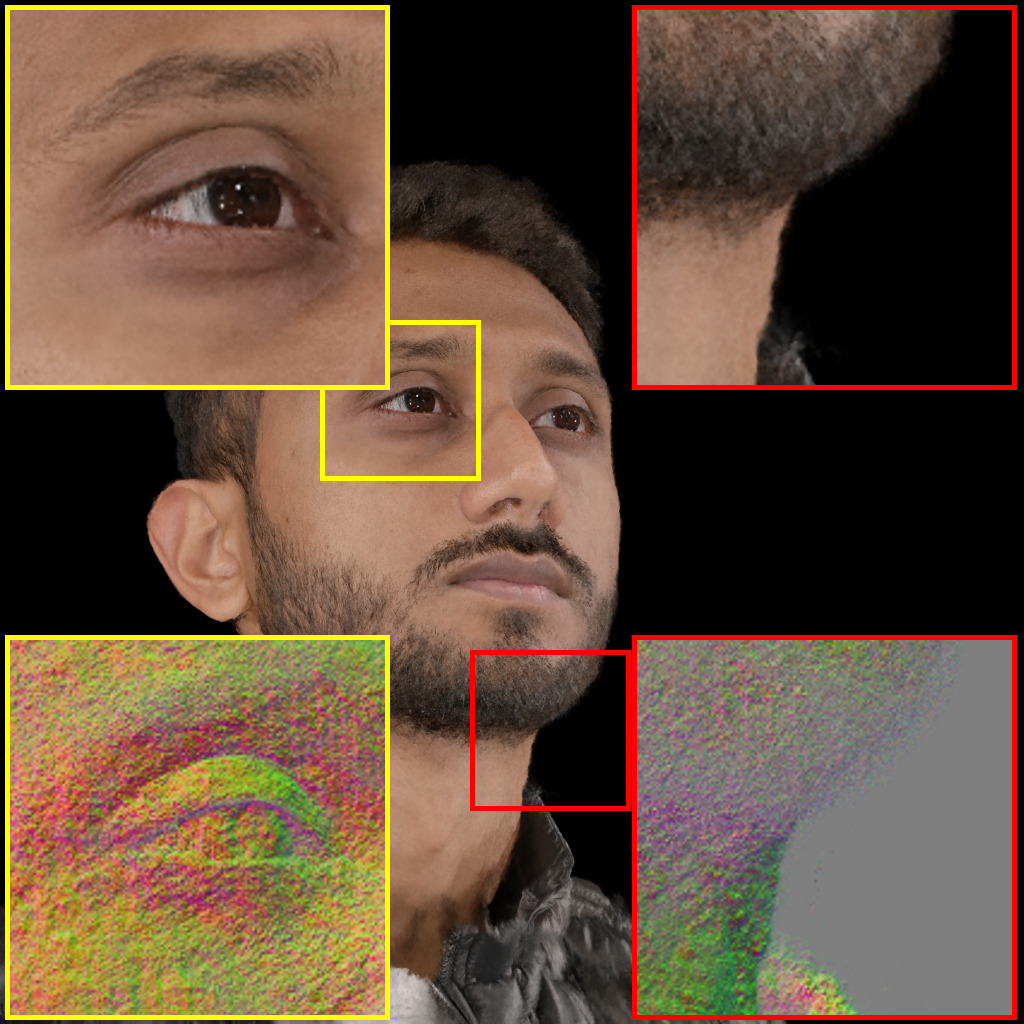}
    \\

$\mathcal{L}_{recon} + \mathcal{L}_{prop}$ & $+\;\mathcal{L}_v$ & $+ \;\mathcal{L}_\text{normal}$ \vspace{1em}
\end{tabular}
\caption{Visual results when applying regularisers. Training without regularisers ($\mathcal{L}_{recon} + \mathcal{L}_{prop}$, first column) leads to strong colour distortions for unseen views. Adding a regularisation loss on the model weights that process the view direction mitigates the colour distortions but yields fuzzy surfaces ($\mathcal{L}_v$, second column). Our final model employs an additional regulariser on predicted normals~\cite{verbin2022ref} to obtain well-defined surfaces ($\mathcal{L}_\text{normal}$, last column).
\label{fig:ablation_regularization} }
\end{figure*}

%% file: tables/ablation_regularization.tex
\begin{table}
\small 
\begin{center}
\begin{tabular}{cc | ccc}
\hline
\textbf{$\mathcal{L}_{v}$} &  \textbf{$\mathcal{L}_\text{normal}$} &
    \multicolumn{1}{c}{\textbf{PSNR} $\uparrow$} &
    \multicolumn{1}{c}{\textbf{SSIM} $\uparrow$} & 
    \multicolumn{1}{c}{\textbf{LPIPS} $\downarrow$}\\
\hline
 $\times$ & $\times$ &	23.91 &	0.7787	&0.2233\\
 $\times$ & $\checkmark$ & 24.79	&0.7839	&0.2066	\\
 $\checkmark$	& $\times$	&25.53	&0.7996	&0.1963	\\
 \hline
 $\checkmark$ & $\checkmark$ & \bf{25.69}&	\bf{0.8040} &	\bf{0.1905}
\end{tabular}
\end{center}
\caption{Ablation on regularisation when finetuning the model. The scores have been computed on models trained on two views with resolution $1024\times 1024$ and averaged across six views of three holdout subjects. Please refer to Fig.~\ref{fig:ablation_regularization} for visuals.}
\label{tab:ablation_regularization}
\end{table}

%% file: figures/ablation/ablation_frozen_latent.tex
\begin{figure*}
\centering
\newcommand{\height}{3.5cm}
\begin{tabular}{ccccc}
\rotatebox{90}{$\argmin_{\mathbf{\theta}_{\text{target}}}$} 
    & \includegraphics[height=\height]{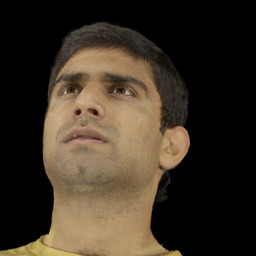} 
    & \includegraphics[height=\height]{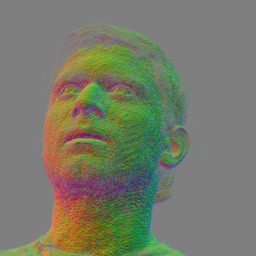}
    & \includegraphics[height=\height]{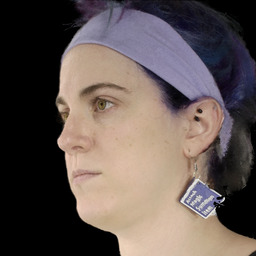} 
    & \includegraphics[height=\height]{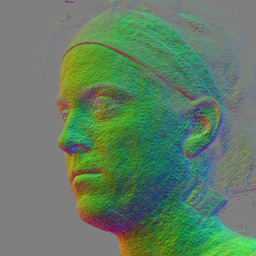}
    \\
\rotatebox{90}{$\argmin_{\mathbf{\theta}_{\text{target}}, \mathbf{w}_{\text{target}}}$ (Ours)} 
    & \includegraphics[height=\height]{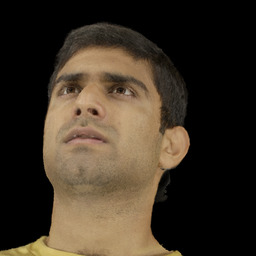} 
    & \includegraphics[height=\height]{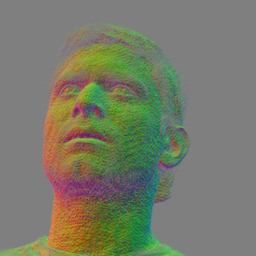}
    & \includegraphics[height=\height]{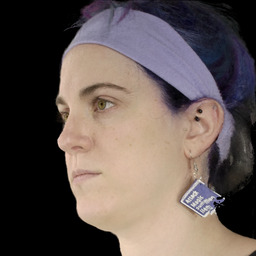} 
    & \includegraphics[height=\height]{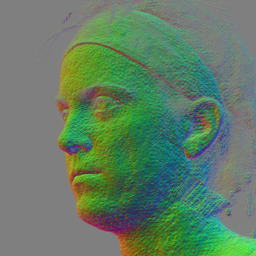}
    \\
& Novel view & Normals & Novel view & Normals
\end{tabular}
\caption{Effect of optimising only the model parameters $\mathbf{\theta}_{\text{target}}$ (top row) and optimising both the model parameters and the latent code $\mathbf{w}_{\text{target}}$ (bottom row, Ours). The visual results are very similar. Tbl.~\ref{tab:frozenlatent} lists quantitative metrics.\label{fig:ablation_frozen_latent}}
\end{figure*}

%% file: tables/ablation_frozenlatent.tex
\begin{table*}
\small 
\setlength{\tabcolsep}{3pt}
\begin{center}

\begin{tabular}{l ccc|ccc|ccc}
\hline
\textbf{Objective} & 
    \multicolumn{3}{c}{\textbf{PSNR} $\uparrow$} & 
    \multicolumn{3}{c}{\textbf{SSIM} $\uparrow$} & 
    \multicolumn{3}{c}{\textbf{LPIPS} $\downarrow$} \\
 Subject  & A & B & C & A & B & C & A & B & C \\
\hline
$\argmin_{\mathbf{\theta}_{\text{target}}}$ & 26.07 & 27.21	& 22.90	& 0.7949 & \textbf{0.8000}	& 0.7998 & \textbf{0.1823} & 0.1651 & 0.2126\\
$\argmin_{\mathbf{\theta}_{\text{target}}, \mathbf{w}_{\text{target}}}$ (Ours) & \textbf{26.55} & \textbf{27.30} & \textbf{23.22} & \textbf{0.8113}	& 0.7996 & \textbf{0.8009} & 0.1962	& \textbf{0.1650}	& \textbf{0.2102}	\\
\end{tabular}				
\end{center}
\caption{The model finetuning performs best when optimising both the model parameters $\Theta_{target}$ and the latent code $\bf{w}_{target}$. 
All metrics were computed after finetuning to two views at 1K resolution. Visually, the optimisation results look very similar, see Fig.~\ref{fig:ablation_frozen_latent}.\label{tab:frozenlatent}
}
\end{table*}

%% file: figures/ablation/ablation_initialization.tex
\begin{figure*}
\centering
\newcommand{\height}{2.5cm}
\begin{tabular}{cccccc}
\includegraphics[height=\height]{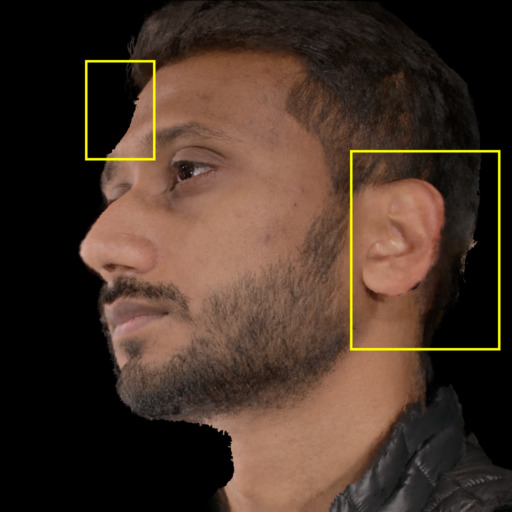} &
\includegraphics[height=\height]{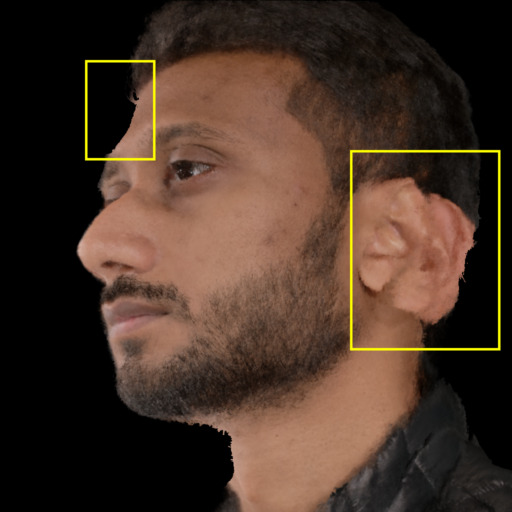} &
\includegraphics[height=\height]{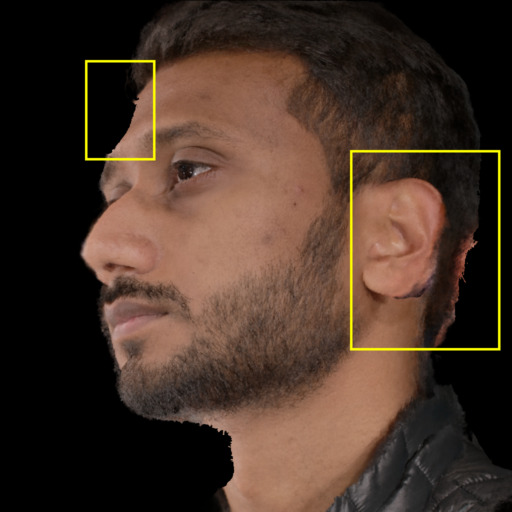} &
\includegraphics[height=\height]{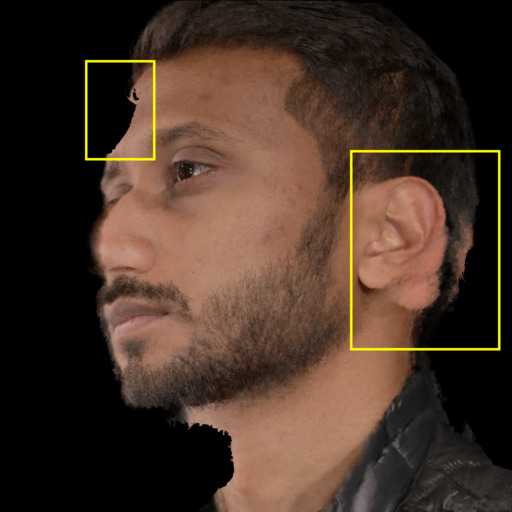} &
\includegraphics[height=\height]{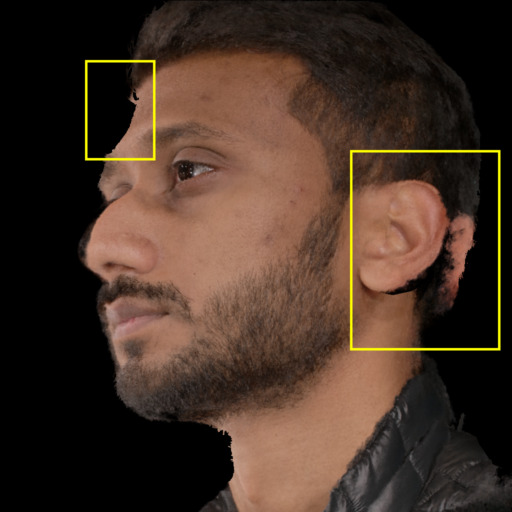} &
\includegraphics[height=\height]{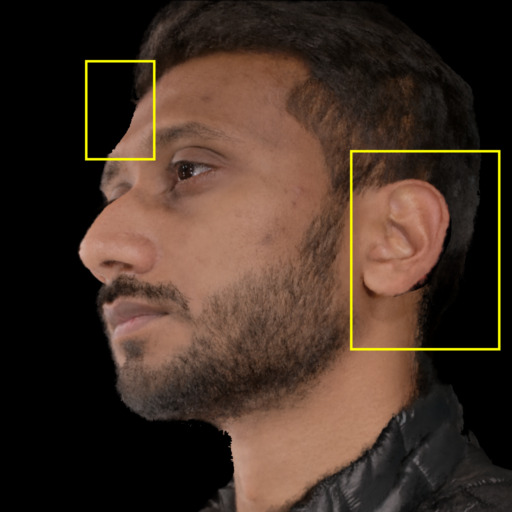} \\
\includegraphics[height=\height]{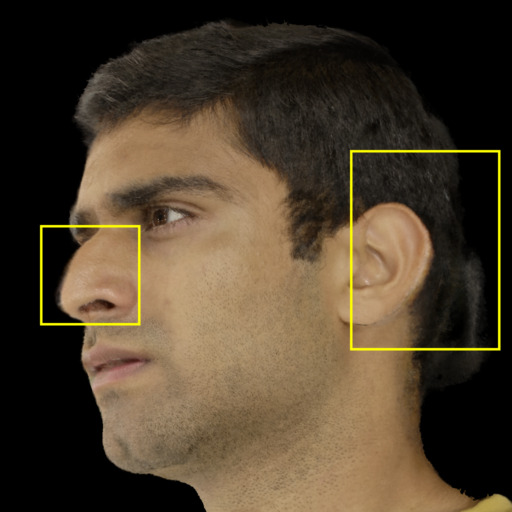} &
\includegraphics[height=\height]{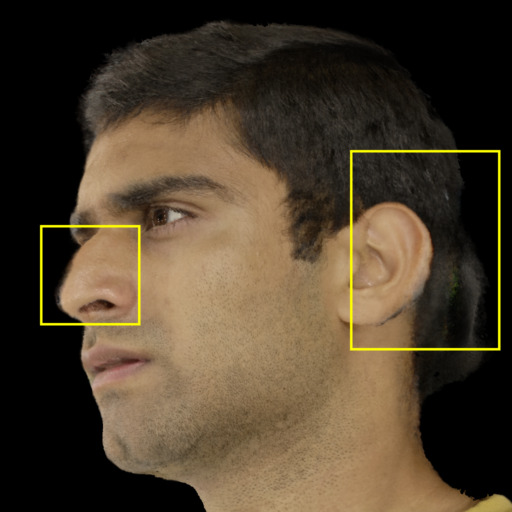} &
\includegraphics[height=\height]{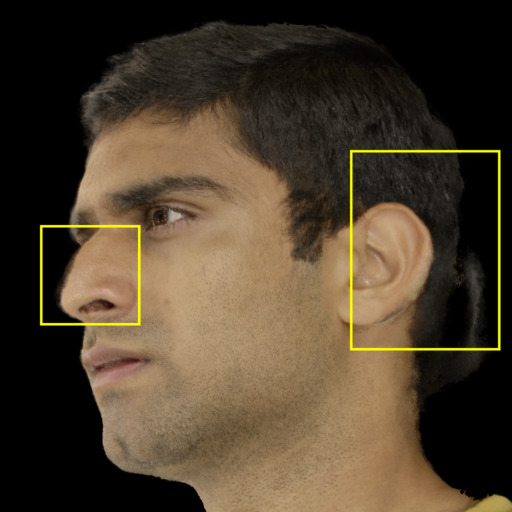} &
\includegraphics[height=\height]{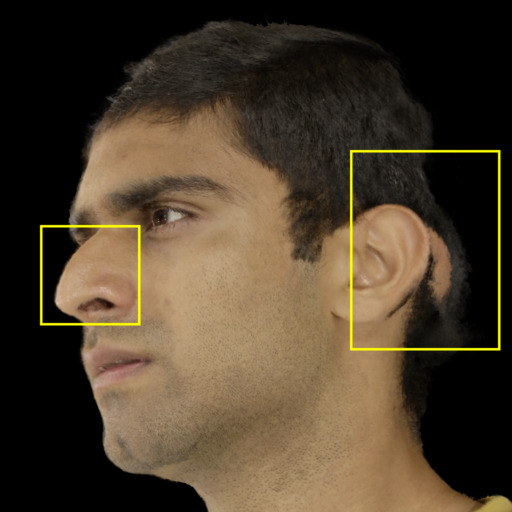} &
\includegraphics[height=\height]{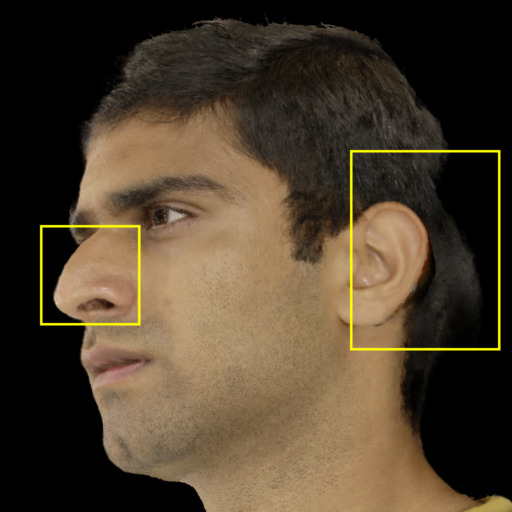} &
\includegraphics[height=\height]{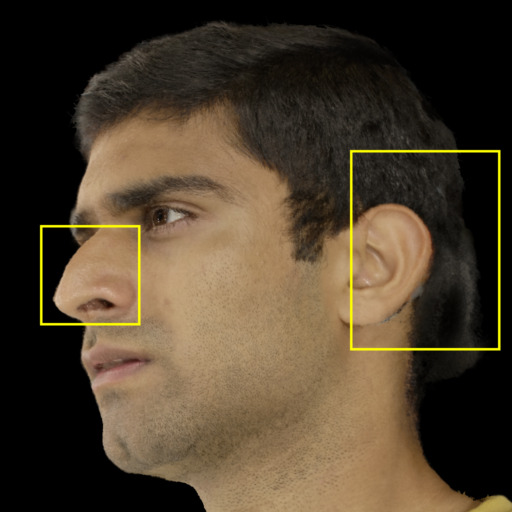} \\
Mean & Noise & Zeros & Furthest & Nearest & Inversion 
\end{tabular}
\caption{\label{fig:ablation_initialization} Visual comparison of different initialisation techniques.
When the geometry is not initialised correctly at the start of finetuning, the final result can contain artifacts like a second ear, an unrealistic forehead, and a fuzzy surface.
Starting from the inversion result mitigates these artifacts. Please see the text for an explanation of the different initialisation techniques and Tbl.~\ref{tab:ablation_initialization} for metrics.}
\end{figure*}

%% file: tables/ablation_initialization.tex
\begin{table*}
\small 
\setlength{\tabcolsep}{3pt}
\begin{center}

\begin{tabular}{l ccc|ccc|ccc}
\hline
\textbf{Initialisation} & 
    \multicolumn{3}{c}{\textbf{PSNR} $\uparrow$} & 
    \multicolumn{3}{c}{\textbf{SSIM} $\uparrow$} & 
    \multicolumn{3}{c}{\textbf{LPIPS} $\downarrow$} \\
 Subject  & A & B & C & A & B & C & A & B & C \\
\hline
Mean & 25.39 & 26.44 & 22.00 & 0.7963 & 0.7913 & 0.7927 & 0.1917 & 0.1749 & 0.2210 \\
Noise & 25.21 & 26.32 & 22.44 & 0.7993 & 0.7911 & 0.7966 & 0.206 & 0.1766 & 0.2169 \\
Zeros & 25.32 & 26.37 & 22.25 & 0.7956 & 0.7927 & 0.7939 & 0.1917 & 0.1732 & 0.2183 \\
Furthest & 24.07 & 25.57 & 22.09 & 0.7884 & 0.7829 & 0.7915 & 0.1997 & 0.1875 & 0.2250 \\
Nearest & 25.49 & 25.68 & 22.05 & 0.7934 & 0.7818 & 0.7948 & \textbf{0.1915} & 0.1852 & 0.2240 \\
\hline
Inversion (\textbf{Ours}) & \textbf{26.55} & \textbf{27.30} & \textbf{23.22} & \textbf{0.8113} & \textbf{0.7996} & \textbf{0.8009} & 0.1962 & \textbf{0.1650} & \textbf{0.2102}\\
\end{tabular}				
\end{center}
\caption{Ablation on initialisation strategies for $\mathbf{w}_{\text{target}}$ for finetuning. This table lists metrics computed on face crops of 6 holdout views at resolution $1024\times 1024$. 
\emph{Furthest} (\emph{nearest}) indicate initialising the latent code with the least (most) similar training subject.
Figure~\ref{fig:ablation_initialization} shows visuals examples.
\label{tab:ablation_initialization}}
\end{table*}

%% file: figures/ablation/ablation_expression.tex
\begin{figure}

\begin{center}
\small
\setlength{\tabcolsep}{2pt}

\newcommand{\height}{2.3cm}
\newcommand{\smallheight}{0.53cm}
\begin{tabular}{cccccc}
\begin{tabular}{cccccc}
\includegraphics[height=\smallheight]{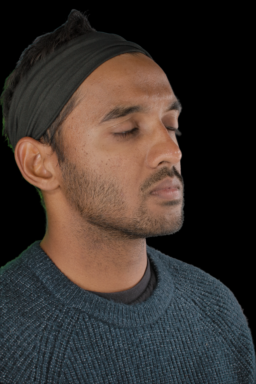} 
    \includegraphics[height=\smallheight]{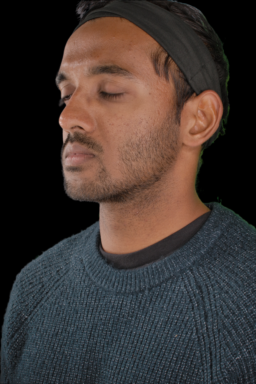} \\
\includegraphics[height=\smallheight]{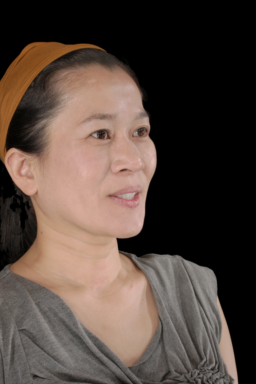} 
    \includegraphics[height=\smallheight]{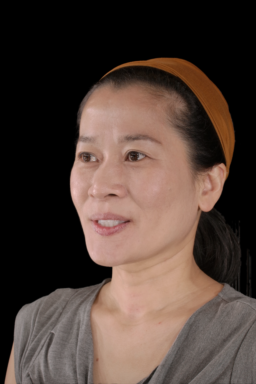} \\
\includegraphics[height=\smallheight]{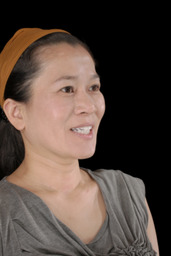} 
    \includegraphics[height=\smallheight]{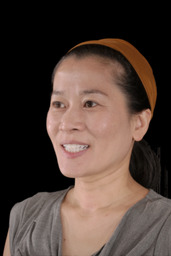} \\
Inputs\end{tabular}

\begin{tabular}{ccc}
\includegraphics[height=\height]{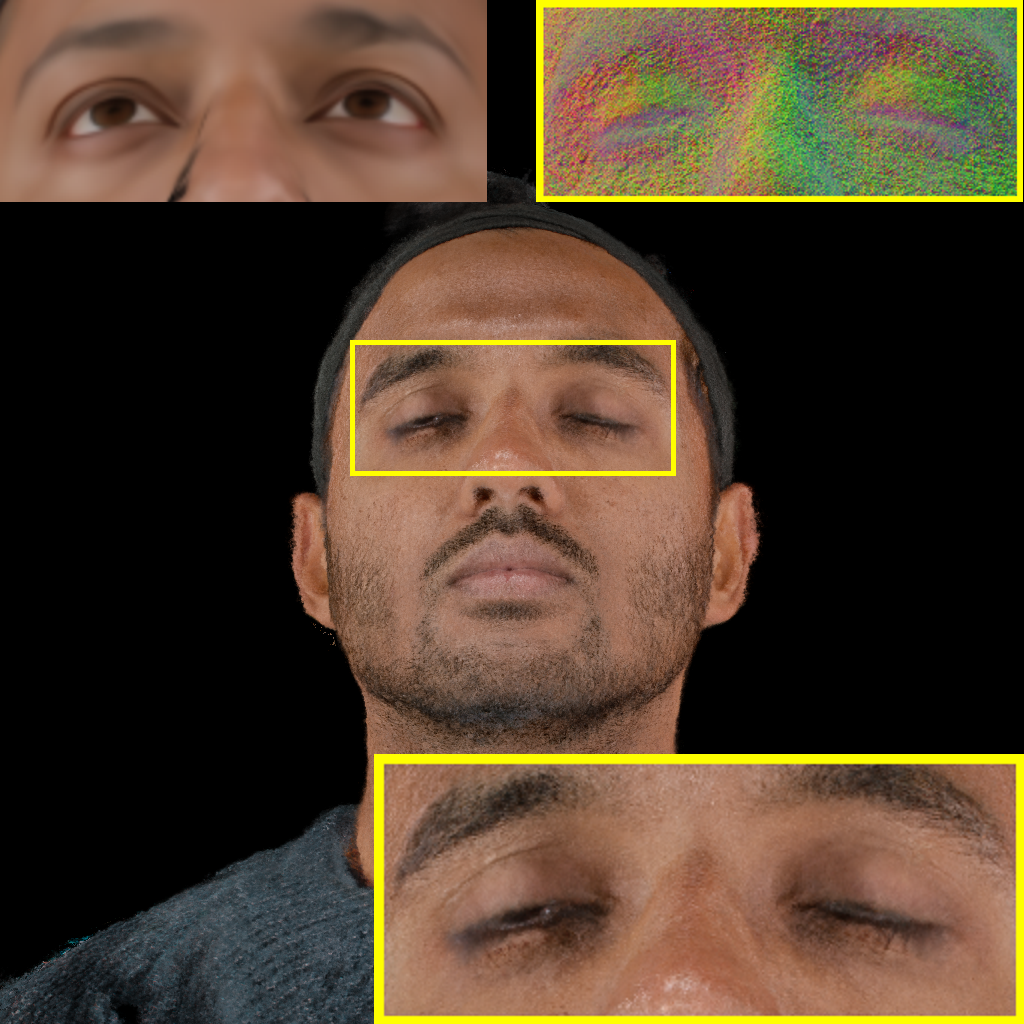} &
\includegraphics[height=\height]{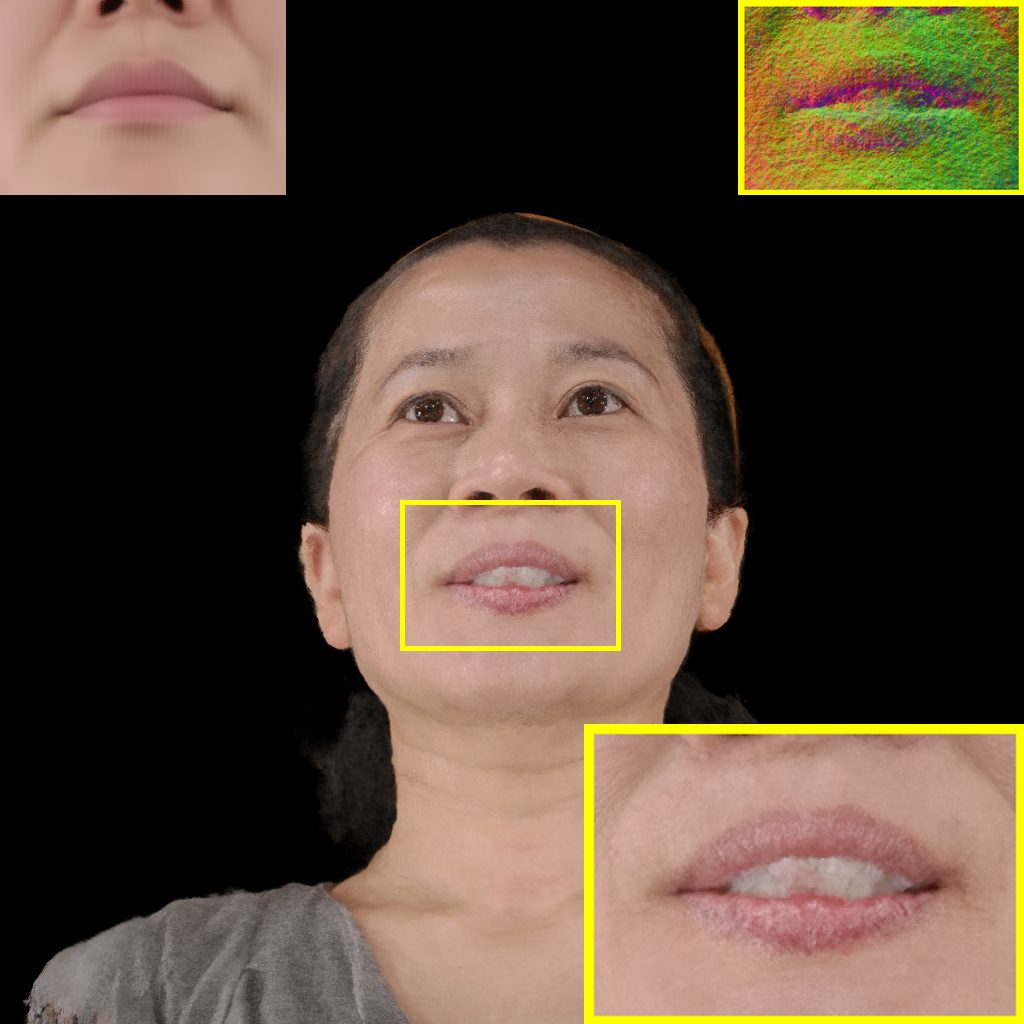} &
\includegraphics[height=\height]{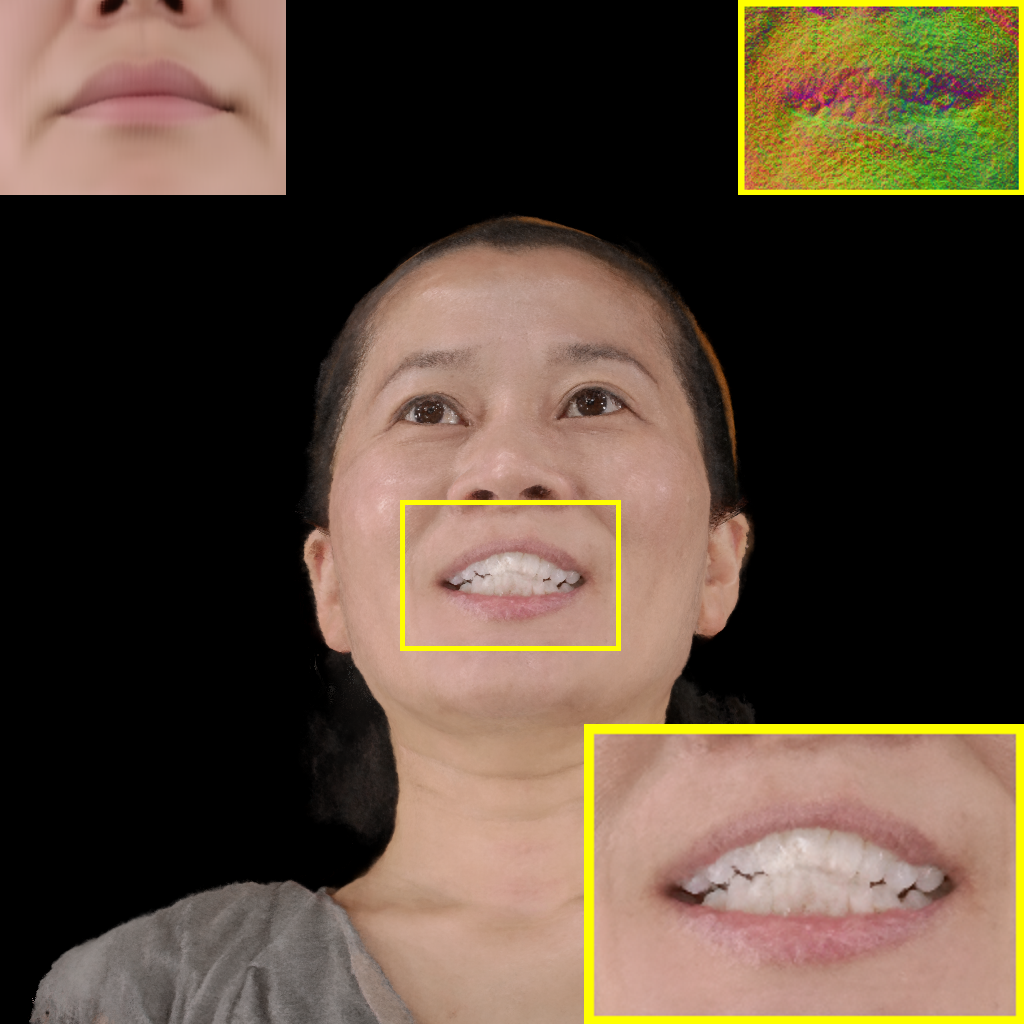}

\end{tabular}
\end{tabular}
\end{center}
\caption{\label{fig:ablation_expression}
Out-of-distribution facial expressions. Our model was trained on neutral faces with a closed mouth. It can handle mild expressions but fails for strong expressions and teeth.
We show a novel view with insets of the inversion result (top-left), normals (top-right), and a zoom-in patch (bottom-right).
}

\end{figure}

%% file: tables/ablation_views.tex
\begin{table}
\small 
\begin{center}
\begin{tabular}{c | ccc}
\hline
\textbf{\# Views} & 
    \multicolumn{1}{c}{\textbf{PSNR} $\uparrow$} &
    \multicolumn{1}{c}{\textbf{SSIM} $\uparrow$} & 
    \multicolumn{1}{c}{\textbf{LPIPS} $\downarrow$}\\
\hline
1 &	23.37 &	0.7658 &	0.2189 \\
2 &	25.69 &	0.8040	& 0.1905 \\
3 &	27.16 &	0.8275 &	0.1675 \\
5 &	28.33&	0.8445&	0.1651 \\
7 &	29.24	&0.8600	& 0.1539	
\end{tabular}
\end{center}
\caption{Ablation on the performance for different number of views when  finetuning the model. The scores are computed on models trained on images with resolution $1024\times 1024$.}
\label{tab:ablation_views}
\end{table}

%% file: tables/comparison256.tex
\begin{table*}[t]
\small 
\setlength{\tabcolsep}{3pt}
\begin{center}

\begin{tabular}{r ccc|ccc|ccc}
\hline
\textbf{Method} & 
    \multicolumn{3}{c}{\textbf{PSNR} $\uparrow$} & 
    \multicolumn{3}{c}{\textbf{SSIM} $\uparrow$} & 
    \multicolumn{3}{c}{\textbf{LPIPS} $\downarrow$} \\
Subject  & A & B & C & A & B & C & A & B & C \\
\hline
KeypointNeRF~\cite{keypointnerf} &
24.47 & 23.42 & 20.33 & 0.7887 & 0.7736 & 0.7387 & 0.1866 & 0.1991 & 0.2462\\
\hline
\textbf{Ours} & \textbf{28.31} & \textbf{29.00} & \textbf{23.92 }&	\textbf{0.8703} & \textbf{0.8814} & \textbf{0.8321} &	\textbf{0.1025} & \textbf{0.0937} & \textbf{0.1484}\\
\end{tabular}				
\end{center}
\caption{Comparison with KeypointNeRF~\cite{keypointnerf} on our dataset. Despite considerable efforts, their implementation did not produce high-quality results at 1K resolution, hence, we compare on resolution $256\times256$. Please refer to Fig.~\ref{fig:comparison256} for visuals and Tbl.~\ref{tab:comp_holobooth} for results at 1K resolution.
\label{tab:comparison256}
}
\end{table*}

%% file: tables/comparison_holobooth_views.tex
\begin{table*}
\small 
\setlength{\tabcolsep}{3pt}
\begin{center}

\begin{tabular}{c r ccc|ccc|ccc}
\hline
 & \textbf{Method} & 
    \multicolumn{3}{c}{\textbf{PSNR} $\uparrow$} & 
    \multicolumn{3}{c}{\textbf{SSIM} $\uparrow$} & 
    \multicolumn{3}{c}{\textbf{LPIPS} $\downarrow$} \\
\# Views & Subject  & A & B & C & A & B & C & A & B & C \\
\hline
\multirow{6}{*}{2} &
Learnit & 22.07 & 21.18 & 16.86 & 0.7870 & 0.7765 & 0.7513 & 0.3068 & 0.3195 & 0.3635 \\
& EG3D-based prior & 20.25 & 20.60 & 18.24 & 0.7633 & 0.7575 & 0.7556 & 0.2678 & 0.2853 & 0.3159 \\
& RegNeRF & 20.63 & 19.93 & 20.63 &	0.7468 & 0.7361 & 0.7468 & 	0.2791 & 0.2993	 &0.2791 \\
& FreeNeRF& 17.24 & 14.48 & 13.35 & 0.7091 & 0.6619 & 0.6675 & 0.2711 & 0.3140 & 0.3428 \\
& KeypointNeRF & 23.80 & 23.45 & 21.11 & 0.7964 & 0.7832 & 0.7838 & 0.2542 & 0.2628 & 0.2969 \\
&\textbf{Ours} & \textbf{26.55} & \textbf{27.30} & \textbf{23.22} & \textbf{0.8113}	& 	\textbf{0.7996} & \textbf{0.8009} & \textbf{0.1962}	& \textbf{0.1650}	& \textbf{0.2102}	\\
\hline
\multirow{5}{*}{3}
& Learnit & 22.99 & 22.53 & 19.15 & 0.7939 & 0.7847 & 0.7775 & 0.2981 & 0.3031 & 0.3473 \\
& EG3D-based prior & 22.26 & 21.91 & 19.60 & 0.7902 & 0.7781 & 0.7823 & 0.2649 & 0.2819 & 0.3057 \\
& RegNeRF & 22.62 & 23.12 & 20.26 & 0.7794  & 0.7654 & 0.7714 & 0.2654 & 0.2768 & 0.3043\\
& FreeNeRF & 24.71 & 21.74 & 21.52 & 0.7962 & 0.7582 & 0.7757 & 0.2150 & 0.2314 & 0.2622 \\
& KeypointNeRF & 24.62 & 24.52 & 22.19 & 0.8013 & 0.7904 & 0.7913 & 0.2364 & 0.2449 & 0.2751 \\
&\textbf{Ours} & \textbf{27.89} & \textbf{28.86} & \textbf{24.72} & \textbf{0.8268} & \textbf{0.8305} & \textbf{0.8252} & \textbf{0.1633} & \textbf{0.1498} & \textbf{0.1893} \\
\hline
\multirow{5}{*}{5} &
Learnit & 23.03 & 23.01 & 18.54 & 0.7935 & 0.7874 & 0.7742 & 0.2991 & 0.3011 & 0.3494 \\
& EG3D-based prior & 20.16 & 21.32 & 19.13 & 0.7938 & 0.7832 & 0.7783 & 0.2694 & 0.2829 & 0.3137 \\
& RegNeRF &24.85 & 23.56 & 20.93 & 0.7944 & 0.7787 & 0.7908 & 0.2611 & 0.2753 & 	0.2919	\\
& FreeNeRF & 28.10 & 27.37 & 24.14 & 0.8291 & 0.8217 & 0.8274 & 0.1760 & 0.2022 & 0.2245  \\
& KeypointNeRF & 24.38 & 24.29 & 22.29 & 0.7969 & 0.7867 & 0.7864 & 0.2388 & 0.2434 & 0.2743 \\
&\textbf{Ours} & \textbf{29.55} & \textbf{29.27} & \textbf{26.17} & \textbf{0.8466} & \textbf{0.8452} & \textbf{0.8417} & \textbf{0.1560} & \textbf{0.1483} & \textbf{0.1910}\\
\hline
\multirow{5}{*}{7} & 
Learnit & 23.60 & 23.10 & 18.31 & 0.7984 & 0.7887 & 0.7659 & 0.2961 & 0.3000 & 0.3506 \\
& EG3D-based prior & 20.05 & 21.26 & 19.45 & 0.7991 & 0.7890 & 0.7890 & 0.2690 & 0.2815 & 0.3130 \\
& RegNeRF & 27.73 & 26.36 & 24.55 & 0.8229 & 	0.8055 & 0.8225	 & 0.2437 & 0.2589 & 0.2671		\\
& FreeNeRF & 28.09 & 25.03 & 20.03 & 0.8392 & 0.8027 & 0.7936 & 0.1704 & 0.2292 & 0.2458 \\
& KeypointNeRF & 23.84 & 23.97 & 22.11 & 0.7902 & 0.7811 & 0.7793 & 0.2430 & 0.2477 & 0.2829 \\
& \textbf{Ours} & \textbf{29.54} & \textbf{30.42} & \textbf{27.76} & \textbf{0.8564} & \textbf{0.8639} & \textbf{0.8598} & \textbf{0.1510} & \textbf{0.1353} & \textbf{0.1755}\\
\end{tabular}				
\end{center}
\caption{Comparison with related works at 1K resolution on our studio dataset. We compare with Learnit~\cite{tancik2021learned}, EG3D-based prior~\cite{eg3d}, RegNeRF~\cite{regnerf}, FreeNeRF~\cite{Yang2023FreeNeRF}, and KeypointNeRF~\cite{keypointnerf} on different number of input views ranging from two to seven. Our method outperforms the related works by a clear margin. For a visual comparison, please refer to Figures \ref{fig:comparison_1k_suba},\ref{fig:comparison_1k_subb}, and \ref{fig:comparison_1k_subc}.
\label{tab:comp_holobooth}}
\end{table*}

%% file: tables/comparison_facescape.tex
\begin{table*}
\small 
\setlength{\tabcolsep}{3pt}
\begin{center}

\begin{tabular}{r cccc|cccc|cccc}
\hline
\textbf{Method} & 
    \multicolumn{4}{c}{\textbf{PSNR} $\uparrow$} & 
    \multicolumn{4}{c}{\textbf{SSIM} $\uparrow$} & 
    \multicolumn{4}{c}{\textbf{LPIPS} $\downarrow$} \\
Subject & 122 & 212 & 340 & 344 & 122 & 212 & 340 & 344 & 122 & 212 & 340 & 344 \\
\hline
 EG3D-based prior~\cite{chan2022efficient}&	23.27	& 26.15	& 22.68 &	24.54&	0.8678&	0.9030	&0.8862&	0.8844&	0.1504&	0.1281&	0.1228&	0.1357\\
 KeypointNeRF~\cite{keypointnerf}&23.46	 &24.59 &	23.53 &	22.10 &	0.9171	 &0.9372	 &0.9187	 &0.9025 & 0.0940	 &0.0681	 &0.0743	 &0.0919\\
 RegNeRF~\cite{regnerf}& 24.77 & 28.97 & 24.95 & 25.60 & 0.8903 & 0.9390 & 0.9129 & 0.8908 & 0.1334 & 0.0892 & 0.1001 & 0.1232\\
 DINER~\cite{diner}& 25.79 & 29.78 & 26.27 & \textbf{26.45} & 0.9382 &  0.9597 & 0.9434 & \textbf{0.9324} & 0.0672 & 0.0672 & 0.0540 & \textbf{0.0677}\\
\hline
\textbf{Ours}& \textbf{27.40} & \textbf{32.03}	 & \textbf{26.69}	 & 25.51  & \textbf{0.9359}  & 	\textbf{0.9721}  & 	\textbf{0.9489}  & 	0.9135 & 	\textbf{0.0671}  & 	\textbf{0.0355}	 & \textbf{0.0533} & 	0.0761\\
\end{tabular}
\end{center}
\caption{Comparison with the state-of-the-art for novel view synthesis from sparse views on Facescape~\cite{facescape}.
This table supplements the main paper with individual metrics for each of the four test subjects. For a visual comparison, please refer to Fig.~\ref{fig:comparison_facescape_full}.\label{tab:facescape}}
\end{table*}

%% file: tables/ablation_prior.tex
\begin{table}
\small 
\begin{center}
\begin{tabular}{rc | ccc}
\hline
\textbf{\# Identities} & \textbf{Resolution} &
    \multicolumn{1}{c}{\textbf{PSNR} $\uparrow$} &
    \multicolumn{1}{c}{\textbf{SSIM} $\uparrow$} & 
    \multicolumn{1}{c}{\textbf{LPIPS} $\downarrow$}\\
\hline
 15& $512\times 768$ & 24.25 &	0.7917 &	0.2187  \\
 350 & $512\times 768$ & 24.62	& 0.7926	 & 0.1985 \\
 750 & $512\times 768$ & 25.43 & 0.7935 & 0.2035 \\
 1450 & $256\times 384$ &\bf{25.99} &	0.8034 &	\bf{0.1810}\\ 
 1450 & $512\times 768$  & 25.69&	\bf{0.8040} &	0.1905
\end{tabular}
\end{center}
\caption{Ablation on the prior model. We train variants of our prior model at a lower resolution and with fewer identities. The metrics are computed after finetuning to two views at resolution $1024\times 1024$.}
\label{tab:ablation_prior}
\end{table}

%% file: figures/comparison/comparison_holobooth_subA.tex
\begin{figure*}[ht]

\begin{center}
\small
\setlength{\tabcolsep}{2pt}

\newcommand{\height}{2.8cm}
\begin{tabular}{cccccc}
& \textbf{Subject A}\\
\rotatebox{90}{GT}  & \includegraphics[height=\height]{figures/comparison/holobooth/gt_0_gt}  \\
\\
\rotatebox{90}{EG3D}
  & \includegraphics[height=\height]
 {figures/comparison/holobooth/eg3d_in2_0_pred}  
    & \includegraphics[height=\height]{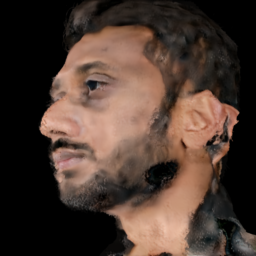} 
    & \includegraphics[height=\height]{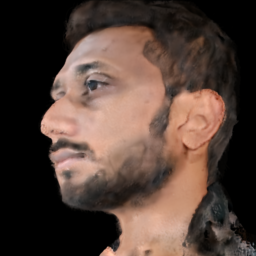} 
    & \includegraphics[height=\height]{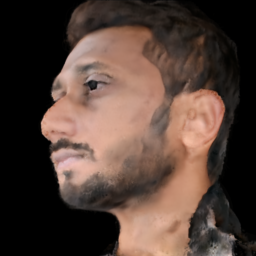} 
    \\    
\rotatebox{90}{Learnit}
  & \includegraphics[height=\height]
 {figures/comparison/holobooth/learnit_in2_0_pred}  
    & \includegraphics[height=\height]{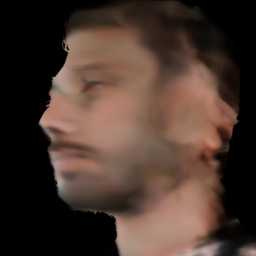} 
    & \includegraphics[height=\height]{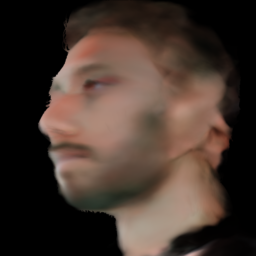} 
    & \includegraphics[height=\height]{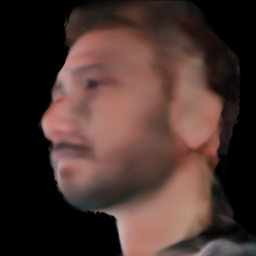} 
    \\    
\rotatebox{90}{KeypointNeRF}
  & \includegraphics[height=\height]
 {figures/comparison/holobooth/keypointnerf_in2_0_pred}  
    & \includegraphics[height=\height]{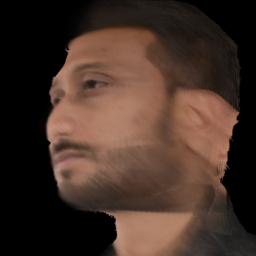} 
    & \includegraphics[height=\height]{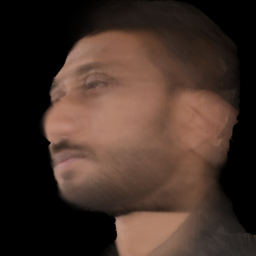} 
    & \includegraphics[height=\height]{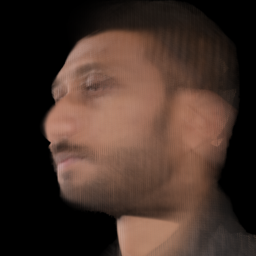} 
    \\    
\rotatebox{90}{RegNeRF}
  & \includegraphics[height=\height]
 {figures/comparison/holobooth/regnerf_in2_0_pred} 
    & \includegraphics[height=\height]{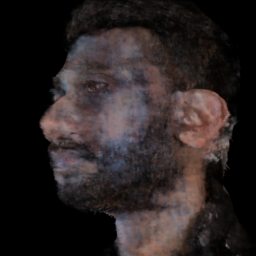} 
    & \includegraphics[height=\height]{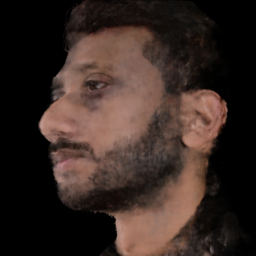} 
    & \includegraphics[height=\height]{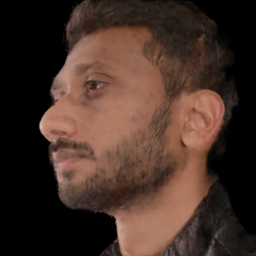} 
    \\    
\rotatebox{90}{FreeNeRF}
  & \includegraphics[height=\height]
 {figures/comparison/holobooth/freenerf_in2_0_pred} 
    & \includegraphics[height=\height]{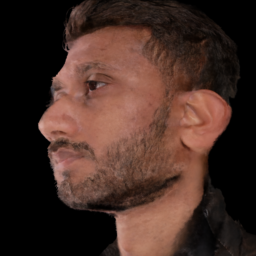} 
    & \includegraphics[height=\height]{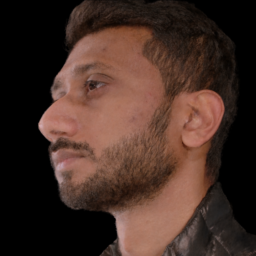} 
    & \includegraphics[height=\height]{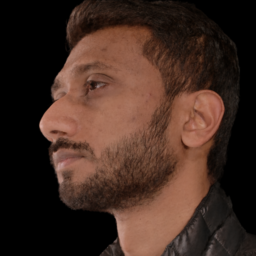} 
    \\   
    \\
\rotatebox{90}{\textbf{Ours}} & \includegraphics[height=\height]{figures/comparison/holobooth/ours_in2_0_pred} 
    & \includegraphics[height=\height]{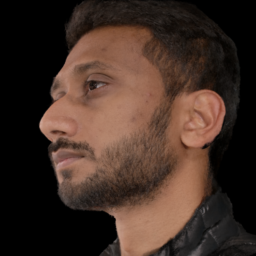} 
    & \includegraphics[height=\height]{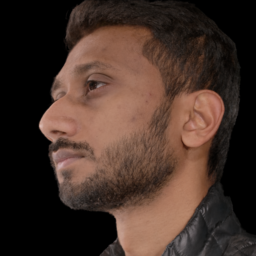} 
    & \includegraphics[height=\height]{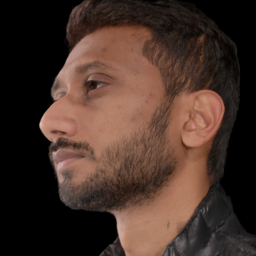} 
    \\
& 2 Views & 3 Views & 5 Views & 7 Views 
\end{tabular}
\end{center}

\caption{\label{fig:comparison_1k_suba} 
Comparison with related works at 1K resolution on our studio dataset. We compare with Learnit~\cite{tancik2021learned}, EG3D~\cite{eg3d}, RegNeRF~\cite{regnerf}, FreeNeRF~\cite{Yang2023FreeNeRF}, and KeypointNeRF~\cite{keypointnerf} on different number of input views ranging from two to seven. Please see Tbl.~\ref{tab:comp_holobooth} for metrics.}
\end{figure*}

%% file: figures/comparison/comparison_holobooth_subB.tex
\begin{figure*}[ht]

\begin{center}
\small
\setlength{\tabcolsep}{2pt}

\newcommand{\height}{2.8cm}
\begin{tabular}{cccccc}
& \textbf{Subject B}\\
\rotatebox{90}{GT}  & \includegraphics[height=\height]{figures/comparison/holobooth/gt_2_gt}  \\
\\
\rotatebox{90}{EG3D}
  & \includegraphics[height=\height]
 {figures/comparison/holobooth/eg3d_in2_2_pred}  
    & \includegraphics[height=\height]{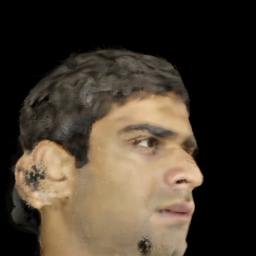} 
    & \includegraphics[height=\height]{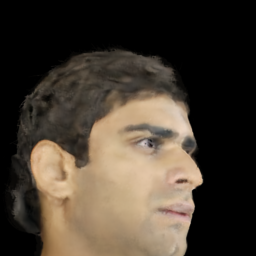} 
    & \includegraphics[height=\height]{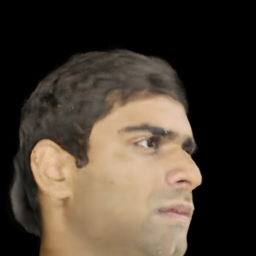} 
    \\    
\rotatebox{90}{Learnit}
  & \includegraphics[height=\height]
 {figures/comparison/holobooth/learnit_in2_2_pred}  
    & \includegraphics[height=\height]{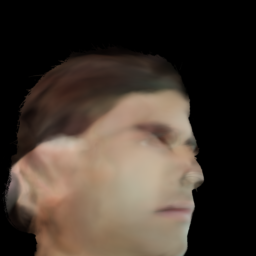} 
    & \includegraphics[height=\height]{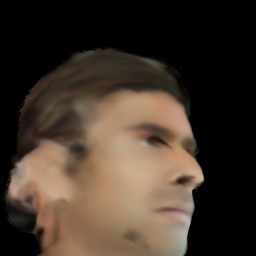} 
    & \includegraphics[height=\height]{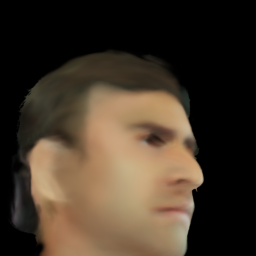} 
    \\    
\rotatebox{90}{KeypointNeRF}
  & \includegraphics[height=\height]
 {figures/comparison/holobooth/keypointnerf_in2_2_pred}  
    & \includegraphics[height=\height]{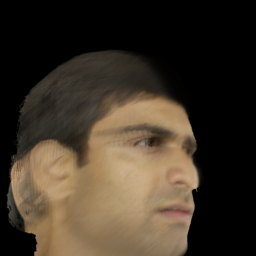} 
    & \includegraphics[height=\height]{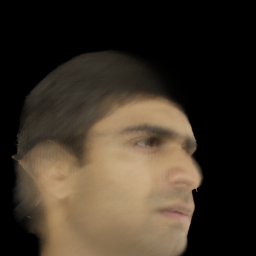} 
    & \includegraphics[height=\height]{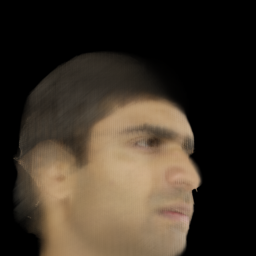} 
    \\    
 \rotatebox{90}{RegNeRF}& \includegraphics[height=\height]
{figures/comparison/holobooth/regnerf_in2_2_pred} 
    & \includegraphics[height=\height]{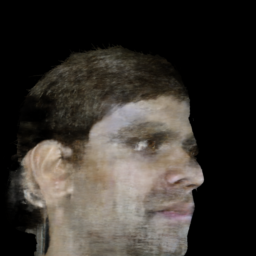} 
    & \includegraphics[height=\height]{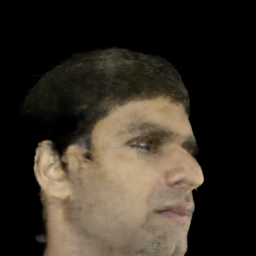} 
    & \includegraphics[height=\height]{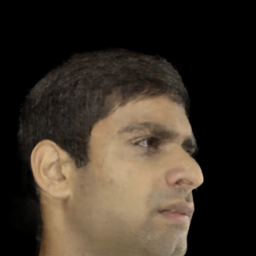} 
   \\
\rotatebox{90}{FreeNeRF}
  & \includegraphics[height=\height]
 {figures/comparison/holobooth/freenerf_in2_2_pred} 
    & \includegraphics[height=\height]{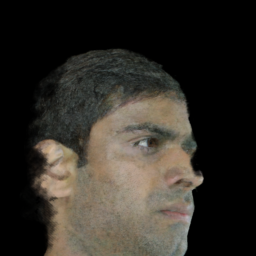} 
    & \includegraphics[height=\height]{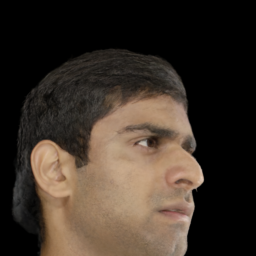} 
    & \includegraphics[height=\height]{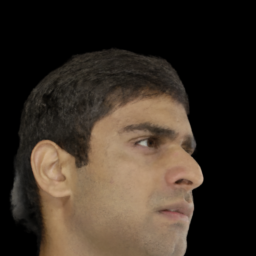} 
    \\    
    \\
\rotatebox{90}{\textbf{Ours}} & \includegraphics[height=\height]{figures/comparison/holobooth/ours_in2_2_pred} 
    & \includegraphics[height=\height]{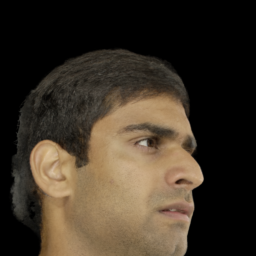} 
    & \includegraphics[height=\height]{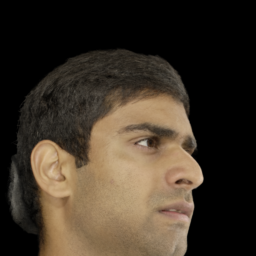} 
    & \includegraphics[height=\height]{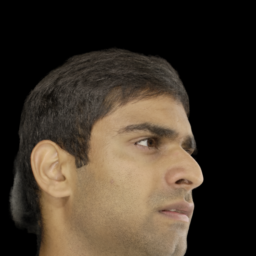} 
    \\    
    
& 2 Views & 3 Views & 5 Views & 7 Views 
\end{tabular}
\end{center}

\caption{\label{fig:comparison_1k_subb}Comparison with related works at 1K resolution on our studio dataset. We compare with Learnit~\cite{tancik2021learned}, EG3D~\cite{eg3d}, RegNeRF~\cite{regnerf}, FreeNeRF~\cite{Yang2023FreeNeRF}, and KeypointNeRF~\cite{keypointnerf} on different number of input views ranging from two to seven. Please see Tbl.~\ref{tab:comp_holobooth} for metrics.}

\end{figure*}

%% file: figures/comparison/comparison_holobooth_subC.tex
\begin{figure*}[ht]

\begin{center}
\small
\setlength{\tabcolsep}{2pt}

\newcommand{\height}{2.8cm}
\begin{tabular}{cccccc}
& \textbf{Subject C}\\
\rotatebox{90}{GT}  & \includegraphics[height=\height]{figures/comparison/holobooth/gt_1_gt}  \\ 
\\
\rotatebox{90}{EG3D}
  & \includegraphics[height=\height]
 {figures/comparison/holobooth/eg3d_in2_1_pred}  
    & \includegraphics[height=\height]{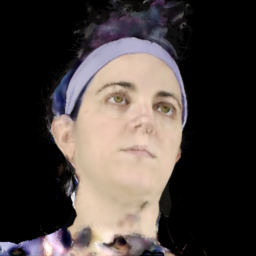} 
    & \includegraphics[height=\height]{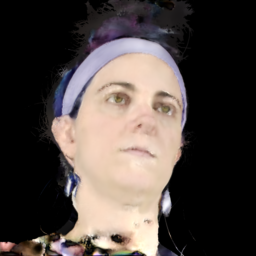} 
    & \includegraphics[height=\height]{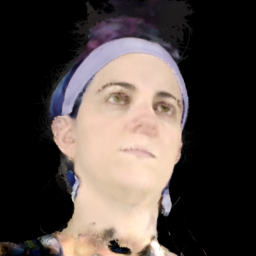} 
    \\
\rotatebox{90}{Learnit}
  & \includegraphics[height=\height]
 {figures/comparison/holobooth/learnit_in2_1_pred}  
    & \includegraphics[height=\height]{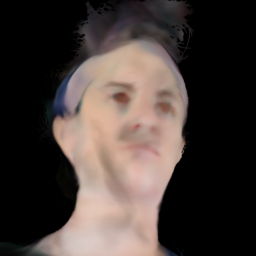} 
    & \includegraphics[height=\height]{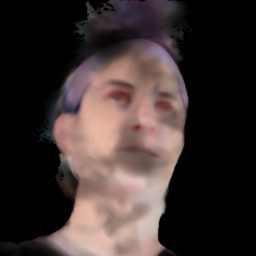} 
    & \includegraphics[height=\height]{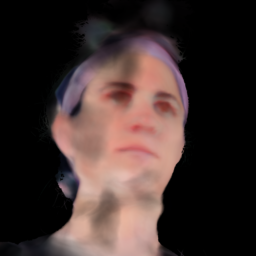} 
    \\    
\rotatebox{90}{KeypointNeRF}
  & \includegraphics[height=\height]
 {figures/comparison/holobooth/keypointnerf_in2_1_pred}  
    & \includegraphics[height=\height]{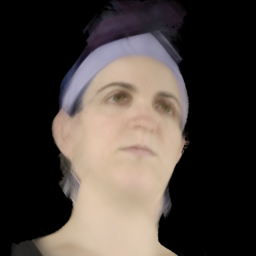} 
    & \includegraphics[height=\height]{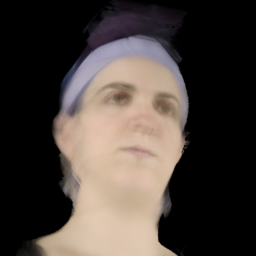} 
    & \includegraphics[height=\height]{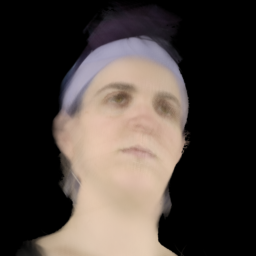} 
    \\    
\rotatebox{90}{RegNeRF}
 &\includegraphics[height=\height]
{figures/comparison/holobooth/regnerf_in2_1_pred} 
    & \includegraphics[height=\height]{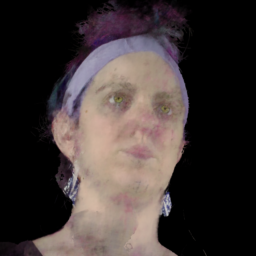} 
    & \includegraphics[height=\height]{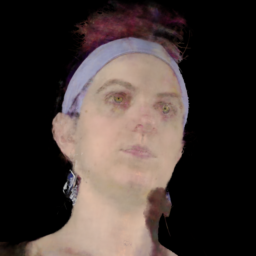} 
    & \includegraphics[height=\height]{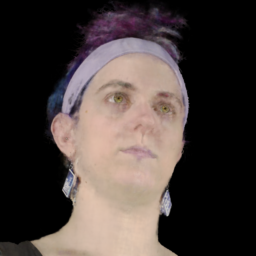} 
    \\    
\rotatebox{90}{FreeNeRF}
  & \includegraphics[height=\height]
 {figures/comparison/holobooth/freenerf_in2_1_pred} 
    & \includegraphics[height=\height]{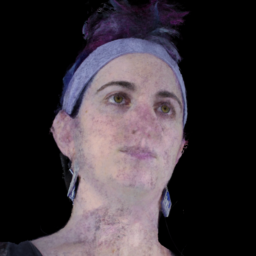} 
    & \includegraphics[height=\height]{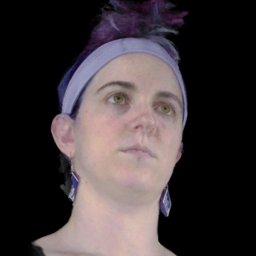} 
    & \includegraphics[height=\height]{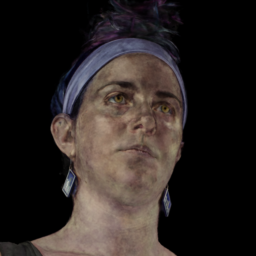} 
    \\    
    \\
\rotatebox{90}{\textbf{Ours}} & \includegraphics[height=\height]{figures/comparison/holobooth/ours_in2_1_pred} 
    & \includegraphics[height=\height]{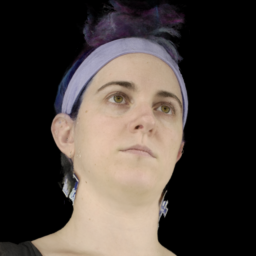} 
    & \includegraphics[height=\height]{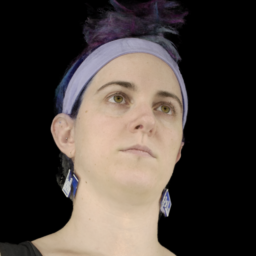} 
    & \includegraphics[height=\height]{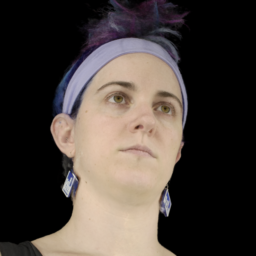} 
    \\
    
& 2 Views & 3 Views & 5 Views & 7 Views 
\end{tabular}
\end{center}

\caption{\label{fig:comparison_1k_subc}Comparison with related works at 1K resolution on our studio dataset. We compare with Learnit~\cite{tancik2021learned}, EG3D~\cite{eg3d}, RegNeRF~\cite{regnerf}, FreeNeRF~\cite{Yang2023FreeNeRF}, and KeypointNeRF~\cite{keypointnerf} on different number of input views ranging from two to seven. Please see Tbl.~\ref{tab:comp_holobooth} for metrics.}

\end{figure*}